\documentclass[twoside]{article}

\usepackage[accepted]{aistats2022}
\usepackage[utf8]{inputenc} 
\usepackage[T1]{fontenc}    
\usepackage{hyperref}       
\usepackage{booktabs}       
\usepackage{amsfonts}       
\usepackage{nicefrac}       
\usepackage{microtype}      
\usepackage{xcolor}         
\usepackage{xfrac}

\usepackage{microtype}
\usepackage{graphicx}
\usepackage{booktabs} 

\usepackage{float}

\usepackage{natbib}
\usepackage{subcaption}
\usepackage{hyperref}
\usepackage{url}
\usepackage{amsfonts}       
\usepackage{nicefrac}       
\usepackage{mathrsfs}
\usepackage{amsmath, amssymb, amsthm, amsfonts}
\usepackage{amssymb}
\usepackage{amsmath}
\usepackage{amsthm}
\usepackage{nicefrac}
\usepackage{enumerate}
\usepackage{graphicx}
\usepackage{epsfig}
\usepackage{tabu}
\usepackage{wrapfig}
\usepackage{empheq}
\usepackage{ragged2e}
\usepackage{multicol}
\usepackage{mathtools}

\newtheorem{theorem}{Theorem}[section]
\newtheorem{lemma}[theorem]{Lemma}

\newtheorem{definition}[theorem]{Definition}
\newtheorem{corollary}[theorem]{Corollary}
\newtheorem{fact}[theorem]{Fact}
\newtheorem{remark}[theorem]{Remark}

\newenvironment{informal-proof}{\textit{Informal Proof:}}{\hfill$\square$}

\numberwithin{equation}{section}

\usepackage[disable]{todonotes} 
\presetkeys{todonotes}{}{}

\usepackage{comment}
\usepackage{commath} 

\PassOptionsToPackage{hyphens}{url}

\usepackage{etoolbox} 
\usepackage{titlesec}

\newcommand{\tf}{ \Tilde{f} }           
\newcommand{\dnp}[1]{\Lambda_{np}^{ (#1) }}  
\newcommand{\tg}{\Tilde{g}}             
\newcommand{\al}{\tilde{L}}               
\newcommand{\qp}[2]{\tau_{#1} \left( #2 \right) } 
\newcommand{\td}{F^{*\prime}}                
\newcommand{\tarfunc}[1]{F^{#1}}            
\newcommand{\tarfunci}[1]{F^{*}_{#1} \rb{ \vecx{#1} }}
\newcommand{\tftp}[1]{\mu_{#1}^*}                            
\newcommand{\tfhw}[1]{ u_{#1}^* }								
\newcommand{\tfhmw}[1]{ v_{#1}^* }

\newcommand{\funci}[1]{f_{#1} ( x_{1:#1} )}
\newcommand{\approxfunci}[1]{ \tilde{f}_{#1} ( x_{1:#1} ) }
\newcommand{\approxfunc}{ \tilde{f} (x) }
\newcommand{\dfi}[1]{ \nabla_{#1} f_{#1} \rb{ \vecx{#1} } }
\newcommand{\dfit}[2]{ \nabla_{#1} f_{#1}^{#2} \rb{ \vecx{#1} } }
\newcommand{\dgi}[1]{ \nabla_{#1} g_{#1} \rb{ \vecx{#1} } }
\newcommand{\dgit}[2]{ \nabla_{#1} g_{#1}^{#2} \rb{ \vecx{#1} } }
\newcommand{\dfv}[1]{\nabla #1}

\newcommand{\qpi}[2]{ \tau_{#1} \rb{ \vecx{#2} } }
\newcommand{\nrmqpi}[2]{ \tilde{\tau}_{ #1 } \rb{ \vecx{#2} } }
\newcommand{\dfqpi}[2]{ \nabla_{#2} f_{#2} \rb{ \qpi{#1}{#2} } }
\newcommand{\dfqpit}[3]{ \nabla_{#2} f_{#2}^{#3} \rb{ \qpi{#1}{#2} } }
\newcommand{\dgqpi}[2]{ \nabla_{#2} g_{#2} \rb{ \qpi{#1}{#2} } }
\newcommand{\dgqpit}[3]{ \nabla_{#2} g_{#2}^{#3} \rb{ \qpi{#1}{#2} } }

\newcommand{\alm}[2]{\tilde{L} \rb{ \dfv{#1}, #2 } }

\newcommand{\vecx}[1]{x_{1:#1}}
\newcommand{\nni}[1]{N ( \vecx{#1}; \theta_{#1} ) }					 
\newcommand{\nnit}[2]{N (\vecx{#1}; \theta_{#1}^{#2} ) }					 
\newcommand{\pni}[1]{P ( \vecx{#1}; \theta_{#1} ) }					
\newcommand{\pnit}[2]{ P ( \vecx{#1}; \theta_{#1}^{#2} ) }		
\newcommand{\RN}[1]{%
  \textup{\uppercase\expandafter{\romannumeral#1}}%
}
\newcommand{\pnil}[2]{P_{#1} ( \vecx{#2}; \theta_{#2} ) }					
\newcommand{\pnilt}[3]{ P_{#1} ( \vecx{#2}; \theta_{#2}^{#3} ) }		

\newcommand{\xx}{\mathbf{x}}
\newcommand{\zz}{\mathbf{z}}

\newcommand{\E}{\mathbb{E}}             
\newcommand{\R}{\mathbb{R}}
\newcommand{\N}{\mathcal{N}}            

\newcommand{\mcx}{\mathcal{X}}
\newcommand{\mcf}{\mathcal{F}}          
\newcommand{\mcd}{\mathcal{D}}          
\newcommand{\empR}{\hat{ \mathcal{R} }} 

\newcommand{\initw}[1]{ \bar{w}_{#1} }
\newcommand{\initb}[1]{ \bar{b}_{#1} }
\newcommand{\inita}[1]{ \bar{a}_{#1} }
\newcommand{\initam}[2]{ \bar{a}_{#1, #2} }			
\newcommand{\initwm}[2]{ \bar{w}_{#1, #2} }			
\newcommand{\initbm}[2]{ \bar{b}_{#1, #2} }			
\newcommand{\dotp}[2]{ \langle #1, #2 \rangle }	
\newcommand{\nrmxi}[1]{ \tilde{x}_{1:#1} }								
\newcommand{\inittht}{ \bar{\theta} }
\newcommand{\initW}{\bar{W}}
\newcommand{\initB}{\bar{B}}

\newcommand{\LB}{\bar{ \Lambda }}
\newcommand{\mHB}[1]{ \overline{ \mathcal{H} }^{(#1)} }
\newcommand{\tht}[1]{\theta^{(#1)}}
\newcommand{\Wt}[1]{W^{(#1)}}
\newcommand{\Bt}[1]{B^{(#1)}}
\newcommand{\wt}[2]{ w_{#1}^{(#2)} }
\newcommand{ \bt }[2]{ b_{#1}^{(#2)} }

\newcommand{\xt}[1]{ x^{ (#1) } }
\newcommand{\lbd}[2]{\Lambda_{#1}^{ (#2) }}

\newcommand{\Nt}[1]{N^{(t)}}
\newcommand{\Pt}[1]{P^{(t)}}
\newcommand{\ft}[1]{f^{(#1)}}
\newcommand{\gt}[1]{g^{(#1)}}
\newcommand{\fpt}[1]{f^{\prime (#1) }}

\newcommand{\neurvalvector}[2]{ (\initW + \Wt{t}) #1 + \initB + \Bt{t} }
\newcommand{\neuralvalzero}{ \initw{r} x + \initb{r} }
\newcommand{\neuralvalvectorzero}{ \initW x + \initB }
\newcommand{\neurval}[2]{ ( \initw{r} + \wt{r}{t} ) #1 + \initb{r} + \bt{r}{t} }                              

\newcommand{\neuralvalzerom}[1]{ \dotp{ \initwm{#1}{r} }{ \nrmxi{#1} } + \initbm{#1}{r} }
\newcommand{\neuralvalzeromwv}[1]{ \dotp{ \initwm{#1}{r} }{ \vecx{#1} } + \initbm{#1}{r} }		
\newcommand{\neurvalm}[3]{ \dotp{ \initwm{#1}{r} + w_{#1, r}^{#2} }{ #3 } + \rb{ \initbm{#1}{r} + b_{#1, r}^{#2} }  }
\newcommand{\zeroone}[1]{ \pmb{\alpha}_{#1}}				

\newcommand{\squarefunc}{\zeta}

\newcommand{\RomanNumUp}[1]{\text{\MakeUppercase{\romannumeral #1}}}                                    
\newcommand{\RomanNumLow}[1]{\text{\MakeLowercase{\romannumeral #1}}}                                   
\newcommand{\onevector}{\mathbf{1}}
\newcommand{\ind}[1]{\mathbb{I} \left[ #1 \right] }  
\newcommand{\detm}[1]{\det \left( #1 \right)  }
\newcommand{\rb}[1]{ \left( #1 \right) }        
\newcommand{\sqb}[1]{ \left[ #1 \right] }
\newcommand{\deff}{ \mathclap{\normalfont \footnotesize \mbox{def}} } 
\newcommand{\comp}[2]{ \mathfrak{C}_{ #1 } \rb{ #2 } }

\newcommand{\cgl}{\Gamma} 

\newcommand{\samples}{\mathcal{X}}

\newcommand{\deriv}{\mathrm{d}}

\newcommand{\relu}{\ensuremath{\mathsf{ReLU}}}
\newcommand{\elu}{\ensuremath{\mathsf{ELU}}}
\newcommand{\eluone}{\ensuremath{\mathsf{ELU}+1}}

\makeatletter
\newcommand{\changeoperator}[1]{%
  \csletcs{#1@saved}{#1@}%
  \csdef{#1@}{\changed@operator{#1}}%
}
\newcommand{\changed@operator}[1]{%
  \mathop{%
    \mathchoice{\textstyle\csuse{#1@saved}}
               {\csuse{#1@saved}}
               {\csuse{#1@saved}}
               {\csuse{#1@saved}}%
  }%
}
\makeatother

\changeoperator{sum}

\newcommand{\scriptfrac}{\genfrac{}{}{}2}
\newcommand{\inlinefrac}[2]{\scriptfrac{#1}{#2} }

\begin{document}

%
\runningtitle{Learning and Generalization in Overparameterized Normalizing Flows}

%
\runningauthor{Kulin Shah, Amit Deshpande, Navin Goyal}

\twocolumn[

\aistatstitle{Learning and Generalization in Overparameterized Normalizing Flows}

\aistatsauthor{ Kulin Shah \And Amit Deshpande \And  Navin Goyal }

\aistatsaddress{ Microsoft Research India \And  Microsoft Research India \And Microsoft Research India } ]


\begin{abstract}
  In supervised learning, it is known that overparameterized neural networks with one hidden layer provably and efficiently learn and generalize, when trained using stochastic gradient descent with a sufficiently small learning rate and suitable initialization. In contrast, the benefit of overparameterization in unsupervised learning is not well understood. Normalizing flows (NFs) constitute an important class of models in unsupervised learning for sampling and density estimation. In this paper, we theoretically and empirically analyze these models when the underlying neural network is a one-hidden-layer overparametrized network. Our main contributions are two-fold: (1) On the one hand, we provide theoretical and empirical evidence that for constrained NFs (this class of NFs underlies many NF constructions) with the one-hidden-layer network, overparametrization hurts training. (2) On the other hand, we prove that unconstrained NFs, a recently introduced model, can efficiently learn any reasonable data distribution under minimal assumptions when the underlying network is overparametrized and has one hidden-layer.

\end{abstract}

\section{Introduction}
 
Neural network models trained using gradient-based algorithms have been very effective in both supervised and unsupervised learning.
This is surprising for two reasons:
First, the optimization of training loss is typically non-smooth and non-convex and yet gradient-based methods often succeed in making the training loss very small. Second, even large neural networks whose number of parameters are more than the size of training data often generalize well on the unseen test data, instead of overfitting the seen training data. Recent work in supervised learning attempts to  theoretically analyze these phenomena. 

In supervised learning, the empirical risk minimization with quadratic or cross-entropy loss is a non-convex optimization problem even for one hidden layer fully connected network. In the last few years, it was realized that when the network is overparametrized, i.e. the hidden-layer size is large compared to the dataset size or some measure of complexity of the data, one can provably show efficient training and generalization for these networks. This hinges on the fact that overparametrization makes the optimization problem close to a convex one. See, e.g., \cite{JacotNTK, du2018gradient, allen2019learning, quanquan, Arora_Generalization}. 

The role of overparameterization and its effect on provable training and generalization guarantees for neural networks is far less understood in unsupervised learning. Generative modeling of a probability distribution when we are given samples drawn from that distribution is an important, classical problem in statistics and unsupervised learning. The goal of a generative model is to generate new samples from the distribution and give a probability density estimate at any queried point. Popular categories of generative models based on neural networks include Generative Adversarial Networks (GANs) \cite{GoodfellowGenerative2014}, Variational AutoEncoders (VAEs) (e.g., \cite{KingmaWelling2014}), and Normalizing Flows (NFs) (e.g., \cite{RezendeMohamed2015}). All categories of models, especially GANs, have shown an impressive capability to generate samples of photo-realistic images but GANs and VAEs cannot give probability density estimates for new data points. All categories present various challenges in training such as mode collapse, posterior collapse, training instability, etc., e.g., \cite{Bowman2016,Salimans2016,arora2018do,lucic2018are}.  

Unlike GANs and VAEs, NFs can do both sampling and density estimation, leading to a potentially wider range of applications; see, e.g., the surveys \cite{KobyzevSurvey, PapamakariosSurvey}. 
Theoretical understanding of learning and generalization in generative models remains a natural and important open question even after some recent work (\cite{buhai2020benefits, kong2020expressive, risteski2020representationalNF, lee2021universal}). 
Appendix~\ref{sec:related-work} contains further literature review. In this paper, we focus on the theoretical analysis of NFs. For \emph{constrained} NFs which underlies a large class of NF constructions, we show that theoretical analysis in the overparametrized regime runs into difficulties. This is also seen in experiments where overparametrization hurts the performance of constrained NFs in many settings. In contrast, a recent class of NFs called \emph{unconstrained} NFs, admits provable training and generalization guarantees in the overparametrized setting. 
Before stating our contributions in detail, we introduce NFs followed by a very brief discussion of overparametrized supervised learning to provide the necessary context. 

\textbf{Normalizing Flows.}
The general idea behind normalizing flows (NFs) is as follows: let $X \in \R^d$ be a random variable coming from the data distribution and $Z \in \R^d$ be a random variable associated with \emph{base} distribution which can be the standard Gaussian or exponential distribution. Given i.i.d. samples of $X$, the goal is to learn a differentiable invertible map $f_X: \R^d \to \R^d$ that transports the distribution of $X$ to the distribution of $Z$: in other words, the distribution of $f_X^{-1}(Z)$ and $X$ are same. (We tacitly assume that the distribution of $X$ is nice enough to allow for the existence of $f_X$.) We assume that function $f_X$ is autoregressive, means $f_X$ is of the form $f_X(x) = \rb{f_{X, 1}(x_1), f_{X, 2}(\vecx{2}), \ldots, f_{X, d}(\vecx{d})}$ where $f_{X, i}: \R^i \to \R$ and $\vecx{i}$ is first $i$ dimension of a data sample $x$ from $X$ (i.e., if $x=\rb{ x_1, x_2, \ldots, x_d }$, then $\vecx{i} = \rb{ x_1, \ldots, x_i }$). 
The nice thing about autoregressive functions is that their invertibility is easily ensured by making $f_{X, i}(\vecx{i})$ a strictly monotonically increasing function in $x_i$ for any fixed value of $\vecx{(i-1)}$. 
We will call such an $f$ \emph{monotonic autoregressive function}. Such a function is also called a Knothe--Rosenblatt map and is known to exist and be unique under very general conditions sufficient for our purposes, in particular for any pair of probability measures on $\R^d$ with density; see Chapter 2 in \cite{Santambrogio}. 

Learning of $f_X$ is done by representing a monotonic autoregressive map $f$ by neural networks, setting up an appropriate loss function, and doing gradient-based training with the aim of achieving $f = f_X$. A number of approaches have been suggested for carrying out this general plan. We distinguish between two classes of approches: (1) \emph{Represent $f$ directly using neural networks.}  In this approach there are $d$ neural networks $N_1, \ldots, N_d$ with $f_i(x_{1:i}) = N_i(x_{1:i})$. Since the functions represented by standard neural networks are not necessarily monotone, the design of the neural network is \textit{constrained} to make it monotone. For example, if $\{ a_r, w_r, b_r \}_{r=1}^m$ are the parameters of the neural networks, with $a_r, w_r, b_r \in \R$ for each $r$, and $\rho$ is a monotonically increasing activation function, then the univariate one-hidden layer network of the form $\sum_{r=1}^m a_r \, \rho \rb{ w_r x + b_r }$ can be made monotonically increasing by ensuring positivity of $a_r$ and $w_r$. This can be done in multiple ways: for example, instead of $a_r, w_r$, one can use $a_r^2, w_r^2$ in the above expression; 
see, e.g., \citep{huang2018neural, cao2019block}. (2) \emph{Represent the Jacobian matrix $\inlinefrac{ \partial f(x) }{ \partial x }$ using neural networks.} In this approach, we model diagonal entries of the Jacobian by neural networks $\inlinefrac{ \partial f_i \rb{ \vecx{i} } }{ \partial x_i } = \phi ( N_i \rb{ \vecx{i} } )$ where $\phi: \R \to \R^+$ takes on only positive values. Positivity of $\inlinefrac{ \partial f_i \rb{ \vecx{i} } }{ \partial x_i } $ implies monotonicity of $f_i \rb{ \vecx{i} }$ with respect to $x_i$. Note that the parameters are \textit{unconstrained} in this approach. This approach is used by \cite{FrenchPaper}.

We will refer to the models in the first class as \emph{constrained normalizing flows} (\textbf{CNFs}) and those in the second class as \emph{unconstrained normalizing flows} (\textbf{UNFs}). 

Most existing analyses for overparametrized neural networks in the supervised setting consider a linear approximation of the neural network, termed \emph{pseudo-network} in \cite{allen2019learning}. 
The convexity property of loss function for pseudo-network and closeness between neural network and pseudo network help in proving convergence and generalization of neural network.

\subsection{Our Contributions} 

In this paper, we study both CNFs and UNFs theoretically when the underlying network has one hidden-layer and empirically validate our theoretical findings. We now describe our contributions.

\textbf{Architectural variants.} 
The practical CNF and UNF architectures can be quite detailed involving multiple layer neural networks and stacking of flows. It is difficult to get a theoretical handle on such models---presently there are no satisfactory results even for  two-hidden layers networks in the supervised learning setting. In this paper, we identify very simple and natural NF models (gleaned from the existing architectures) reducing the architecture to the essentials and yet providing satisfactory results in experiments. These models are the starting point of our analyses.
A natural approach to analyze NFs is to adapt the successful techniques from supervised learning to NFs. 
While there is a natural definition of pseudo-network in the case of CNFs, for UNFs this is not clear. We are able to define linear approximations of the neural network to analyze the training of both CNFs and UNFs. 
However, one immediately encounters some new roadblocks: the loss surface of the pseudo-networks is non-convex in both CNFs and UNFs for the simple NF models mentioned above.
Therefore, analyzing pseudo-networks still remains difficult.
Barring a major breakthrough in non-convex optimization for deep learning, one way to proceed is to find architectural variants of simple NFs that may lead to pseudo-networks with convex optimization problems without adverse effect on their empirical performance. We follow this path and identify novel variations that make the optimization problem for associated pseudo-network convex. It is pertinent that our variations are arguably natural.

\textbf{Architectural variants for CNFs.} To resolve the non-convexity arising from using $a_r^2, w_r^2$ as parameters, we simply impose the constraints $a_r \geq \epsilon$ and $w_{r} \geq \epsilon$ for all $r \in [m]$ where $[m] = \{1, \ldots, m\}$. To solve this constrained optimization  problem, we use projected SGD, which in this case incurs essentially no extra cost over SGD due to the simplicity of the constraints. In our experiments, this variation slightly \emph{improves} the training of NFs compared to the reparametrization approach mentioned above and may be of a separate interest in practical settings. 

\textbf{Architectural variants for UNFs.} Similarly, for UNFs we identify two problems in the model of \cite{FrenchPaper} that  make the theoretical analysis difficult. We resolve these as follows: (1) Change in numerical integration method. Instead of Clenshaw--Curtis quadrature method for numerical integration employed in \cite{FrenchPaper}, we use the simple rectangle quadrature. This change makes the model slightly slower (in our experiments, it typically uses twice as many samples and time to get similar performance). (2) Change in the base distribution. We use the exponential distribution as the base distribution instead of the standard Gaussian distribution. In experiments, this does not cause any changes in performance. Note that NFs require \textit{only} efficient sampling and density estimation from the base distribution but the Gaussian is far from the only distribution to have those properties. 

Our results about these variants point to a dichotomy between these two classes of NFs: 

\textbf{Overparametrization hurts CNFs.} Our theoretical findings provide evidence that overparametrization makes training slower. To be more precise, we show that in a bounded number of training iterations or for bounded change in weights such that neural networks and pseudo networks are close, overparameterized CNFs can not learn the target function.
We also point out 
the reasons that lead overparametrization to adversely affect the training of CNFs. Our experimental results also validate our theoretical results and confirm that overparameterization in CNF makes training slower. Note that in supervised learning, it is known that overparameterization makes training faster \citep{neyshabur2015search, allen2019learning}. Therefore, the finding that overparametrization is significantly detrimental to CNFs is \textit{novel} and we are not aware of \textit{any other} settings where overparametrization has such a strong negative effect. 
Thus, for theoretical analysis of CNFs, one must work with moderate-sized networks. 
But this is likely to be difficult as analysis of such networks has remained open even for supervised learning leading us to a ``barrier''. 

\textbf{Analysis of overparametrized UNFs.} We theoretically analyze UNFs and prove that overparameterized networks for UNFs indeed learn the data distribution. To our knowledge, this is the \textit{first ``end-to-end'}' analysis of an NF model---and in fact for \emph{any} neural generative model using gradient-based algorithms for a sufficiently large class of distributions (please see Appendix~\ref{sec:related-work} for additional extensive related work). This proof, while following the high-level scheme of supervised learning proofs, requires several new ideas, conceptual as well as technical, due to different settings and will be discussed in the sequel.

To summarize, our contributions include:
\begin{itemize}
	\item We identify difficulties in the theoretical analysis of existing NF models. We resolve these by proposing new versions of these models without loss of experimental efficacy.
	\item We identify a ``barrier'' to the training convergence and generalization analysis of CNFs: overparametrization is detrimental to CNFs. 
	\item We provide efficient training convergence and generalization analysis for UNFs. To our knowledge, this is the \textit{first} result on training and generalization of NFs.
	\item We experimentally validate our theoretical claims. 
\end{itemize}

\paragraph{Paper outline.} 
Sec.~\ref{sec:Prelim_Results} contains preliminaries, Sec.~\ref{section:constrained-normalizing-flow-main} contains our results on CNFs and Sec.~\ref{sec:unconstrained-normalizing-flow} contains results on UNFs. Sec.~\ref{section:experiments} briefly describes our empirical studies. We conclude in \ref{sec:conclusion}. Appendix \ref{sec:outline-appendix} contains outline of the appendix. 

\section{Preliminaries}\label{sec:Prelim_Results}
In this section, we will continue our description of the problem of learning probability distributions using NFs and introduce necessary notation.
\subsection{Problem of learning distributions in Normalizing Flows}
Recall that the goal of NFs is to learn a probability distribution given via i.i.d. samples from the distribution. Let $X$ be 
the random variable corresponding to the data distribution we want to learn. We denote the probability density (we often just say density) of $X$ at $u \in \R^d$ by $p_X(u)$. We will work with distributions whose densities have a finite support.\footnote{ This is often without any real loss of generality because, for most purposes, light-tailed distribution (e.g., the Gaussian distribution) can be assumed to have a finite support. (Exception to this are heavy-tailed distributions which are seldom encountered; we believe our work here could be extended to deal with such distributions too.)} We will furthermore assume $p_X(u) = 0$ when $\norm{u}_2 \geq 1$, without loss of generality. Let $Z$ be a random variable with either standard Gaussian or the \emph{standard exponential distribution}. There seems to be no well-accepted definition of multidimensional exponential distribution; for our purposes the following natural definition will serve well.
The density of the standard exponential distribution at $z = \rb{ z_1, z_2, \ldots, z_d } \in \R^d$ is given by $e^{- \sum_{i=1}^d z_i}$ when all $z_i \geq 0$, and by $0$, otherwise. We will refer to the distribution of  $Z$ as the \emph{base} distribution. 

Let $f: \R^d \to \R^d$ be monotonic autoregressive as defined previously; thus, $f$ is invertible. 
Let $p_{f, Z}(\cdot)$ be the density of the random variable $f^{-1}(Z)$. 
Let $z = f(x)$.
Then the standard change of density formula using the invertibility of $f$ gives 
\begin{align}\label{eqn:change_of_variable}
    p_{f, Z}(x) = p_Z( f(x) ) \; \Big| \det \rb{ \frac{ \partial f(x) }{ \partial x } } \Big| .
\end{align}
We would like to choose $f$ so that $p_{f, Z}=p_X$. As mentioned before, such an $f$ always exists and is unique and we will denote it by $\tarfunc{*}$. 
If we can find $\tarfunc{*}$, then we can generate samples of $X$ using $F^{*-1}(Z)$ since generating the samples of $Z$ is easy and so is the inversion of $F^*$ using monotonic autoregressive property. 
Similarly, we can evaluate density $p_X (x)$ using standard change of variable with $\tarfunc{*}$ because $p_{\tarfunc{*}, Z}(x) = p_X (x)$.
To find $\tarfunc{*}$ from the data, we set up the maximum log-likelihood objective:
\begin{align} \label{eqn:LL}
    \nonumber & \max_f \tfrac{1}{n} \sum_{x \in \samples} \log p_{f, Z}( x ) \\
    & = \max_f \tfrac{1}{n} \big[ \sum_{x \in \samples} \log p_{Z}(f( x )) + \sum_{ x \in \samples } \log \big( \det \big( \tfrac{\partial  f(x) }{ \partial x } \big) \big) \big],
\end{align}
where training set $\samples \subset \R^d$ contains $n$ i.i.d. samples of $X$, and the maximum is over differentiable invertible functions. 
When $Z$ is standard exponential and $f$ is monotonic autoregressive, 
then \eqref{eqn:LL} simplifies to
\begin{align}\label{eqn:LL_exponential}
\nonumber \min_f \, & L(f, \samples) = \tfrac{1}{n} \sum_{x \in \samples} L(f, x)  \text{ and } \\ 
& L (f, x) = \sum_{i=1}^d \big( \funci{i} - \log  \big( \tfrac{\partial \funci{i} }{ \partial x_i } \big) \big).
\end{align} 

We denote average loss by $L (f, \samples) = \inlinefrac{1}{n} \sum_{x \in \samples} L(f, x)$. Informally, we expect that as $n \to \infty$, the optimum $f_n$ in the above optimization problem satisfies $p_{f_n, Z} \to p_X$. 
To make the above optimization problem tractable, instead of $f$ we work with $d$ neural networks $N_1, N_2, \ldots, N_d$ as previously touched upon in our brief description of CNFs and UNFs. 
All our networks will have one hidden layer with the following basic form:
\begin{align*}
N\rb{ x; \theta } = \sum_{r=1}^m \inita{r} \, \rho ( \dotp{ \initw{r} + w_{r} }{ x } + ( \initb{r} + b_{r} ) ).
\end{align*}
 
Here $m$ is the size of the hidden layer, $\rho$ is a strictly increasing activation function, the weights $\inita{r}, \initw{r}, \initb{r}$ are the initial weights chosen at random according to some distribution specified later, and $w_{r}, b_{r}$ are offsets from the initial weights. We only train $w_{r}$ and $b_{r}$, and the outer weights remain frozen at their initial values.
Let $\inittht=(\initw{1}, \ldots, \initw{m}; \initb{1}, \ldots, \initb{m})$ denote the vector of initial parameters 
and similarly $\theta = (w_1, \ldots, w_m; b_1, \ldots, b_m)$ denote the matrix of offsets from the initial weights. 
Similarly, we denote offsets at time step $t$ by $\tht{t}$ and the corresponding network by $\Nt{t}(x)$ or $N( x; \tht{t})$. 

 %
 
\subsection{Supervised learning analysis} 
\label{subsec:supervised-learning-analysis}

 
 We now very briefly outline a proof technique for analyzing training and generalization for one-hidden layer neural networks for supervised learning (e.g. \cite{allen2019learning}). 
 For simplicity, we restrict the discussion to the realizable setting. 
 Data $x \in \R^d$ is generated by some distribution $D$ and the labels $y = h(x)$ are generated by some unknown function $h: \R^d \to \R$. The function $h$ is assumed to have small ``complexity'' $C_h$ which (informally speaking) measures the required size of a one-hidden-layer neural network with smooth activations to approximate $h$. The loss function is the square loss on the training set $\mcx$, that is, $L_s(\Nt{t}, \mcx) = \inlinefrac{1}{n} \sum_{x \in \mcx } L_s(\Nt{t}, x)$ with $L_s(\Nt{t}, x) = (N(x; \tht{t}) -y)^2$. The training is done using SGD to update the parameters $\theta$ of the neural network. 
 
 The problem of optimizing the square loss is non-convex even for one-hidden layer networks. One instead works with the \emph{pseudo-network} $P(x; \theta)$ which is the linear approximation of $N(x; \theta)$:
 \begin{align*}
    P(x; \theta) =& \sum_{r=1}^m \inita{r} ( \rho( \dotp{ \initw{r} }{ x } + \initb{r} ) \\ 
    &+ \rho'( \dotp{ \initw{r} }{ x } + \initb{r} ) \rb{ \dotp{ w_r }{ x } + b_r } ).
 \end{align*}
Similarly to $\Nt{t}$ and $N(x; \tht{t})$, we can also define $\Pt{t}$ and $P(x; \tht{t})$ with parameters $\tht{t}$.  When the network is overparameterized, i.e. the network size $m$ is sufficiently large compared to $C_h$, and the learning rate is small ($\eta = O(\sfrac{1}{m})$), SGD iterates when applied to $L_s(\Nt{t}, x^{(t)})$ and $L_s(\Pt{t}, x^{(t)})$ remain close throughout. Moreover, the problem of optimizing $L_s(\Pt{t}, \samples)$ is a convex problem in $\tht{t}$ for all $t$ and thus can be analyzed with the existing methods. An \emph{approximation} theorem then states that there exist parameters 
 $\theta^*$ with small norm such that the pseudo-network with parameters $\theta^*$ is close to the target function. This together with the analysis of SGD shows that the pseudo-network, and hence the neural network too, achieves small training loss. 
 Then by a Rademacher complexity argument that the neural network after $T = O(\sfrac{C_h}{\epsilon^2 } )$ time steps has population loss within $\epsilon$ of the optimal loss, thus obtaining a generalization result.


\section{Constrained Normalizing Flow} \label{section:constrained-normalizing-flow-main}
In this section, we will first describe problems in analyzing current CNF architectures. Then, we will describe a new architectural variant which is easy to analyze and our theoretical result on CNF. 

\subsection{Problems in analyzing CNF architectures}

In CNFs, monotonic autoregressive functions $f(x) = ( \funci{1}, \funci{2}, \ldots, \funci{d} )$ are represented by $d$ neural networks
via $\funci{i} = N_i(x_{1:i}) = \nni{i}$ where $\nni{i}$ is given by
\begin{align*}
\nni{i} = \tau \sum_{r=1}^m \initam{i}{r} \, \rho ( \dotp{ \initwm{i}{r} + w_{i, r} }{ \vecx{i} } + \left( \initbm{i}{r} + b_{i, r} ) \right),
\end{align*} 
where $\tau$ is a normalization constant chosen to compensate for the effect of overparameterization. We use $\theta_i$ to denote parameters of $N_i(x_{1:i})$ and $\theta$ to denote parameters of all neural networks. To make $\funci{i}$ monotonically increasing in $x_i$ for each fixed $x_{1:i-1}$, we ensure that $\initam{i}{r, i} \geq 0$, $\initwm{i}{r, i} + w_{i, r, i} \geq 0$ for all $r$. One way to do this is by replacing $\initam{i}{r}$ and $\initwm{i}{r, i}+w_{i, r, i}$ by their functions that take on only positive values. For example, the square function would give us the neural network
\begin{align*}
N_i(x_{1:i}) = \tau \sum_{r=1}^m \initam{i}{r}^2 \, \rho( \dotp{ \squarefunc \rb{ \initwm{i}{r} + w_{i, r} } }{ \vecx{i} } + \initbm{i}{r} + b_{i, r} ),
\end{align*}
where $\squarefunc : \R^i \to \R^i$ is given by $\squarefunc \rb{ y_1, y_2, \ldots, y_i } = \rb{ y_1, \ldots, y_{i-1}, y_i^2 }$ . After reparameterization, parameters have no constraints, and so this network can be trained using SGD. But we need to specify the (monotone) activation $\rho$ to complete our description of CNF. 

\paragraph{Activation function.} Unlike supervised learning, the choice of the activation function needs more care for CNFs as we will now see. 
Let $\sigma(x)$ denote the {\relu} activation. If we choose $\rho = \sigma$, then in \eqref{eqn:LL_exponential} we have 
\begin{align*}
 \frac{\partial \funci{i}}{\partial x_i} 
   =& \; \tau \sum_{r=1}^m \initam{i}{r}^2 \, ( \initwm{i}{r, i} + w_{i, r, i} )^2 \mathbb{I} [ \dotp{ \squarefunc ( \initwm{i}{r} \\
   & + w_{i, r} ) }{ \vecx{i} } + \initbm{i}{r} + b_{i, r} \geq 0 ].
\end{align*}
The derivative $\inlinefrac{\partial \funci{i}}{\partial x_i} $ and consequently $\log ( \det ( \inlinefrac{\partial f(x)}{\partial x} ) )$ are discontinuous functions of $x$ and $\theta$. Gradient-based optimization algorithms are not applicable to problems with discontinuous objectives, and indeed this is reflected in experimental failure of such models. By the same argument, any activation with a discontinuous derivative is not admissible. Convex activations with continuous derivative (e.g.  $\elu(x)$) also cannot be used because then $\nni{i}$ is also a convex function of $x_i$, which need not be the case for the optimal $f$. Hence in such cases, $\nni{i}$ can not approximate $f$.
To our knowledge, among the commonly used activations $\tanh$ (and the closely-related sigmoid) is the only one that does not suffer from either of these defects and also works well in practice \cite{BlockAutoRegressive}. 

\paragraph{Non-convexity of pseudo-network.} Pseudo-network with activation $\tanh$ is given by
\begin{align*}
  \pni{i} & =  \; \tau  {\textstyle \sum_{r=1}^m } \initam{i}{r}^2 \big( \tanh ( \dotp{ \squarefunc \rb{ \initwm{i}{r} } }{ \vecx{i} } + \initbm{i}{r} ) \\
    &+ \tanh' ( \dotp{ \squarefunc \rb{ \initwm{i}{r} } }{ \vecx{i} } + \initbm{i}{r} ) \; ( \dotp{ \squarefunc \rb{ \initwm{i}{r} + w_{i, r} } \\ 
    &- \squarefunc \rb{ \initwm{i}{r} } }{ \vecx{i} } + b_{i, r}  ) \big).
\end{align*}
      Note that $\pni{i}$ is not linear in $w_{i, r}$. Hence, it is not obvious that the loss function for the pseudo-network will remain convex in parameters; indeed, non-convexity can be confirmed in experiments. 


\subsection{A variant of CNF architecture}


To overcome the non-convexity issue, we propose another formulation of CNFs. Here we use standard form of the neural network, but ensure the constraints $\initam{i}{r} > 0$ 
and $\initwm{i}{r, i} > 0$ by the choice of the initialization distribution and $\initwm{i}{r, i} +w_{i, r, i} \geq \epsilon$ by using \emph{projected SGD} for optimization. 
\begin{align*}
&\nni{i} \\ =& \; \tau \sum_{r=1}^m \initam{i}{r} \, \tanh \big( \dotp{  \initwm{i}{r} + w_{i, r} }{ \vecx{i} } + ( \initbm{i}{r} + b_{i, r} ) \big), \\
& \text{with constraints $\initwm{i}{r, i} + w_{i, r, i} \geq \epsilon$, for all $r$}.
\end{align*}
$\epsilon>0$ is a small constant to ensure strict monotonicity of $\nni{i}$. These constraints are very simple and projected SGD incurs very little overhead. The pseudo-network in this formulation is given by
$$\pni{i} = P_c (\vecx{i})  + \pnilt{\ell}{i}{}$$ with constraints $\initwm{i}{r, i} + w_{i, r, i} \geq \epsilon$ for all $r$, where 
\begin{align*}
P_c (\vecx{i}) = & \; \tau \sum_{r=1}^m \initam{i}{r}  \tanh ( \dotp{ \initwm{i}{r}  }{ \vecx{i} }  + \initbm{i}{r} ) \;\; \text{ and } \\
\pnilt{\ell}{i}{} = & \; \tau \sum_{r=1}^m \initam{i}{r} \tanh' ( \dotp{ \initwm{i}{r}  }{ \vecx{i} } \\
& + \initbm{i}{r} ) \rb{ \dotp{ w_{i, r} }{ \vecx{i} } + b_{i, r}  }.
\end{align*}
Pseudo-network $\pni{i}$ is linear in $\theta_i$, therefore the objective 
in \eqref{eqn:LL_exponential} with $f_i$ replaced by $\pni{i}$ is convex in $\theta_i$ and hence, in $\theta$.
Note that $P_c (\vecx{i})$ does not change during training, therefore $\pnilt{\ell}{i}{}$ must approximate the target function with $P_c( \vecx{i} )$ subtracted. 

\subsection{Theoretical analysis of CNF}
Our results for CNFs are negative: we identify barriers in the analysis of highly over-parameterized CNFs and show that surmounting these barriers entails analyzing moderately overparameterized neural networks---a long-open problem even in supervised learning. 
Let $\tarfunc{*}$ denote the target function and $C(\tarfunc{*})$ denote some complexity measure of $\tarfunc{*}$. Initial weights $\initam{i}{r}$ and $\initwm{i}{r, i}$ are sampled from \emph{half-normal} distribution with parameters $\rb{0, \epsilon_a^2}$ and $\rb{0, \sigma_{wb}^2}$, respectively. The half-normal random variable $Y$ with parameters $\rb{\mu, \sigma^2}$ is given by simply $\abs{Y^\prime}$ where $Y^\prime \sim \N \rb{\mu, \sigma^2}$. Here $\N \rb{\mu, \sigma^2}$ denote the Gaussian distribution with mean $\mu$ and variance $\sigma^2$.  The bias term $\initbm{i}{r}$ is sampled from $\N \rb{0, \sigma_{wb}^2 }$.  We divide our analysis into two cases based on the value of $\sigma_{wb}$: (1) $\sigma_{wb}$ is between $\inlinefrac{1}{ \sqrt{m} } $ and $  \inlinefrac{\epsilon}{  C \rb{ \tarfunc{*} } \sqrt{ \log \rb{ m d } } } $, (2) $\sigma_{wb}$ is between $\inlinefrac{\epsilon}{ C ( \tarfunc{*} ) \sqrt{ \log \rb{ m d } } } $ and $1$. In case (1) we have: 


\begin{theorem}
\label{theorem:smaller-variance-problem-cnf}

For any $\epsilon > 0$, for any $i \in [d]$, any hidden layer size $m \geq \Omega \big( \mathrm{poly} \rb{ C( \tarfunc{*} ), \inlinefrac{1}{\epsilon}} \big)$, by choosing learning rate $\eta = O \big( \inlinefrac{\epsilon}{ m \tau \epsilon_a^2 \log m } \big)$ and $T= O ( \inlinefrac{ C( \tarfunc{*} ) }{ \epsilon^2 } )$, with at least probability 0.9, there exist constants $\alpha_i \in \R^i$ and $\beta \in \R$ for which projected SGD after $T$ iterations gives
\begin{align} \label{eq:nn-optimization-cnf}
| \nnit{i}{(T)} - ( \dotp{ \alpha_i }{ \vecx{i} } + \beta ) | \leq O \rb{ \epsilon },
\end{align}
for all $x$ with $\| x \|_2 \leq 1$.
\end{theorem}

Theorem \ref{theorem:smaller-variance-problem-cnf} tells us that if we choose $\eta$ and $T$ as suggested in the theorem statement then the function learned by overparametrized neural networks is close to a linear function. Recall from Sec.~\ref{subsec:supervised-learning-analysis} that choosing similar values of $\eta$ and $T$ in supervised learning enables the provable successful training of the neural network. 
The same issue in approximation arises for \emph{all} activations with continuous derivative. More details about case (1) is given in Appendix \ref{sec:problem-training-cnf}. The result in case (2) is given by the next theorem. 

\begin{theorem}
\label{theorem:larger-variance-cnf-problem}
For any constant $c > 0$ and any $\eta > 0$, $T>1$, if norm of change in parameters 
$\| \tht{T} \|_{1, 2} \leq O ( \inlinefrac{1}{\epsilon_a \sigma_{wb} \tau m^c \log m } ),$
then for all $i \in [d]$ and for all $x$ with $\norm{x}_2 \leq 1$, we have
\begin{align*}
\abs[0]{ \pnilt{\ell}{i}{(T)} } \leq O \big( \tfrac{1}{\sigma_{wb} m^c \sqrt{\log \rb{m d} } } \big).
\end{align*}
\end{theorem}

Most extant theoretical analyses require that 
 the change in weights from initialization is small so that the pseudo-network remains close to the neural network. Small change implies $\abs[0]{\pnilt{\ell}{i}{(T)}} = O( \inlinefrac{1}{m^c} )$ for some constant $c>0$. 
 Therefore, $P_{\ell}(x; \tht{T})$ can not in general approximate the target function (with $P_c( \vecx{i} )$ subtracted). And the same happens with $\nnit{i}{(T)}$ because it is close to $\pnit{i}{(T)}$. 
  More details about case (2) is provided in Appendix \ref{sec:problem-training-cnf}. 

%
We also show the negative effect of overparameterization for CNF in experiments (Section \ref{section:experiments}).  



\section{Unconstrained Normalizing Flow}
\label{sec:unconstrained-normalizing-flow}
In this section, we first describe our UNF model that we analyze and then present our main theoretical result on training and generalization of the UNF model. 

\subsection{Our UNF model}
Unlike the constrained case, where we model $f(x)$ using neural networks, here we model the Jacobian $\inlinefrac{\partial f(x)}{ \partial x }$ using $d$ neural networks by setting $$\frac{\partial \funci{i} }{ \partial x_i }=\phi ( \nni{i} ),$$ where $\phi$ is $\eluone$ function given by 
\begin{align*}
\phi(u) &= e^u \,\ind{u < 0} + \rb{u + 1}\, \ind{u \geq 0} \text{ and } \\ 
\nni{i} &= \sum_{r=1}^m \initam{i}{r} \, \rho ( \dotp{ \initwm{i}{r} + w_{i, r} }{ \nrmxi{i} } + \left( \initbm{i}{r} + b_{i, r} ) \right)
\end{align*}
 with $\rho= \relu$. In the expression for $\nni{i}$ instead of $\vecx{i}$, we use $\nrmxi{i} \in \R^{i+1}$ to aid in analysis; the extra coordinate is added to make $\| \nrmxi{i} \|_2 = 1$. 
No normalization factor is needed in the expression for $\nni{i}$ because of the choice of initialization distribution specified later. 
We can reconstruct $f$ by integration:
\begin{align*}
\funci{1} &= \int_{-1}^{x_1} \frac{ \partial f_1 (t) }{ \partial t } \deriv t \hspace{0.5cm} \text{and} \\
 \funci{i} &= \int_{-1}^{x_i} \frac{ \partial f_i (x_1, x_2, \ldots, x_{i-1}, t) }{ \partial t } \deriv t
\end{align*}
 for  $i \in [d]$. The lower limit in our integral is  $-1$ because $\norm{x}_2 \leq 1$ by our assumption on the support of the data distribution. We also denote $\inlinefrac{ \partial \funci{i} }{ \partial x_i }$ by $\dfi{i}$.
The monotonicity of $f$ is achieved by ensuring that $\dfi{i}$ is positive for all $x$. 
 Although positivity was the only useful property of $\phi$ mentioned by \cite{FrenchPaper}, it turns out to have several other properties which we will exploit in our proof: it is 1-Lipschitz and increasing, its derivative is 1-Lipschitz, and its second derivative is non-negative (except at $0$, where it's not defined).

\paragraph{Quadrature.} To reconstruct $f$, from the Jacobian we need to evaluate the integrals. While this cannot be done exactly, good approximation can be obtained via numerical integration (also known as quadrature).  We estimate $\funci{i}$ via the general quadrature formula by $$ \approxfunci{i} = \sum_{j=1}^{ Q } {q}_{j} \dfqpi{j}{i}.$$
Here, $Q$ is the number of quadrature points and the ${q}_{1}, \ldots, {q}_{Q}$ are the corresponding coefficients. 
We use simple rectangle quadrature, which arises in Riemann integration, and uses only positive coefficients with $q_j = \Delta_{x_i} := \inlinefrac{x_i+1}{Q}$ and 
$\qpi{j}{i} = \rb{ x_1, \ldots, x_{i-1}, -1 + j \Delta_{x_i}}$. 

\cite{FrenchPaper} uses Clenshaw--Curtis
quadrature where the coefficients ${q}_{i}$ can be negative.  Compared to Clenshaw--Curtis quadrature, the rectangle quadrature requires more points for similar accuracy (about doubling the number of quadrature points in our experiments). This is a small price to pay because rectangle quadrature makes the problem of minimizing the loss of the pseudo-network (defined shortly) easier to analyze via the positivity of the quadrature coefficients.

\paragraph{Exponential base distribution.} 
Taking the standard Gaussian as a base distribution as in \cite{FrenchPaper} causes two difficulties: it is not clear that the loss function in the pseudo-network is convex (see Remark \ref{rem:gaussian_convex}). Moreover, it is not clear that throughout training the Lipschitz constant of the loss function will remain bounded by an absolute constant and hence independent of the parameters. (This issue also arises in supervised learning, e.g. \cite{allen2019learning}, though the authors seem to have not realized the problem and do not address it.)
Both of these difficulties with the Gaussian can be circumvented by using the exponential as the base distribution. This does not cause any negative effects in our experiments. 

\paragraph{Learner network parameterization and training procedure.} We initialize $\initam{i}{r} \sim \N (0, \epsilon_a^2)$, $\initw{r} \sim \N \rb{0, \inlinefrac{1}{m}}$ and $\initb{r} \sim \N \rb{ 0, \inlinefrac{1}{m} }$, where $\epsilon_a = O ( \inlinefrac{ \epsilon }{ \log m } )$ is a small constant. 
Additionally, using the estimates $\approxfunci{i}$, we get approximate loss function 
\begin{align*}
\al \rb{ \dfv{f}, x } = \sum_{i=1}^d \approxfunci{i} - \sum_{r=1}^d \log \rb{ \dfi{i} }. 
\end{align*}
 Define average approximated loss as $\al \rb{ \dfv{f}, \mcx} =\tfrac{1}{n} \sum_{x \in \mathcal{X}}\al \rb{ \dfv{f}, x }$  and expected approximated loss ass $\al \rb{ \dfv{f}, \mathcal{D}} = \E_{x \sim \mathcal{D}}\al \rb{ \dfv{f}, x }$ .
 The parameters of neural networks are updated using SGD:
 \begin{align*}  
 \tht{t+1} = \tht{t} - \eta \, \nabla_{\theta} \al (\dfv{f}, \xt{t})
 \end{align*}
  where $\nabla_i f_i = \phi ( N ( x_{1:i}; \tht{t}_i ))$, and $\xt{t} \in \mcx$ is chosen uniformly at random from the training set at each step. We assume that our data is generated from a target function $\tarfunc{*} = \rb{ \tarfunci{1}, \tarfunci{2}, \ldots, \tarfunci{d} }$, where $F^*_i : \R^i \to \R$. Thus, $\tarfunc{*-1} \rb{Z}=X$. 
 
\paragraph{Target function class.} We consider target functions whose derivative are given by
\begin{align*}
 \frac{ \partial \tarfunci{i} }{ \partial x_i } = \phi \rb{ \sum_{r=1}^{p_i} \tftp{i, r}  \psi_{i, r} (  \dotp{ \tfhw{i, r} }{ \nrmxi{i} }  } 
\end{align*}
 where
 $\abs[0]{ \tftp{i, r} } \leq 1$,$\norm[0]{ \tfhw{i, r} }_2 \leq 1$ for all $i \in [d]$ and $\psi_{i, r} : \R \to \R$ are smooth functions with Taylor expansion and $p_i$ are positive integers. Our target function class is rich: the argument of $\phi$ is two-layer neural network with smooth activations. 


\paragraph{Target function complexity.} We need to quantify the complexity of the functions: more complex functions allow representing more distributions but are also harder to learn. We begin by defining the complexity of univariate smooth functions used in the definition of target functions. 
 Let $\psi: \R \to \R$ have Taylor expansion $\psi(y) = \sum_{j=0}^{\infty} c_j y^j$, then, for $\epsilon > 0$, its complexity $C_0( \psi, \epsilon )$ is given by $$C_0( \psi, \epsilon ) = O\rb{ \rb{ \sum_{i=0}^{\infty} (i + 1)^{1.75} |c_i|} \mathrm{poly} \rb{ \frac{1}{\epsilon} } }$$ which is a weighted norm of the Taylor coefficients.  For example, when $\psi(y)$ is one of $\mathrm{poly}(y), \sin(y), e^y - 1, \tanh (y)$, it is known that $C_0( \psi, \epsilon ) = O ( \mathrm{poly} ( \inlinefrac{1}{\epsilon} ) )$ \citep{Arora_Generalization, allen2019learning}. Very roughly, $C_0(\psi, \epsilon)$ captures how many samples are needed to learn $\psi$ up to error $\epsilon$.
For $\tarfunc{*}$ in our target class, complexity $C(\tarfunc{*}, \epsilon)$ is defined to be $\mathrm{poly} ( d, \max_{i \in [d]} p_i, \max_{i \in [d], r\in [p_i] } C_0( \psi_{i, r}, \epsilon ) )$.

\begin{figure*}
\centering
    \begin{tabular}{ccc}
         \includegraphics[width=0.95\columnwidth]{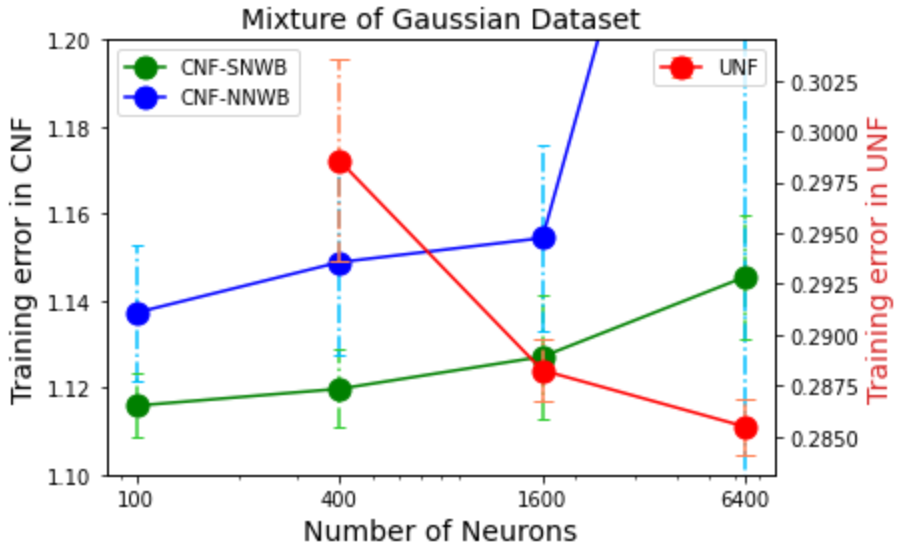} & \includegraphics[width=0.95\columnwidth]{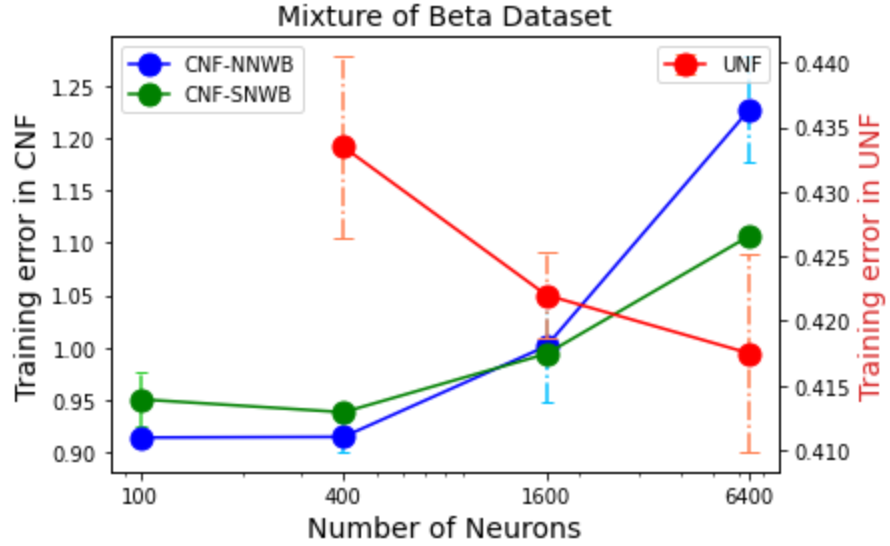} 
    \end{tabular}
\caption{Effect of over-parameterization on training of CNF and UNF on mixture of Gaussian (left figure) and mixture of Beta (right figure) dataset}
\label{fig:flow-mog-mob-width-error}
\end{figure*}

%
%

\subsection{Theoretical analysis of UNF}

We state the main theorem for UNFs informally. (For the complete version, see Theorem~\ref{theorem:final-theorem} in the appendix.)  

\begin{theorem} \label{thm:unf_theorem}  
For any $\epsilon > 0$ and for any target function $\tarfunc{*}$ with finite $\inlinefrac{ \partial F_i^*(x_{1:i}) }{ \partial x_i }$ for all $i \in [d]$, hidden layer size $m \geq \inlinefrac{C (\tarfunc{*}, \epsilon)}{\epsilon^2}$, the number of samples $n \geq \inlinefrac{ { C (\tarfunc{*}, \epsilon)  }}{\epsilon^2}$, the number of quadrature points $Q \geq O( \inlinefrac{C (\tarfunc{*}, \epsilon) }{\epsilon} )$ and total time steps $T \geq O( \inlinefrac{ C (\tarfunc{*}, \epsilon) }{ \epsilon^2 } )$ 
with probability at least $0.9$, we have
\begin{align*}
    \E_{\mathrm{sgd}} \Big[ \tfrac{1}{T} \sum_{t=0}^{T-1}   \E_{ x \sim \mcd }  L (\ft{t}, x)  \Big] - \E_{ x \sim \mcd } \left[ L (\tarfunc{*}, x) \right] = O(\epsilon).
\end{align*}
\end{theorem}

Recall that $\text{KL} ( p_{\tarfunc{*}, Z} || p_{\ft{t}, Z} ) = \E_X \log\inlinefrac{p_{\tarfunc{*}, Z}(X)}{p_{\ft{t}, Z}(X)}$, which gives $\E_{\mathrm{sgd}} [ \inlinefrac{1}{T} \sum_{t=0}^{T-1}   \text{KL} ( p_{\tarfunc{*}, Z} || p_{\ft{t}, Z} ) ] = O(\epsilon)$. 
Using Pinsker's inequality, we can also bound the total variation distance between the learned and data distributions $p_{f_t, Z}$ and $p_{\tarfunc{*}, Z}$.  
The theorem can be interpreted as saying that the target density $p_{\tarfunc{*}, Z}$ of $X = F^{*-1}(Z)$ is close to the density given by the learned function, namely $p_{\ft{t}, Z}$ (which is the density of $(\ft{t})^{-1}(Z)$). Note that Theorem \ref{thm:unf_theorem} gives the learning guarantee for all probability distributions which has a two-layer low complexity neural network with smooth activation as the derivative of the target function $\tarfunc{*}$. An example of such functions is any positive low degree polynomial with small coefficients.  

\paragraph{Proof Outline.} The general outline of the proof follows that for supervised learning mentioned earlier,
but details differ substantially and require new ideas. First, unlike prior work which only works with one neural network, NFs have $d$ neural networks which are trained jointly. But we show that each neural network behaves essentially independently which allows us to analyze each neural network separately. 
Therefore, for each neural network $\dfi{i}$, we define its pseudo-network by 
\begin{align*}
 \dgi{i} = \frac{\partial g_i \rb{ \vecx{i} } }{ \partial x_i } = \phi( \pni{i} ).
 \end{align*}
Note that our definition of pseudo-network is not a straightforward generalization from the supervised case: $\dgi{i}$ is not a linear approximation of $\dfi{i}$ because we are not taking linear approximation of final activation $\phi$. 
For every $i \in [d]$, we show the existence of pseudo-networks close to the target function 
\begin{align*}
  \frac{ \partial \tarfunci{i} }{ \partial \vecx{i} }  \approx \phi \rb{ \pnit{i}{*} }
\end{align*}
for some parameters $\theta_i^*$ and for all $x$ (Lemma~\ref{lemma:existence-pseudo-network}).
  However, for this we cannot directly use prior work: since our pseudo-network approximation is used in quadrature, it needs to be pointwise (close in $L_\infty$) unlike only on average (close in $L_1$) as in the prior work.
Next, we show that for each $i \in [d]$, the corresponding neural network and pseudo-network remain close during optimization and the same holds for the gradients of their respective loss functions (Section \ref{sec:coupling} on coupling). Specifically, for all $i \in [d]$, all $t \in [T]$ and all $x$, we show that 
\begin{align*}
 \dfit{i}{(t)} &\approx \dgit{i}{(t)} \hspace{10mm} \text{(Lemma \ref{lemma:coupling-df-dg})} \\
\nabla_{\theta_i} \rb{ \dfit{i}{(t)} } &\approx \nabla_{\theta_i} \rb{ \dgit{i}{(t)} } \text{(Lemma \ref{lemma:coupling-delta-f-delta-g})} \\
\al \rb{ \nabla \ft{t} , x } &\approx \al \rb{ \nabla \gt{t} , x } \hspace{9mm}  \text{(Lemma \ref{lemma:coupling-loss})} \\
\nabla_{\theta} \alm{\ft{t}}{ x } &\approx \nabla_{\theta} \alm{ \gt{t}}{ x } \hspace{4mm}  \text{(Lemma \ref{lemma:coupling-gradient-loss})}.
\end{align*}
 Using coupling and independence of neural networks mentioned above, we show that SGD achieves near-minimum training loss  (Theorem \ref{theorem:optimization-empirical-approximated-L}), that is, for sufficient large $T$,  
 \begin{align*}
     \frac{1}{T} \sum_{t=0}^{T-1} \E_{\mathrm{sgd}} & [ \al( \dfv{ \ft{t} } , \mcx) ] \leq \al( \dfv{ \tarfunc{*} } , \mcx) + O( \epsilon ).
 \end{align*}
Compared to the supervised setting the details in these sections are considerably more involved due to the presence of $\dfv{f}$ and $\tilde{f}$ and other features of the loss function. Finally, the full generalization result is proven in Theorem \ref{theorem:final-theorem} showing that for sufficiently large $T$, population loss $ L (\ft{T}, \mathcal{D})$ is close to $L (\tarfunc{*}, \mathcal{D})$: 
\begin{align*}
     \frac{1}{T} \sum_{t=0}^{T-1} \E_{\mathrm{sgd}} \left[ L (\ft{t}, \mcd) \right] \leq L (\tarfunc{*}, \mcd) + O(\epsilon).
\end{align*}
This is proven by stringing together several approximate equalities. First, we show that the loss $\al (\dfv{ \tarfunc{*} }, x)$ (and $\al ( \dfv{ \ft{t} } , x)$) using the approximation via quadrature is close to the true loss $L (\tarfunc{*}, x)$ (respectively $ L (\ft{t},  x)$):
\begin{align*}
\al (\dfv{ \tarfunc{*} }, x) \approx L (\tarfunc{*}, x) \;\; \text{  and  } \;\;
\al ( \dfv{ \ft{t} } , x) \approx L (\ft{t},  x)
\end{align*}
It is also shown that
the empirical and population versions of approximate loss are close: 
\begin{align*}
\al ( \dfv{\ft{t}} , \mathcal{X}) &\approx \al ( \dfv{\ft{t}} , \mathcal{D}) \hspace{10mm} \text{(Lemma \ref{lemma:generalization-neural-net-func})} \\ 
\al (\dfv{ \tarfunc{*} }, \mathcal{X}) &\approx \al (\dfv{ \tarfunc{*} }, \mathcal{D}) \hspace{11mm} \text{(Lemma \ref{lemma:generalization-target-func})}.
\end{align*} 
These results together with the optimization result mentioned earlier give Theorem \ref{theorem:final-theorem}.

\section{Experiments}
\label{section:experiments}

In Sec.~\ref{section:constrained-normalizing-flow-main}, we theoretically show that overparameterized neural networks in CNFs can not approximate the target function in the bounded time steps or in the bounded change in weights, and in Sec.~\ref{sec:unconstrained-normalizing-flow}, we show that highly overparameterized neural networks provably learn target distribution. We now give empirical evidence of these claims. In Fig.~\ref{fig:flow-mog-mob-width-error}, we plot training error after a fixed number of training iterations for a different amount of over-parameterization for both CNF and UNF models on a mixture-of-Gaussian and a mixture-of-Beta distribution datasets. The left and right $y$-axes represent training error in CNF and UNF models, respectively. CNF-SNWB and CNF-NNWB denote CNF models with standard normal and normalized normal $(\N \rb{0, \inlinefrac{1}{m}})$ initialization of parameters, resp. We see that as we increase overparameterization in CNF models, training error becomes \textit{larger} after a fixed number of training iterations, which means that larger CNF models need \textit{larger} number of training iterations to learn the target function. But in UNFs, by increasing overparameterization, training error becomes \textit{smaller}, which means that larger UNF models need \textit{smaller} number of training iterations to learn the target function. Thus, our experimental results suggest that overparameterization in CNFs makes training \textit{slower} and overparameterization in UNFs makes training \textit{faster}.
These experiments were done for a fixed learning rate. Similar patterns were observed for various different settings of learning rates except when training becomes unstable in CNFs.
 Since results in supervised learning also suggest that overparameterization makes training faster \cite{neyshabur2015search}, our results on CNF are novel and surprising. Results on CNFs as well as results on UNFs on additional synthetic and real datasets, deeper models, various initializations, different learning rates and full experimental setup are given in Appendix \ref{sec:additional-experiments}.


\section{Conclusions and Limitaions} \label{sec:conclusion}
We gave the first end-to-end theoretical analysis of normalizing flows. We introduced the dichotomy between CNFs and UNFs: overparametrization seems to be hurting training of CNFs but for UNFs overparametrization does not hurt and we can analyze UNFs when the underlying network has one hidden-layer. We also proposed NF variants with desirable properties and these may find use in future work.

The main limitations of our work are the following which also suggests the main open problems: (1) A clear theoretical and empirical understanding of the role of overparameterization in CNFs remains an interesting open direction. As shown by our negative theoretical results, it seems necessary to analyze CNFs in the moderately overparametrized setting. However, this setting is not well-understood even in the supervised case. 
(2) For UNFs our analysis requires the overparametrized setting. (3) For the analysis we distill NF architectures to essentials---while this permits us to zero in on the main phenomena the more practical architectures are far more elaborate and performant and pose new theoretical challenges. (4) Our work assumes the autoregressive structure of the flow models. However, the role of overparameterized neural networks in other normalizing flow models such as coupling flows, residual flows, and other generative models such as VAEs is not well understood. (5) Our theoretical results have a one-hidden layer flow model but invertible flow models can be sequentially composed to construct an invertible map and in practice, flows models are sequentially composed to learn flexible target distributions. Extending our theoretical results for such models is an open problem.


\bibliography{references}


\clearpage
\appendix

\thispagestyle{empty}

\onecolumn \makesupplementtitle

\section{Outline}
\label{sec:outline-appendix}
In this section, we give outline of details and proofs of supplementary. We define common notations between Constrained Normalizing Flows results and Unconstrained Normalizing Flow results in Appendix \ref{sec:notation}. Our results on CNFs from Section \ref{section:constrained-normalizing-flow-main} from the main paper are discussed in detail in Theorem \ref{theorem:smaller-variance-problem-cnf-appendix} (Section \ref{subsec:smaller-variance-cnf-problem}) and Theorem \ref{theorem:larger-variance-cnf-problem-appendix} (Section \ref{subsec:large-variance-cnf-problem}) and their proofs. 

We give details about our result on UNFs (in Section \ref{sec:unconstrained-normalizing-flow}) in Theorem \ref{theorem:final-theorem} and its proof (Section \ref{sec:generalization}). Our analysis begins with showing that if change in weights and biases from the initialization is small for a neural network, then training dynamics of the pseudo-network (linear approximation of neural network) is close to training dynamics of the neural network in Section \ref{sec:coupling}. In Section \ref{sec:existence}, we show that with high probability there exist a pseudo-network which can approximate the derivative of target function. 
In Section \ref{sec:optimization}, we show that optimization problem for the pseudo-network is convex; therefore, combining results from Section \ref{sec:existence} and Section \ref{sec:coupling} will give us the result that the loss of UNFs on the training data is close to the 	loss of target function. In section \ref{sec:generalization}, we prove generalization guarantees to test datasets and complete the proof of Theorem \ref{theorem:smaller-variance-problem-cnf-appendix}.

We also provide experimental results to verify our theoretical claims on UNFs and CNFs in Section \ref{section:experiments} and Section \ref{sec:additional-experiments}. Discussion of related work is given in Section \ref{sec:related-work}.

\section{Notations}
\label{sec:notation}
In this section, we define commonly used notations. We denote $(\pmb{\alpha}, \pmb{\beta})$ as a concatenation of 2 vectors $\pmb{\alpha}$ and $\pmb{\beta}$. For any 2 vectors $\pmb{\alpha}$ and $\pmb{\beta}$, $\pmb{\alpha} \odot \pmb{\beta}$ denotes element wise multiplication of $\pmb{\alpha}$ and $\pmb{\beta}$ vector. We use $\| \pmb{\alpha} \|_1$, $\| \pmb{\alpha} \|_2$ and $\| \pmb{\alpha} \|_{\infty}$ to denote $L_{1}$, $L_{2}$ and $L_{\infty}$ norm of vector $\pmb{\alpha}$. For any matrix $M \in \R^{m \times d}$, we denote matrix norm as
\begin{align*}
\norm{M}_{p, q} = \rb{ \sum_{ i \in [m]} \norm{ m_i }_p^q }^{1/q},
\end{align*}
where $m_i \in \R^d$ denotes row vector of matrix $M$.
We denote vector $\onevector = \rb{1, 1, \ldots, 1} \in \R^m$. Big-$O$ and Big-$\Omega$ notation to hide only constants. We use $\log$ to denote natural logarithm. For any constant $n$, $[n]$ is denoted by set $\{1,2,\ldots,n\}$. We use $\N ( \mu, \sigma )$ to denote Gaussian distribution with mean $\mu$ and variance $\sigma$. We use $\ind{E}$ to denote the indicator of the event $E$. We say a function $f: \R^d \to \R$ is $L$-Lipschitz continuous if $\abs{ f(x) - f(y) } \leq  L \norm{x - y}_2$ for all $x, y \in \R^d$.

\section{Preliminaries}

Recall that $X$ is the random variable corresponding to the data distribution and $Z$ is a random variable with standard Gaussian or multivariate exponential distribution. There seems to be no well-accepted definition of standard exponential distribution; for our purposes the following natural defintion will serve well.
The density of the standard exponential distribution at $z = \rb{ z_1, z_2, \ldots, z_d } \in \R^d$ is given by $e^{- \sum_{i=1}^d z_i}$ when all $z_i \geq 0$, and by $0$, otherwise. Let flow $f: \R^d \to \R^d$ be an monotonic autoregressive function. Then standard change of density formula using invertibility of $f$ gives
\begin{align*}
p_{f, Z}(\xx) = p_{Z}(\zz) \detm{ \frac{ \partial f(\xx) }{ \partial \xx } }.
\end{align*}
To make $f(x)=( \funci{1}, \funci{2}, \ldots, \funci{d} )$ an monotonic autoregressive function, we force function $\funci{i}$ to be monotonic with respect to $x_i$ for any fixed $\vecx{(i-1)}$ where $x_i$ is $i^{th}$ dimension of $x$. Recall that $\vecx{i}$ represents the vector including first $i$ elements of vector $x$ for any $i \in [1, d]$.

Unlike the constrained case where we model $f$ using a neural network, in unconstrained case we model derivative of function using $d$ neural networks. In  normalizing flow, for all $i \in [1, d]$, we model $ \frac{ \partial \funci{i} }{ \partial x_i } $ using a neural network $\nni{i}$. To be specific, 
\begin{align*}
 \dfi{i} = \frac{ \partial \funci{i} }{ \partial x_i } = \phi \rb{ \nni{i} }.
\end{align*}
We denote $\dfv{f}$ as $\rb{ \dfi{1}, \ldots, \dfi{r}, \ldots, \dfi{d}}$.  Here, $\phi$ is the \elu+1 function given by $\phi(x)=e^x \ind{ x \leq 0 } + \rb{x+1} \ind{x > 0}$ for all $x \in \R$. we use a one-hidden-layer neural network in $\nni{i} $, which is given by
\begin{align*}
\nni{i} = \sum_{r=1}^m \initam{i}{r} \sigma \rb{  \dotp{ \initwm{i}{r} + w_{i,r}}{\nrmxi{i} }  + \rb{ \initbm{i}{r} + b_{i,r} }  }
\end{align*}
We construct $\nrmxi{i} \in \R^{i+1}=(x_1, x_2, \ldots, x_i, \sqrt{1 - \| \vecx{i} \|^2 } )$ such that $ \| \nrmxi{i} \|_2 = 1$. We can reconstruct $f$ by integration:
\begin{align*}
\funci{1} = \int_{-1}^{x_1} \inlinefrac{ \partial f_1 (t) }{ \partial t } \deriv t \hspace{0.4cm} \text{and} \hspace{0.4cm} \funci{i} = \int_{-1}^{x_i} \inlinefrac{ \partial f_i (x_1, x_2, \ldots, x_{i-1}, t) }{ \partial t } \deriv t \text{ for }  1 < i \leq d. 
\end{align*}
The lower limit in our integral is  $-1$ because $\norm{x}_2 \leq 1$ by our assumption on the support of the data distribution. Note that to reconstruct $f$ from the Jacobian, we need to evaluate the integrals. While this cannot be done exactly, good approximation can be obtained via numerical integration (also known as quadrature).  We estimate $\funci{i}$ via the general quadrature formula by 
\[ 
\label{eq:hhello}
\approxfunci{i} = \sum_{j=1}^{ Q } {q}_{j} \dfqpi{j}{i}. 
\]
Here, $Q$ is the number of quadrature points and the ${q}_{1}, \ldots, {q}_{Q}$ are the corresponding coefficients. 
We use simple rectangle quadrature, which arises in Riemann integration, and uses only positive coefficients with $q_j = \Delta_{x_i} := \inlinefrac{x_i+1}{Q}$ and 
$\qpi{j}{i} = \rb{ x_1, \ldots, x_{i-1}, -1 + j \Delta_{x_i}}$. 
The loss function for normalizing flows is given by
\begin{align*}
\al \rb{ \dfv{f}, x } = -\log \rb{ p_Z \rb{ \approxfunc } } - \log \rb{ \Pi_{i=1}^d \dfi{i} }
\end{align*}
Using standard exponential distribution as a base distribution, we get
\begin{align}
\label{eq:divide-L-in-Li}
\al \rb{ \dfv{f}, x } = \sum_{i=1}^d \approxfunci{i} - \sum_{r=1}^d \log \rb{ \dfi{i} } = \sum_{i=1}^d \al_i \rb{ \dfv{f}, x }
\end{align}
where 
\begin{align*}
\al_i \rb{ \dfv{f}, x } = \approxfunci{i} - \log \rb{ \dfi{i} }.
\end{align*}
 For our theoretical result, we consider target functions whose derivative are given by
\[ 
\frac{ \partial \tarfunci{i} }{ \partial x_i } = \phi ( \sum_{r=1}^{p_i} \tftp{i, r}  \psi_{i, r} (  \dotp{ \tfhw{i, r} }{ \nrmxi{i} }  ) ),
\] 
where $\abs[0]{ \tftp{i, r} } \leq 1$,$\norm[0]{ \tfhw{i, r} }_2 \leq 1$ for all $i \in [d]$ and $\psi_{i, r} : \R \to \R$ are smooth functions with Taylor expansion and $p_i$ are positive integers. Our target function class is rich: the argument of $\phi$ is two-layer neural network with smooth activations. 
 
 We need to quantify the complexity of the functions: more complex functions allow representing more distributions but are also harder to learn. We begin by defining the complexity of univariate smooth functions used in the definition of target functions. 
 Let $\psi: \R \to \R$ have Taylor expansion $\psi(y) = \sum_{j=0}^{\infty} c_j y^j$, then its complexity $C_0( \psi, \epsilon )$ for $\epsilon > 0$ is given by $O( (\sum_{i=0}^{\infty} (i + 1)^{1.75} |c_i|) \mathrm{poly} ( \inlinefrac{1}{\epsilon} ) )$ which is a weighted norm of the Taylor coefficients.  For example, when $\psi(y)$ is one of $\mathrm{poly}(y), \sin(y), e^y - 1, \tanh (y)$, it is known that $C_0( \psi, \epsilon ) = O ( \mathrm{poly} ( \inlinefrac{1}{\epsilon} ) )$ \cite{allen2019learning}. Very roughly, $C_0(\psi, \epsilon)$ captures how many samples are needed to learn $\psi$ up to error $\epsilon$.
For $\tarfunc{*}$ in our target class, complexity $C(\tarfunc{*}, \epsilon)$ is defined to be $\mathrm{poly} ( d, \max_{i \in [d]} p_i, \max_{i \in [d], r\in [p_i] } C_0( \psi_{i, r}, \epsilon ) )$.

For each neural network $\dfi{i}$, we define its pseudo-network by 
    $\dgi{i} = \inlinefrac{\partial g_i \rb{ \vecx{i} } }{ \partial x_i } = \phi( \pni{i} ).$, 
where 
\[
\pni{i} = \sum_{r=1}^m \initam{i}{r} \sigma \rb{ \neuralvalzerom{i} }
\]
Note that our definition of pseudo-network is not the straightforward generalization from the supervised case: $\dgi{i}$ is not a linear approximation of $\dfi{i}$ because we are not taking linear approximation of final activation $\phi$. 

\section{Coupling}
\label{sec:coupling}
In this section, we will establish closeness between training dynamics of neural networks and pseudo network, which we will call as coupling. First, we will establish the coupling between $\dfi{i}$ and $\dgi{i}$ (Lemma \ref{lemma:coupling-df-dg}). Using coupling between $\dfi{i}$ and $\dgi{i}$, we prove coupling between $\al_i \rb{ \nabla \ft{t} , x }$ and $\al_i \rb{ \nabla \gt{t} , x }$ (Lemma \ref{lemma:coupling-loss}). We also prove coupling between gradient $\nabla_{\theta} \alm{\ft{t}}{ x }$ and $\nabla_{\theta} \alm{ \gt{t}}{ x }$ in Lemma \ref{lemma:coupling-gradient-loss}, which will be used in proving global optimization of neural network in Section \ref{sec:optimization}.

We define $\lambda_1$ as \todo{explain why $\lambda_1$ is being defined, why it is relevant to the proof}
\begin{equation}
\label{eq:define-tilde-L1-tilde-L2}
\begin{aligned}
    \lambda_1 &= \sup_{t \in [T], i \in [d], r \in [m], \wt{i, r}{t}, \bt{i ,r}{t}, \abs{x} \leq 1 } \frac{ \phi' \rb{ \nnit{i}{(t)} } }{ \phi \rb{ \nnit{i}{(t)} } },
\end{aligned}
\end{equation}
which will be used later in the proof of coupling between $\dfi{i}$ and $\dgi{i}$. The upper bound on $\lambda_1$ is useful to bound derivative of $\alm{f}{x}$ w.r.t. $w_{i, r}$. We get the following upper bound on $\lambda_1$:
\begin{align}
\label{eq:tilde-L2-value}
    \nonumber \lambda_1 &= \sup_{t \in [T], i \in [d], r \in [m], \wt{i, r}{t}, \bt{i, r}{t}, \abs{x} \leq 1 } \frac{ \phi' \rb{ \nnit{i}{(t)} } }{ \phi \rb{ \nnit{i}{(t)} } } \\
    \nonumber &= \sup_{t \in [T], i \in [d], r \in [m], \wt{i, r}{t}, \bt{i, r}{t}, \abs{x} \leq 1 } \frac{ \exp \rb{ \nnit{i}{(t)} } \ind{ \nnit{i}{(t)} < 0 } +  \ind{ \nnit{i}{(t)} \geq 0 } }{ \exp \rb{ \nnit{i}{(t)} } \ind{ \nnit{i}{(t)} < 0 } + \rb{ \nnit{i}{(t)} + 1 } \ind{ \nnit{i}{(t)} \geq 0 } } \\
    \nonumber &= \sup_{t \in [T], i \in [d], r \in [m], \wt{i, r}{t}, \bt{i, r}{t}, \abs{x} \leq 1 } \ind{ \nnit{i}{(t)} < 0} + \frac{ \ind{ \nnit{i}{(t)} \geq 0} }{ \nnit{i}{(t)} + 1 } \\
    &\leq 1.
\end{align}

\todo{Is $\LB$ a good notation? }
Define $\LB$ as
\begin{align}
\label{eq:define-delta-bar}
\LB := 6 c_1 \epsilon_a \sqrt{2 \log m} 
\end{align}
for any fixed constant $c_1 > 10$. \todo{Need to be clear about the meaning of "some". If $c_1$ can be chosen to be any constant $>1$, then we should say "for any positive constant $c_1 > 1$. This comment applies to other usages of "some constant" in the paper.}

Recall that loss function in case of CNFs is given by
\begin{align*}
\alm{f}{x} &= \sum_{i=1}^d \approxfunci{i} - \sum_{i=1}^d  \log \rb{ \dfi{i} }, \\
&= \sum_{i=1}^d \rb{ \sum_{j=1}^Q \Delta_x \dfqpit{j}{i}{(t)} } - \sum_{i=1}^d  \log \rb{ \dfi{i} }, \\
&= \sum_{i=1}^d \rb{ \sum_{j=1}^Q \Delta_x \phi \rb{ N \rb{ \qpi{j}{i}, \tht{t}_i } } } - \sum_{i=1}^d  \log \rb{ \phi \rb{ N \rb{ \vecx{i}, \tht{t}_i } } }, \\
\end{align*}

\begin{lemma} (Bound on change in weights) \label{lemma:bound-wt-bt} For every $i \in [d]$, for all $r \in [m]$, for any positive constant $c_1 \geq 10$ and for every $\vecx{i}$ with $\norm{ \vecx{i} }_2 \leq \frac{1}{2}$, with at least $1  - \frac{1}{c_1}$ probability over random initialization, bound on change in weights after $t$ steps with learning rate $\eta$ is given by
\begin{align*}
        \norm{ \wt{i, r}{t} }_2 & \leq \eta \LB t, \\
    \abs{ \bt{i, r}{t} } &\leq \eta \LB t.
\end{align*}
\end{lemma}
\begin{proof}
By taking derivative of $\alm{f}{x}$ w.r.t. $w_{i, r}$,  we get \todo{The reader has to lot of work to understand the equation below. Before the lemma, we need to recall and explicitly write the expression for $\al$ which is then used below. In the main paper, we only gave the abstract formula for $\al$ and not the one obtained by substituting for $f$ its spcific choice.}
\begin{align*}
    \norm{ \frac{ \partial \alm{\ft{t}}{x} }{ \partial w_{i, r} } }_2 \leq&~ \norm{ \rb{ \sum_{j=1}^Q \Delta_x \phi' \rb{ N \rb{ \qpi{j}{i}, \tht{t}_i } }  \, \initam{i}{r} \sigma' \rb{ \neurvalm{ i }{(t)}{\nrmqpi{j}{i}} } \nrmqpi{j}{i} } }_2 \\
    &+ \norm{ \frac{1}{ \phi \rb{ \nnit{i}{(t)} } } \rb{ \phi' \rb{ \nnit{i}{(t)} } \,\initam{i}{r} \sigma' \rb{ \neurvalm{i}{(t)}{ \nrmxi{i} } } \nrmxi{i}  } }_2 \\
    \leq & \sum_{j=1}^{ Q } \norm{ \Delta_x \phi' \rb{ N \rb{ \qpi{j}{i} ; \tht{t}_i} } \, \initam{i}{r} \sigma' \rb{ \neurvalm{i}{(t)}{ \nrmqpi{j}{i} }  } \nrmqpi{j}{i} }_2  \\
    &+ \abs{ \frac{ \phi' \rb{ N \rb{ \vecx{i} ; \tht{t}_i } } }{ \phi \rb{ N \rb{ \vecx{i} ; \tht{t}_i } } } } \norm{  \initam{i}{r} \sigma' \rb{ \neurvalm{i}{(t)}{\nrmxi{i}}  } \nrmxi{i}  }_2. \\
\end{align*}
Using $\abs{q_j} \leq \frac{2}{Q} $, $\| \nrmqpi{j}{i} \| = 1$, $\| \nrmxi{i} \| = 1$ and $\abs[0]{ \phi'\rb{ N( \vecx{i} ; \tht{t}_i ) } /\phi\rb{ N( \vecx{i} ; \tht{t}_i ) }} \leq 1$ (by \eqref{eq:tilde-L2-value}), we get
\begin{align*}
    \norm{ \frac{ \partial \alm{\ft{t}}{x} }{ \partial w_{i, r} } }_2 \leq 3 \abs{ \initam{i}{r} }.
\end{align*}
Using Lemma \ref{lemma:concentration-folded-normal}, with probability at least $1-\frac{1}{c_1}$ we get
\begin{align}
\label{eq:bound-der-loss-wrt-w}
     \norm{ \frac{ \partial \alm{\ft{t}}{ x } }{ \partial w_{i, r} } }_2 \leq \LB
\end{align}
where $\LB$ is defined in \eqref{eq:define-delta-bar}.
Using the same reasoning for $b_{i, r}$, with probability at least $1-\frac{1}{c_1}$ we get
\begin{align}
\label{eq:bound-der-loss-wrt-b}
    \nonumber \abs{ \frac{ \partial \alm{\ft{t}}{ x} }{ \partial b_{i, r} } } =& \abs{  \sum_{j=1}^{ Q } \Delta_x \phi'\rb{ N \rb{ \qpi{j}{i} ; \tht{t}_i } }  \initam{i}{r} \sigma' \rb{ \neurvalm{i}{(t)}{\nrmqpi{j}{i}} } } \\
    \nonumber &+ \abs{ \frac{1}{ \phi \rb{ \nnit{i}{(t)} } } \rb{ \phi' \rb{\nnit{i}{(t)} } \initam{i}{r} \sigma' \rb{ \neurvalm{i}{(t)}{ \nrmxi{i} } }  } } \\
    \nonumber  \leq & \; 3 \abs{ \initam{i}{r} }. \\
    \leq& \LB.
\end{align}
Using \eqref{eq:bound-der-loss-wrt-w}, \eqref{eq:bound-der-loss-wrt-b} and the fact that we are using SGD, we obtain
\begin{equation}
\label{eq:upper-bound-on-wt-bt}
\begin{aligned}
    \norm{ \wt{i, r}{t} }_2 & \leq \eta \LB t, \\
    \abs{ \bt{i, r}{t} } &\leq \eta \LB t.
\end{aligned}
\end{equation}
\end{proof}

\begin{lemma}
\label{lemma:bound-change-in-act-patterns}
(Bound on the number of changes in activation patterns) For every $i \in [d]$ and for all $r \in [m]$, suppose $\norm{ w_{i, r} }_2 \leq \Delta_{i}$ and $\abs{ b_{i, r}} \leq \Delta_i $. Then, for every $\vecx{i}$ such that $\norm{ \vecx{i} } \leq \frac{1}{2}$, with probability at least $1  - \exp \rb{ - \frac{ 32 (c_4 - 1)^2  m^2 \Delta_i^2 }{ \pi }  }$ over random initialization, 
the number of activation patterns that change is at most $ c_4 \frac{ 4 \Delta_i \sqrt{m} }{\sqrt{ \pi }}$. In other words, for at most 
 $ c_4 \frac{ 4 \Delta_i \sqrt{m} }{\sqrt{ \pi }}$ fraction of $r \in [m]$, we have
\begin{align*}
    \ind{ \neurvalm{i}{}{\nrmxi{i}} \geq 0 } \neq \ind{\neuralvalzerom{i} \geq 0 }
\end{align*}
for any positive constant $c_4 \geq 1$.
\end{lemma}

\begin{proof} Define \todo{Explain in words what this definition means.}
\begin{align}
\label{eq:define-mathcal-h}
\mathcal{H}_i := \{ r \in [m] \,\mid\, \abs{ \neuralvalzerom{i} } \geq 4 \Delta_i \} .   
\end{align}
The set $\mathcal{H}_i$ contains indices of neurons for which indicator function doesn't change its value if change in weights is bounded by $\Delta_i$. For every $\vecx{i}$ such that $\norm{ \vecx{i} }_2 \leq 1$ and for all $r \in [m]$, $\abs{ \neuralvalzerom{i} } \leq 2 \Delta_i $. For all $r \in \mathcal{H}_i$, we have 
\begin{align}\label{eqn:Htproperty}
\ind{ \neurvalm{i}{(t)}{ \nrmxi{i} } \geq 0 } = \ind{\neuralvalzerom{i} \geq 0 }.
\end{align}
Now, we need to bound the size of $\mathcal{H}_i$. We know that for all $x$ with $\norm{ \vecx{i} }_2 \leq 1$, $\neuralvalzerom{i}$ is Gaussian with $\E \left[ \neuralvalzerom{i} \right] = 0$ and $\text{Var} \left[ \neuralvalzerom{i} \right] = \frac{2}{m}$. Using Lemma \ref{lemma:anti-concentration-zero-mean-normal}, we get
\begin{align*}
    \text{Pr} \rb{ \abs{ \neuralvalzerom{i} } \leq 4 \Delta_i } &\leq  \frac{ 4 \Delta_i \sqrt{m} }{\sqrt{ \pi }}. 
\end{align*}
Using Fact \ref{fact-hoeffding-on-binomial} (Hoeffding's inequality) for $\mathcal{H}_i$ (where $\overline{ \mathcal{H} }_i = [m] \backslash \mathcal{H}_i )$ for any positive constant $c_4 \geq 1$, we get
\begin{align*}
    \text{Pr} \rb{ \abs{ \overline{ \mathcal{H} }_i } \geq c_4 m \frac{ 4 \Delta_i \sqrt{m} }{\sqrt{ \pi }} } &\leq  \exp \rb{ -2 m \rb{ (c_4 - 1) \rb{ \frac{ 4 \Delta_i \sqrt{m} }{\sqrt{ \pi }} } }^2 }, \\
    &\leq  \exp \rb{ - \frac{ 32 (c_4 - 1)^2  m^2 \Delta_i^2 }{ \pi }  },
\end{align*}
which gives
\begin{align*}
    \text{Pr} \rb{ \abs{\mathcal{H}_i} \geq  m \rb{1 - c_4 \frac{ 4 \Delta_i \sqrt{m} }{\sqrt{ \pi }} } } &\geq 1 - \exp \rb{ - \frac{ 32 (c_4 - 1)^2  m^2 \Delta_i^2 }{ \pi }  }.
\end{align*}
\end{proof}

\begin{lemma} \label{lemma:coupling-f-prime-and-g-prime} (Bound on the difference between $ \dfit{i}{(t)} $ and $ \dgit{i}{(t)} $) For every $i \in [d]$, for all $x$ with $\norm{x}_2 \leq \frac{1}{2}$ and for every time step $t\geq1$, with probability at least $1 - \frac{1}{c_1}$ over random initialization, for any positive constants $c_1 > 10$, we have
\begin{align*}
    \abs{ \phi \rb{ \nnit{i}{(t)} } - \phi \rb{ \pnit{i}{(t)} } }  \leq  24 c_1 \epsilon_a \Delta_i \abs{ \overline{ \mathcal{H} }_i^{(t)} } \sqrt{ 2 \log m }.
\end{align*}
    
\end{lemma}

\begin{proof}
Using 1-Lipschitz continuity of $\phi$, we get
\begin{align*}
    \abs{ \phi \rb{ \nnit{i}{} } - \phi \rb{ \pnit{i}{} } } \leq &\; \abs{ \nnit{i}{} - \pnit{i}{} }. 
\end{align*}
We bound $\abs{ \nnit{i}{} - \pnit{i}{} }$:
\begin{align}
\label{eq:coupling-N-P}
    \nonumber \abs{ \nnit{i}{} - \pnit{i}{} } \leq & \Bigg| \sum_{ r \in [m] } \initam{i}{r} \rb{ \neurvalm{i}{}{ \nrmxi{i} } } \ind{ \neurvalm{i}{}{ \nrmxi{i} } \geq 0 } \\ 
    \nonumber &- \sum_{ r \in [m] }  \initam{i}{r} \rb{ \neurvalm{i}{}{\nrmxi{i}} } \ind{ \neuralvalzerom{i} \geq 0 } \Bigg| \\
    \nonumber \leq & \Bigg| \sum_{ r \in \overline{\mathcal{H}}_i } \initam{i}{r} \rb{ \neurvalm{i}{(t)}{\nrmxi{i}} } \Big( \ind{ \neurvalm{i}{}{ \nrmxi{i} } \geq 0 } \\ 
    \nonumber &- \ind{ \neuralvalzerom{i} \geq 0 } \Big) \Bigg| \\
    \nonumber \stackrel{ (\RomanNumLow{1}) }{\leq} &  \abs{ \overline{\mathcal{H}}_i^{(t)} }  \rb{ 2 c_1 \epsilon_a \sqrt{ 2 \log m } } \rb{  4 \Delta_i  + 2 \Delta_i } \rb{2} \\
    \leq & 24 c_1 \epsilon_a \Delta_i \abs{ \overline{\mathcal{H}}_i^{(t)} } \sqrt{ 2 \log m },
\end{align}
where inequality ($\RomanNumLow{1}$) uses Lemma \ref{lemma:concentration-folded-normal} to upper bound $|\initam{i}{r}|$ with probability at least $1 - \frac{1}{c_1}$.
\end{proof}

\begin{lemma}
\label{lemma:coupling-df-dg}
    (Final bound on the difference between $ \dfit{i}{(t)} $ and $ \dgit{i}{(t)} $) For every $i \in [d]$, for all $x$ with $\norm{x}_2 \leq 1$ and for every time step $t\geq1$, with probability at least $1 - \frac{1}{c_1} - \exp \rb{ - \frac{ 32 (c_4 - 1)^2 \eta^2 m^2 \LB^2 t^2 }{ \pi }  }$ over the random initialization, and some positive constants $c_1 > 10$ and $c_4 \geq 1$, we have
    \begin{align}\label{eqn:dnpt}
        \abs{ \phi \rb{ \nnit{i}{(t)} } - \phi \rb{ \pnit{i}{(t)} } }  \leq \abs{ \nnit{i}{(t)} - \pnit{i}{(t)} } \leq \frac{ 192 \eta^2 m^{1.5} \LB^2 c_1 c_4 \epsilon_a t^2 \sqrt{ \log m } }{ \sqrt{\pi } }.
    \end{align}
\end{lemma}

\todo{$\dnp{t}$ is presently hidden. Needs to brought out more prominently.}
\begin{proof}
Using Lemma \ref{lemma:bound-change-in-act-patterns} and Lemma \ref{lemma:coupling-f-prime-and-g-prime}, we get
\begin{align}
    \nonumber \abs{ \phi \rb{ \nnit{i}{(t)} } - \phi \rb{ \pnit{i}{(t)} } } \leq & 24 c_1 \epsilon_a \Delta_i \abs{ \mHB{t}_i } \sqrt{ 2 \log m } \\
    \nonumber \stackrel{ (\RomanNumLow{1}) }{ \leq }  & 24 c_1 \epsilon_a \Delta_i \rb{ c_4 m \frac{ 4 \Delta_i \sqrt{m} }{\sqrt{ \pi }} } \sqrt{ 2 \log m } \\
    \nonumber = & \frac{ 96 \sqrt{2} c_1 c_4 \epsilon_a \Delta_i^2 m^{1.5} \sqrt{\log m} }{ \sqrt{\pi} } \\
    =& \frac{ 192 \eta^2 m^{1.5} \LB^2 c_1 c_4 \epsilon_a t^2 \sqrt{ \log m } }{ \sqrt{\pi } },
\end{align}
where inequality ($\RomanNumLow{1}$) uses Lemma \ref{lemma:bound-change-in-act-patterns} and the inequality follows with at least $1 - \frac{1}{c_1} - \exp \rb{ - \frac{ 32 (c_4 - 1)^2 \eta^2 m^2 \LB^2 t^2 }{ \pi }  }$ probability. 
\end{proof}

We denote the upper bound as $\dnp{t}$:
\begin{align*}
\dnp{t} := \frac{ 192 \eta^2 m^{1.5} \LB^2 c_1 c_4 \epsilon_a t^2 \sqrt{ \log m } }{ \sqrt{\pi } }.
\end{align*}

\begin{lemma} \label{lemma:coupling-loss} (Coupling of the loss functions) 
For every $i \in [d]$, for all $x$ with $\norm{x}_2 \leq 1$ and for every time step $t\geq1$, with probability at least $1 - \frac{1}{c_1} - \exp \rb{ - \frac{ 32 (c_4 - 1)^2 \eta^2 m^2 \LB^2 t^2 }{ \pi }  }$ over the random initialization, loss function of neural network and pseudo-network are close for some positive constant $c_1 > 10$ and $c_4 \geq 1$:
\begin{align*}
    \abs{\al_i \rb{ \nabla \ft{t} , x } - \al_i \rb{ \nabla \gt{t} , x } } \leq & \; 3 \dnp{t}.
\end{align*}
Using eq. \eqref{eq:divide-L-in-Li}, with probability at least $1 - \frac{d}{c_1} -d \exp \rb{ - \frac{ 32 (c_4 - 1)^2 \eta^2 m^2 \LB^2 t^2 }{ \pi }  }$ over the random initialization, we have 
\begin{align*}
	\abs{\al \rb{ \nabla \ft{t} , x } - \al \rb{ \nabla \gt{t} , x } } \leq & \; 3 d \dnp{t}.
\end{align*}
\end{lemma}
    
\begin{proof}
\begin{align*}
    \abs{ \al_i \rb{ \nabla \ft{t} , x } - \al_i \rb{ \nabla \gt{t} , x } } \leq & \abs{  \sum_{j=1}^{ Q } \Delta_x \rb{  \dfqpi{j}{i} }  - \sum_{j=1}^{ Q } \Delta_x \rb{ \dgqpi{j}{i} }   }\\ 
    &+ \abs{ \log \rb{  \dfi{i}  } - \log \rb{ \dgi{i} } } \\
    \stackrel{ (\RomanNumLow{1}) }{ \leq }  &  2 \rb{ \sup_{ i \in [ Q ] } \abs{ \dfi{i} - \dgi{i} } } + \abs{ \nnit{i}{(t)} - \pnit{i}{(t)} } \\
    \stackrel{ (\RomanNumLow{2}) }{ \leq } & \; 3 \dnp{t},
\end{align*}
where inequality (\RomanNumLow{1}) follows from 1-Lipschitz continuity of $\log \rb{ \phi( u ) }$ with respect to $u$. Inequality (\RomanNumLow{2}) uses Lemma \ref{lemma:coupling-f-prime-and-g-prime}. Using the definition of $\al$, with at least probability $1 - \frac{d}{c_1} - d \exp \rb{ - \frac{ 32 (c_4 - 1)^2 \eta^2 m^2 \LB^2 t^2 }{ \pi }  }$, we get 
\begin{align*}
	\abs{ \al \rb{ \nabla \ft{t} , x } - \al \rb{ \nabla \gt{t} , x } } &\leq \sum_{i=1}^d \abs{ \al_i \rb{ \nabla \ft{t} , x } - \al_i \rb{ \nabla \gt{t} , x } } \\
	&\leq 3 d \dnp{t}
\end{align*}
\end{proof}

\begin{lemma}
\label{lemma:coupling-delta-f-delta-g}
(Coupling of the gradients of functions) For every $i \in [d]$, for all $x$ with $\norm{x}_2 \leq 1$ and for every time step $t\geq1$, with probability at least $1 - \frac{1}{c_1}$ over random initialization, gradient of derivative of neural network function and derivative of pseudo-network function with respect to parameters are close for any positive constant $c_1 > 10$
\begin{align*}
    \Big{\|} \nabla_{\theta_i} \rb{ \dfit{i}{(t)} } - \nabla_{\theta_i} \rb{ \dgit{i}{(t)} } \Big{\|}_{2, 1} \leq & \; 4 c_1 \epsilon_a \rb{ m \dnp{t} + 2 \abs{ \mHB{t}_i } } \sqrt{ 2 \log m }.
\end{align*}
\end{lemma}
\begin{proof}
Recall that $\theta_i$ is given by
\begin{align*}
\theta_i = \begin{bmatrix}
\theta_{i, 1} \\
\vdots \\
\theta_{i, r} \\
\vdots \\
\theta_{i, m}
\end{bmatrix}.
\end{align*} 
where $\theta_{i, r} = \rb{ w_{i, r}, b_{i, r} } \in \R^{i+2}$.
\begin{align*}
    \Big{\|} \nabla_{\theta_i} \rb{ \dfit{i}{(t)} } &- \nabla_{\theta_i} \rb{ \dgit{i}{(t)} } \Big{\|}_{2, 1} \leq  \; \Big{\|} \phi' \rb{ \nnit{i}{(t)} } \nabla_{\theta_i} \nnit{i}{(t)}\\ 
    &- \phi' \rb{ \pnit{i}{(t)} } \nabla_{\theta_i} \pnit{i}{(t)} \Big{\|}_{2, 1} \\
	\leq & \; \Big{\|} \phi' \rb{ \nnit{i}{(t)} } \nabla_{\theta_i} \nnit{i}{(t)} - \phi' \rb{ \pnit{i}{(t)} } \nabla_{\theta_i} \nnit{i}{(t)} \Big{\|}_{2, 1}  \\
    &+ \Big{\|} \phi' \rb{ \pnit{i}{(t)} } \nabla_{\theta_i} \nnit{i}{(t)} - \phi' \rb{ \pnit{i}{(t)} } \nabla_{\theta_i} \pnit{i}{(t)} \Big{\|}_{2, 1} \\
    \leq & \; \abs{ \phi' \rb{ \nnit{i}{(t)} } - \phi' \rb{ \pnit{i}{(t)} } } \Big{\|} \nabla_{\theta_i} \nnit{i}{(t)} \Big{\|}_{2, 1} \\ 
    &+ \abs{ \phi' \rb{ \pnit{i}{(t)} } } \Big{\|} \nabla_{\theta_i} \nnit{i}{(t)} - \nabla_{\theta_i} \pnit{i}{(t)} \Big{\|}_{2, 1} \\
    \leq & \; \abs{ \nnit{i}{(t)} - \pnit{i}{(t)} } \Big{\|} \nabla_{\theta_i} \nnit{i}{(t)} \Big{\|}_{2, 1} \\ 
    &+ \Big{\|} \nabla_{\theta_i} \nnit{i}{(t)} - \nabla_{\theta_i} \pnit{i}{(t)} \Big{\|}_{2, 1}, \\
\end{align*}
where the last inequality follows from 1-Lipschitzness of $\phi'$ and  $\phi'(x) \leq 1$ for all $x$. 
Now, we will bound $\Big{\|} \nabla_{\theta_i} \nnit{i}{(t)} - \nabla_{\theta_i} \pnit{i}{(t)} \Big{\|}_{2, 1}$:
\begin{align}
\label{eq:upper-bound-diff-delta-N-delta-P}
    \nonumber \Big{\|} \nabla_{\theta_i} \nnit{i}{(t)} - \nabla_{\theta_i} \pnit{i}{(t)} \Big{\|}_{2, 1} \leq & \; \Big{\|} \Big[ ( \onevector \initam{i}{r} , \initam{i}{r} ) \odot (  \nrmxi{i} , 1 ) \odot ( \onevector \ind{ \neurvalm{ i }{(t)}{\nrmqpi{j}{i}} \geq 0 } \\
         \nonumber  & - \onevector \ind{ \neuralvalzerom{i} \geq 0 }, \ind{  \neurvalm{ i }{(t)}{\nrmqpi{j}{i}} \geq 0 } \\ 
         \nonumber &- \ind{ \neuralvalzerom{i} \geq 0 } ) \Big]^{r=1}_m \Big{\|}_{2, 1} \\
     \nonumber \stackrel{ (\RomanNumLow{1}) }{ \leq } & \rb{ 8 c_1 \epsilon_a \sqrt{ 2 \log m }  } \abs{ \mHB{t}_i } \\
     \leq & \; 8 c_1 \epsilon_a \abs{ \mHB{t}_i } \sqrt{ 2 \log m }, 
\end{align}
where inequality $(\RomanNumLow{1})$ follows from Lemma \ref{lemma:bound-change-in-act-patterns} with atleast $1 - \frac{1}{c_1}$ probability.
Now using Eq.\eqref{eq:upper-bound-diff-delta-N-delta-P}, with atleast $1 - \frac{1}{c_1}$ probability, we get
\begin{align*}
    \Big{\|} \nabla_{\theta_i} & \rb{ \dfit{i}{(t)} } -  \nabla_{\theta_i} \rb{ \dgit{i}{(t)} } \Big{\|}_{2, 1} \leq  \abs{ \nnit{i}{(t)} - \pnit{i}{(t)} } \Bigg{\|} \Bigg[ ( \onevector \initam{i}{r} , \initam{i}{r} ) \odot (  \nrmxi{i} , 1 ) \odot \\ 
    & \rb{ \onevector \ind{ \neurvalm{ i }{(t)}{\nrmqpi{j}{i}} \geq 0 }, \ind{  \neurvalm{ i }{(t)}{\nrmqpi{j}{i}} \geq 0 } } \Bigg]^{r=1}_m \Bigg{\|}_{2, 1}   \\  
    &+ \Big{\|} \nabla_{\theta_i} \nnit{i}{(t)} - \nabla_{\theta_i} \pnit{i}{(t)} \Big{\|}_{2, 1} \\
     \stackrel{\eqref{eq:upper-bound-diff-delta-N-delta-P}}{\leq} & \; 8 c_1 \epsilon_a m \dnp{t} \sqrt{2 \log m}  + 8 c_1 \epsilon_a \abs{ \mHB{t}_i } \sqrt{ 2 \log m }  \\
     = & \; 8 c_1 \epsilon_a \rb{ m \dnp{t} + \abs{ \mHB{t}_i } } \sqrt{ 2 \log m }.
\end{align*}
\todo{the proof has not explained where the probability comes from and that this is used to get an upper bound on $a$. This was already used in a previous lemma, so in a way it doesn't need to be explained, but there's something to be fixed here.}
\end{proof}

\begin{lemma}
\label{lemma:coupling-gradient-loss}
 (Coupling of the gradient of loss) For every $i \in [d]$, for all $x$ with $\norm{x}_2 \leq 1$ and for every time step $t\geq1$, with probability at least $1 - \frac{d}{c_1} - d \exp \rb{ - \frac{ 32 (c_4 - 1)^2 \eta^2 m^2 \LB^2 t^2 }{ \pi } }$ over random initialization, gradient of loss function with neural network and loss function with pseudo-network are close for some positive constant $c_1 > 10$ and $c_4 \geq 1$:
\begin{align*}
    \norm{ \nabla_{\theta} \alm{\ft{t}}{ x } - \nabla_{\theta} \alm{ \gt{t}}{ x } }_{2, 1} \leq \; & \frac{ 192 d \eta m^{1.5} \LB c_1 c_4 \epsilon_a t \sqrt{ \log m } }{ \sqrt{\pi} } + 24 c_1 d \epsilon_a m \dnp{t} \sqrt{2 \log m}. 
\end{align*}
\end{lemma}
\begin{proof} We have
\begin{align*}
    \big{\|} \nabla_{\theta_i} \alm{ \ft{t} }{ x }&- \nabla_{\theta_i} \alm{ \gt{t} }{ x } \big{\|}_{2, 1} =  \; \Bigg{\|}  \sum_{j=1}^{ Q }  \Delta_x \nabla_{\theta_i} \rb{ \dfqpit{j}{i}{(t)} } - \frac{ \nabla_{\theta_i} \rb{ \dfit{i}{(t)} } }{ \dfit{i}{(t)} }  \\
&    -  \sum_{j=1}^{ Q }  \Delta_x \nabla_{\theta_i} \rb{ \dgqpit{j}{i}{(t)} } + \frac{ \nabla_{\theta_i} \rb{ \dgqpit{j}{i}{(t)} } }{ \dgit{i}{(t)} } \Bigg{\|}_{2, 1} \\
    \leq & \; \underbrace{   \Bigg{\|} \sum_{j=1}^{ Q } \Delta_x \nabla_{\theta_i} \rb{ \dfqpit{j}{i}{(t)} } - \sum_{j=1}^{ Q } \Delta_x \nabla_{\theta_i} \rb{ \dgqpit{j}{i}{(t)} } \Bigg{\|}_{2, 1} }_{ \RomanNumUp{1} } \\
&    + \underbrace{ \Bigg{\|} \frac{ \nabla_{\theta_i} \rb{ \dgit{i}{(t)} } }{ \dgit{i}{(t)} } - \frac{ \nabla_{\theta_i} \rb{ \dfit{i}{(t)} } }{ \dfit{i}{(t)} } \Bigg{\|}_{2, 1} }_{ \RomanNumUp{2} } 
\end{align*}
We first bound $\RomanNumUp{1}$ using Lemma \ref{lemma:coupling-delta-f-delta-g}:
\begin{align*}
    \RomanNumUp{1} 
    &\leq  \sum_{j=1}^Q \Delta_x \Big{\|} \nabla_{\theta_i} \rb{ \dfqpit{j}{i}{(t)} }  -  \nabla_{\theta_i} \rb{ \dgqpit{j}{i}{(t)} } \Big{\|}_1 \\
    &\leq 16 c_1 \epsilon_a \rb{ m \dnp{t} + \abs{ \mHB{t}_i } } \sqrt{ 2 \log m },
\end{align*}
Now, we bound $\RomanNumUp{2}$:
\begin{align*}
    \RomanNumUp{2} =& \Bigg{\|} \frac{ \nabla_{\theta_i} \rb{ \dgit{i}{(t)} } }{ \dgit{i}{(t)} } - \frac{ \nabla_{\theta_i} \rb{ \dfit{i}{(t)} } }{ \dfit{i}{(t)} } \Bigg{\|}_{2, 1} \\
    =& \Bigg{\|} \frac{ \exp \rb{ \pnit{i}{(t)} } \ind{ \pnit{i}{(t)} < 0 } + \ind{ \pnit{i}{(t)} \geq 0 } }{ \exp \rb{ \pnit{i}{(t)} } \ind{ \pnit{i}{(t)} < 0 } + \rb{ \pnit{i}{(t)} + 1 } \ind{ \pnit{i}{(t)} \geq 0 } } \nabla_{\theta_i} \pnit{i}{(t)} \\ 
    &- \frac{ \exp \rb{ \nnit{i}{(t)} } \ind{ \nnit{i}{(t)} < 0 } + \ind{ \nnit{i}{(t)} \geq 0 } }{ \exp \rb{ \nnit{i}{(t)} } \ind{ \nnit{i}{(t)} < 0 } + \rb{ \nnit{i}{(t)} + 1 } \ind{ \nnit{i}{(t)} \geq 0 } } \nabla_{\theta_i} \nnit{i}{(t)} \Bigg{\|}_{2, 1}\\
    =& \Bigg{\|} \rb{ \ind{ \pnit{i}{(t)} < 0 } + \frac{ \ind{ \pnit{i}{(t)} \geq 0 } }{ \rb{ \pnit{i}{(t)} + 1 } } } \nabla_{\theta_i} \pnit{i}{(t)} \\ 
    &- \rb{ \ind{ \nnit{i}{(t)} < 0 } + \frac{ \ind{ \nnit{i}{(t)} \geq 0 } }{ \rb{ \nnit{i}{(t)} + 1 } } } \nabla_{\theta_i} \nnit{i}{(t)} \Bigg{\|}_{2, 1} \\
    =& \; \underbrace{ \bigg{\|} \nabla_{\theta_i} \pnit{i}{(t)}  - \nabla_{\theta_i} \nnit{i}{(t)} \bigg{\|}_{2, 1} \ind{ \pnit{i}{(t)} < 0, \nnit{i}{(t)} < 0 } }_{ \RomanNumUp{2}_1 } \\
    & + \underbrace{ \Bigg{\|} \nabla_{\theta_i} \pnit{i}{(t)}  - \frac{ \nabla_{\theta_i} \nnit{i}{(t)} }{ \nnit{i}{(t)} + 1 } \Bigg{\|}_{2, 1} \ind{ \pnit{i}{(t)} < 0, \nnit{i}{(t)} \geq 0 } }_{ \RomanNumUp{2}_2 } \\
    & + \underbrace{ \Bigg{\|} \frac{ \nabla_{\theta_i} \pnit{i}{(t)} }{ \pnit{i}{(t)} + 1 }  -  \nabla_{\theta_i} \nnit{i}{(t)} \Bigg{\|}_{2, 1} \ind{ \pnit{i}{(t)} \geq 0, \nnit{i}{(t)} < 0 } }_{ \RomanNumUp{2}_3 } \\
    & + \underbrace{ \Bigg{\|} \frac{ \nabla_{\theta_i} \pnit{i}{(t)} }{ \pnit{i}{(t)} + 1 }  -  \frac{ \nabla_{\theta_i} \nnit{i}{(t)} }{ \nnit{i}{(t)} + 1 } \Bigg{\|}_{2, 1} \ind{ \pnit{i}{(t)} \geq 0, \nnit{i}{(t)} \geq 0 } }_{ \RomanNumUp{2}_4 }.
\end{align*}
On simplifying $\RomanNumUp{2}_2$, we get
\begin{align}
\label{eq:upper-bound-on-roman2-gradient}
    \nonumber \RomanNumUp{2}_2 \leq & \Bigg( \abs{ \frac{1}{ \nnit{i}{(t)} + 1 } } \bigg{\|} \nabla_{\theta_i} \pnit{i}{(t)}  - \nabla_{\theta_i} \nnit{i}{(t)} \bigg{\|}_{2, 1} \\ 
    \nonumber &+ \abs{ \frac{ \nnit{i}{(t)} }{ 1 + \nnit{i}{(t)} } } \bigg{\|} \nabla_{\theta_i} \pnit{i}{(t)} \bigg{\|}_{2, 1}  \Bigg)  \ind{ \pnit{i}{(t)} < 0, \nnit{i}{(t)} \geq 0 } \\
    \stackrel{\eqref{eqn:dnpt}}{\leq} & \rb{ \bigg{\|} \nabla_{\theta_i} \pnit{i}{(t)}  - \nabla_{\theta_i} \nnit{i}{(t)} \bigg{\|}_{2, 1} +   \dnp{t} \bigg{\|} \nabla_{\theta_i} \pnit{i}{(t)} \bigg{\|}_{2, 1}  }  \ind{ \pnit{i}{(t)} < 0, N(x; \tht{t} ) \geq 0 }.
\end{align}
Similarly, on simplifying $\RomanNumUp{2}_3$, we get
\begin{align}
\label{eq:upper-bound-on-roman3-gradient}
    \nonumber \RomanNumUp{2}_3 &\leq \Bigg( \abs{ \frac{1}{ \pnit{i}{(t)} + 1 } } \bigg{\|} \nabla_{\theta_i} \pnit{i}{(t)}  - \nabla_{\theta_i} \nnit{i}{(t)} \bigg{\|}_{2, 1} \\ 
    &+ \abs{ \frac{ \pnit{i}{(t)} }{ 1 + \pnit{i}{(t)} } } \bigg{\|} \nabla_{\theta_i} \nnit{i}{(t)} \bigg{\|}_{2, 1}  \Bigg)  \ind{ \pnit{i}{(t)} \geq 0, \nnit{i}{(t)} < 0 } \\
    &\leq \rb{ \bigg{\|} \nabla_{\theta_i} \pnit{i}{(t)} - \nabla_{\theta_i} \nnit{i}{(t)} \bigg{\|}_{2, 1} +   \dnp{t} \bigg{\|} \nabla_{\theta_i} \nnit{i}{(t)} \bigg{\|}_{2, 1}  }  \ind{ \pnit{i}{(t)} \geq 0, \nnit{i}{(t)} < 0 } .
\end{align}
On simplifying $\RomanNumUp{2}_4$, we get
\begin{align}
\label{eq:upper-bound-on-roman4-gradient}
    \nonumber \RomanNumUp{2}_4 \leq & \Bigg( \Bigg{\|} \frac{ \nabla_{\theta_i} \pnit{i}{(t)} }{ \pnit{i}{(t)} + 1 } - \frac{ \nabla_{\theta_i} N(x; \tht{t} ) }{ \pnit{i}{(t)} + 1 } \Bigg{\|}_{2, 1} + \Bigg{\|} \frac{ \nabla_{\theta_i} \nnit{i}{(t)} }{ \pnit{i}{(t)} + 1 }  \\ 
    \nonumber &-  \frac{ \nabla_{\theta_i} \nnit{i}{(t)} }{ \nnit{i}{(t)} + 1 }  \Bigg{\|}_{2, 1} \Bigg) \ind{ \pnit{i}{(t)} \geq 0, \nnit{i}{(t)} \geq 0 } \\
    \nonumber \leq & \Bigg( \frac{1}{ \pnit{i}{(t)} + 1 } \bigg{\|} \nabla_{\theta_i} \pnit{i}{(t)} -  \nabla_{\theta_i} \nnit{i}{(t)} \bigg{\|}_{2, 1} \\ 
    \nonumber &+ \frac{ \big{\|} \nabla_{\theta_i} \nnit{i}{(t)} \big{\|}_{2, 1} \dnp{t} }{ \rb{ \pnit{i}{(t)} + 1 } \rb{ \nnit{i}{(t)} + 1 }  } \Bigg) \ind{ \pnit{i}{(t)} \geq 0, \nnit{i}{(t)} \geq 0 } \\
    \leq & \rb{ \Big{\|} \nabla_{\theta_i} \pnit{i}{(t)} -  \nabla_{\theta_i} \nnit{i}{(t)} \Big{\|}_{2, 1} +  \dnp{t}\Big{\|} \nabla_{\theta_i} \nnit{i}{(t)} \Big{\|}_{2, 1}  } \ind{ \pnit{i}{(t)} \geq 0, \nnit{i}{(t)} \geq 0 } .
\end{align}
Using (\ref{eq:upper-bound-on-roman2-gradient}), (\ref{eq:upper-bound-on-roman3-gradient}) and (\ref{eq:upper-bound-on-roman4-gradient}), we have
\begin{align*}
    \RomanNumUp{2} 
    \leq & \Big{\|} \nabla_{\theta_i} \pnit{i}{(t)}  - \nabla_{\theta_i} \nnit{i}{(t)} \Big{\|}_{2, 1} + \dnp{t} \Big{\|} \nabla_{\theta_i} \nnit{i}{(t)} \Big{\|}_{2, 1}  \ind{ \pnit{i}{(t)} \geq 0 } \\
    & + \dnp{t} \Big{\|} \nabla_{\theta_i} \pnit{i}{(t)} \Big{\|}_{2, 1} \ind{ \pnit{i}{(t)} < 0, \nnit{i}{(t)} \geq 0 }.
\end{align*}

Using (\ref{eq:upper-bound-diff-delta-N-delta-P}), we get 
\begin{align}
\label{eq:uppre-bound-on-roman2}
    \nonumber \RomanNumUp{2} \leq & \; 8 c_1 \epsilon_a \abs{ \mHB{t}_i } \sqrt{ 2 \log m } + \dnp{t} \rb{  \Big{\|} \nabla_{\theta_i} \pnit{i}{(t)} \Big{\|}_{2, 1} + \Big{\|} \nabla_{\theta_i} \nnit{i}{(t)} \Big{\|}_{2, 1}  } \\
    \nonumber \leq & \; 8 c_1 \epsilon_a \abs{ \mHB{t}_i } \sqrt{ 2 \log m } + \dnp{t} \Bigg( \Bigg{\|} \Bigg[ ( \onevector \initam{i}{r} , \initam{i}{r} ) \odot (  \nrmxi{i} , 1 ) \odot  \Bigg( \onevector \ind{ \neuralvalzerom{ i } \geq 0 }, \\
    \nonumber &     \ind{ \neuralvalzerom{ i } \geq 0 } \Bigg) \Bigg]^{r=1}_m \Bigg{\|}_{2, 1} + \Bigg{\|} \Bigg[ ( \onevector \initam{i}{r} , \initam{i}{r} ) \odot (  \nrmxi{i} , 1 ) \odot \\ 
    \nonumber & \rb{ \onevector \ind{ \neurvalm{ i }{(t)}{\nrmxi{i} } \geq 0 }, \ind{  \neurvalm{ i }{(t)}{ \nrmxi{i}} \geq 0 } } \Bigg]^{r=1}_m \Bigg{\|}_{2, 1} \Bigg) \\ 
    \nonumber \leq & \; 8 c_1 \epsilon_a \abs{ \mHB{t}_i } \sqrt{ 2 \log m } + \dnp{t} \Big( 8 c_1 \epsilon_a m \sqrt{2 \log m} \Big) \\
    =& \; 8 c_1 \epsilon_a \rb{ \abs{ \mHB{t}_i } + m \dnp{t} } \sqrt{ 2 \log m }. 
\end{align}
Combining bounds on $\RomanNumUp{1}$ and $\RomanNumUp{2}$, we get
\begin{align*}
    \big{\|} \nabla_{\theta_i} \alm{ \ft{t}}{ x}  & -  \nabla_{\theta_i} \alm{ \gt{t} }{ x } \big{\|}_{2, 1} \leq  16 c_1 \epsilon_a \rb{ m \dnp{t} +  \abs{ \mHB{t}_i } } \sqrt{ 2 \log m } \\ 
    &+ 8 c_1 \epsilon_a \rb{ \abs{ \mHB{t}_i } + m \dnp{t} } \sqrt{ 2 \log m } \\
     \leq & \;  24 c_1 \epsilon_a \rb{  m \dnp{t} +  \abs{ \mHB{t}_i } } \sqrt{ 2 \log m }. 
\end{align*}
Using Lemma \ref{lemma:bound-wt-bt} and Lemma \ref{lemma:bound-change-in-act-patterns}, with at least $1 - \frac{1}{c_1} - \exp \rb{ - \frac{ 32 (c_4 - 1)^2 \eta^2 m^2 \LB^2 t^2 }{ \pi } }$ probability, we get
\begin{align}\label{eq:upper-bound-gradient-of-loss}
    \big{\|} \nabla_{\theta_i} \alm{ \ft{t} }{ x } - \nabla_{\theta_i} \alm{ \gt{t} }{ x } \big{\|}_{2, 1}  \leq \frac{ 192 \eta m^{1.5} \LB c_1 c_4 \epsilon_a t \sqrt{ \log m } }{ \sqrt{\pi} } + 24 c_1 \epsilon_a m \dnp{t} \sqrt{2 \log m} .
\end{align}
We can upper bound $\norm{ \nabla_{\theta} \alm{ \ft{t} }{ x } - \nabla_{\theta} \alm{ \gt{t} }{ x } }_{2, 1}$ as 
\begin{align*}
\norm{ \nabla_{\theta} \alm{ \ft{t} }{ x } - \nabla_{\theta} \alm{ \gt{t} }{ x } }_{2, 1} &\leq \sum_{ i=1 }^d \norm{ \nabla_{\theta_i} \alm{ \ft{t} }{ x } - \nabla_{\theta_i} \alm{ \gt{t} }{ x } }_{2, 1} \\
&\leq \frac{ 192 d \eta m^{1.5} \LB c_1 c_4 \epsilon_a t \sqrt{ \log m } }{ \sqrt{\pi} } + 24 c_1 d \epsilon_a m \dnp{t} \sqrt{2 \log m}
\end{align*}
where last inequality follows from $1 - \frac{d}{c_1} - d \exp \rb{ - \frac{ 32 (c_4 - 1)^2 \eta^2 m^2 \LB^2 t^2 }{ \pi } }$.
\end{proof}

We define $\Gamma$ as 
\begin{align*}
 \Gamma := \frac{ 192 d \eta m^{1.5} \LB c_1 c_4 \epsilon_a T \sqrt{ \log m } }{ \sqrt{\pi} } + 24 c_1 d \epsilon_a m \dnp{t} \sqrt{2 \log m}.
\end{align*}
Note that $\Gamma$ is an upper bound on $\big{\|} \nabla_{\theta} \alm{ \ft{T} }{ x } - \nabla_{\theta} \alm{ \gt{T} }{ x } \big{\|}_{2, 1}$.

\section{Approximation}
\label{sec:existence}
In this section, we will prove that each pseudo network can approximate any target function from target class with small offset $\theta^*$ from the weights of initialization. We first prove that expectation of multiplication of a fixed $\omega$ function and $\ind{ \dotp{w}{x} + b \geq 0 }$ can approximate any smooth activation in target function (Lemma \ref{lemma:pseudo-net-approx-smooth-func}). This is used to prove that $\dgit{i}{*}$ can approximate any target function in target class in $L_{\infty}$ norm. Using Lipschitz continuity $\al$ with respect to $\dgit{i}{*}$, we prove that $\al ( \nabla_i g_i^* , x )$ is close to $\al ( \nabla_i \tarfunc{*} , x )$, where $\tarfunc{*}$ is any target function in the target class.

To prove results in this section, we require a number of new techniques on top of techniques from \cite{allen2019learning}. The target functions in \cite{allen2019learning} are more restricted because $L_2-$norm of weights in target function is equal to 1 (i.e., $\abs[0]{ \tftp{i, r} }, \norm[0]{ \tfhw{i, r} }_2 = 1$). In our paper, we relax this condition and allow any weights with their norm bounded by 1 (i.e., $\abs[0]{ \tftp{i, r} }, \norm[0]{ \tfhw{i, r} }_2 \leq 1$). Our proof can easily be extended to weights bounded by any constant. Additionally, our proof requires to bound $L_{\infty}$ approximation error between pseudo network $\dgit{i}{*}$ and target network, which is a stronger condition than $L_1$ approximation error given in \cite{allen2019learning}, and requires a new proof technique. 
\begin{lemma} \label{lemma:hermite-polynomial-property} For any fixed constant $0 < C \leq 1$ and even $i > 0$, for any $x_1 \in [0, C]$ and $b$, we have 
\begin{align*}
& \E_{ \alpha, \beta \sim { \N \rb{0, 1} } } \sqb{ h_i \rb{ \frac{ \alpha x_1 + \beta \sqrt{ C^2 - x_1^2 }  }{ C } }  \ind{ \alpha \geq b }  } = q_i x_1^i \text{ where} \\
& q_i = \frac{  \rb{ i-1 }!! \exp \rb{ -\frac{b^2  }{2} }  }{ C^i \sqrt{2\pi}  } \sum_{ r=1, \text{odd} }^{ (i-1) } \frac{ \rb{ -1 }^{ \frac{i-r-1}{2} } }{r!!}  \binom{i/2 - 1}{ (r-1)/2 } b^r.
\end{align*}
Similarly, for any fixed constant $C > 0$ and odd $i > 0$, for any $x_1 \in [0, C]$ and $b$, we have
\begin{align*}
& \E_{ \alpha, \beta \sim { \N \rb{0, 1} } } \sqb{ h_i \rb{ \frac{ \alpha x_1 + \beta \sqrt{ C^2 - x_1^2 }  }{ C } }  \ind{ \alpha \geq b }  } = q_i x_1^i \text{ where} \\
& q_i = \frac{  \rb{ i-1 }!! \exp \rb{ -\frac{b^2 }{2} }  }{ C^i \sqrt{2\pi}  } \sum_{ r=0, \text{even} }^{ (i-1) } \frac{ \rb{ -1 }^{ \frac{i-r-1}{2} } }{r!!}  \binom{i/2 - 1}{ (r-1)/2 } b^r.
\end{align*}
\end{lemma}

\begin{proof}
Using summation formula from Fact \ref{fact:sum-prod-hermite}, we have
\begin{align*}
h_{i} \rb{ \frac{ \alpha x_1 + \beta \sqrt{C^2 - x_1^2}  }{C} } = \sum_{k=0}^i \binom{i}{k} \rb{ \frac{\alpha x_1 }{C} }^{i-k} h_k \rb{ \beta \sqrt{ 1 - \frac{x_1^2}{C^2} } }. \\
\end{align*}
Expanding $h_k \rb{ \beta \sqrt{ 1 - \frac{x_1^2}{C^2} } }$ using multiplication formula of Hermite polynomial from Fact \ref{fact:sum-prod-hermite}, we get
\begin{align}
\label{eq:hermite-expansion}
h_k \rb{ \beta \sqrt{ 1 - \frac{x_1^2}{C^2} } } = \sum_{j=0}^{ \lfloor \frac{k}{2} \rfloor } \rb{  1 - \frac{ x_1^2  }{ C^2 } }^{ \frac{k - 2j}{2} } \rb{ - \frac{ x_1^2 }{C^2} }^j \binom{k}{2j} \frac{ (2j)! }{ j! } 2^{-j} h_{k-2j} \rb{ \beta }.
\end{align}
Using Fact \ref{fact:hermite-normal-expectation}, for even $k$, we have
\begin{align}
\label{eq:even-exp}
\E_{ \beta \sim \N \rb{0, 1} } \sqb{ h_k \rb{ \beta \sqrt{ 1 - \frac{x_1^2}{C^2} } } } = \rb{ - \frac{x_1^2}{C^2} }^{ k/2 } \frac{k!}{(k/2)!} 2^{-k/2},
\end{align}
and for odd $k$,
\begin{align}
\label{eq:odd-exp}
\E_{ \beta \sim \N \rb{0, 1} } \sqb{ h_k \rb{ \beta \sqrt{ 1 - \frac{x_1^2}{C^2} } } } = 0.
\end{align}
Using Eq. (\ref{eq:hermite-expansion}), Eq. (\ref{eq:even-exp}) and Eq.(\ref{eq:odd-exp}), we get
\begin{align*}
\E_{ \beta \sim \N \rb{0, 1} } \sqb{ h_{i} \rb{ \frac{ \alpha x_1 + \beta \sqrt{C^2 - x_1^2}  }{C} } } &= \sum_{ k=0, \text{even} }^i \binom{i}{k} \rb{ \frac{\alpha x_1}{C} }^{i-k} \rb{ - \frac{x_1^2}{C^2} }^{k/2} \frac{k!}{ (k/2)! } \rb{ -2 }^{ -k/2 } \\
&= \frac{x_1^i}{C^i} \sum_{ k=0, \text{even} }^i \binom{i}{k} \alpha^{i-k} \frac{k!}{ (k/2)! } \rb{ -2 }^{-k/2}.
\end{align*}
Using $\ind{ \frac{\alpha}{C} \geq b }$ in the expectation, we have
\begin{align}
\label{eq:exp-first}
\E_{ \alpha, \beta \sim \N \rb{0, 1} } \sqb{ h_{i} \rb{ \frac{ \alpha x_1 + \beta \sqrt{C^2 - x_1^2}  }{C} } \ind{ \alpha \geq b } } = \frac{x_1^i}{C^i} \sum_{ k=0, \text{even} }^i \binom{i}{k} \E_{ \alpha \sim \N \rb{0, 1} } \sqb{ \alpha^{i-k} \ind{ \alpha \geq b } } \frac{k!}{ (k/2)! } \rb{ -2 }^{-k/2}.
\end{align}
Define $B_{i, b}$ as
\begin{align*}
B_{i, b} := \E_{\alpha \sim \N \rb{0 ,1} } \sqb{ \alpha^i \ind{  \alpha  \geq b }  }.
\end{align*}
Now, we divide our proof in two parts. In (a), we complete the proof for even $i>0$ and in (b), we do it for odd $i$.
\begin{enumerate}[(a)]
\item Using Lemma \ref{lemma:induction-lemma}, for even $i \geq 0$, we have
\begin{align*}
B_{i, b} = \rb{ i-1 }!! \Phi \rb{0, 1;  b} + \phi \rb{0, 1;  b} \sum_{j=1, j \text{odd}}^{i-1} \frac{ (i-1)!! }{ j!! } b^j
\end{align*}
Using Eq. (\ref{eq:exp-first}), we have
\begin{align*}
& \E_{ \alpha, \beta \sim \N \rb{0, 1} } \sqb{ h_{i} \rb{ \frac{ \alpha x_1 + \beta \sqrt{C^2 - x_1^2}  }{C} } \ind{ \alpha \geq b } } \\
=& \frac{x_1^i}{C^i} \rb{ \sum_{k=0, \text{even}}^i \binom{i}{k} B_{i-k, b} \frac{k!}{ (k/2)! } \rb{ -2 }^{-k/2} } \\
=& \frac{x_1^i}{C^i} \rb{ \sum_{k=0, \text{even}}^i \binom{i}{k} \rb{i-k-1}!! \Phi \rb{0, 1; b} \frac{k!}{ (k/2)! } \rb{ -2 }^{-k/2} } \\
&+ \frac{x_1^i}{ C^i } \phi \rb{0, 1; b} \rb{ \sum_{k=0, \text{even}} \binom{i}{k} \rb{ \sum_{j=1, \text{odd}}^{i-k-1} \frac{ (i-k-1)!! }{ j!! } b^j } \frac{k!}{ (k/2)! } (-2)^{-k/2} }.
\end{align*}
Using
\begin{align*}
\sum_{k=0, \text{even}}^i \binom{i}{k} \rb{i-k-1}!! \frac{k!}{ (k/2)! } \rb{ -2 }^{-k/2} &= \sum_{k=0, \text{even}}^i \frac{ i! \rb{i-k-1}!! k!  \rb{ -2 }^{-k/2} }{ (i-k)! k! (k/2)! } \\
&= \sum_{k=0, \text{even}}^i \frac{ i! (-1)^{k/2} }{ (i-k)!! (k/2)! 2^{k/2} } \\ 
&= \rb{ i-1 }!! \sum_{k=0, \text{even}}^i \frac{ i!! (-1)^{k/2} }{ \rb{i-k}!! (k/2)! 2^{k/2} } \\
&= \rb{ i-1 }!! \sum_{k=0, \text{even}}^i \binom{i/2 }{ k/2 } \rb{ -1 }^{k/2} \\
&= 0,
\end{align*}
we get
\begin{align}
\label{eq:even-final-with-cr}
\E_{ \alpha, \beta \sim \N \rb{0, 1} } \sqb{ h_{i} \rb{ \frac{ \alpha x_1 + \beta \sqrt{C^2 - x_1^2}  }{C} } \ind{ \alpha \geq b } } = \frac{x_1^i}{C^i} \rb{ i-1 }!! \phi \rb{0, 1; b} \sum_{r=1, \text{odd}}^{i-1} c_r b^r
\end{align}
where $c_r$ is given by
\begin{align*}
c_r &:= \frac{1}{ \rb{i-1}!! } \sum_{k=0, \text{even}}^{i-r-1} \binom{i}{k} \frac{ \rb{i-k-1}!! k! \rb{ -2 }^{-k/2} }{ r!! (k/2)! } \\
&= \frac{1}{ \rb{i-1}!! } \sum_{k=0, \text{even}}^{i-r-1} \frac{ i! \rb{i-k-1}!! k! \rb{ -2 }^{-k/2} }{ (i-k)! k! r!! (k/2)! } \\
&= \sum_{k=0, \text{even}}^{i-r-1} \frac{ i!! \; \rb{ -2 }^{-k/2} }{ (i-k)!! \; r!! (k/2)! } \\
&= \sum_{k=0, \text{even}}^{i-r-1} \binom{ i/2 }{ k/2 }  \frac{ \rb{ -1 }^{k/2} }{ r!! } \\
&= \sum_{j=0, \text{even}}^{(i-r-1)/2} \binom{ i/2 }{ j }  \frac{ \rb{ -1 }^{j} }{ r!! } \\
&= \sum_{j=0, \text{even}}^{(i-r-1)/2} \binom{ i/2 }{ j }  \frac{ \rb{ -1 }^{j} }{ r!! } \\
&= \sum_{j=0, \text{even}}^{(i-r-1)/2} \binom{j - i/2 - 1}{ j }  \frac{ 1 }{ r!! } \\
&= \frac{ 1 }{ r!! }  \binom{ -i/2 +\rb{ i - r - 1}/2 }{ \rb{ i - r - 1}/2  }  \\
&= \frac{ \rb{-1}^{ \rb{i-r-1}/2 } }{r!!} \binom{ i/2 -1 }{ \rb{ i - r - 1}/2  } \\
&= \frac{ \rb{-1}^{ \rb{i-r-1}/2 } }{r!!} \binom{ i/2 -1 }{ \rb{ r - 1}/2  } .
\end{align*}
Using value of $c_r$ in Eq.(\ref{eq:even-final-with-cr}), we get the required result.
\item By Lemma \ref{lemma:induction-lemma} for odd $i > 0$, we get
\begin{align*}
B_{i, b} = \phi \rb{0, 1; b} \sum_{j=0, \text{even}}^{i-1} \frac{ (i-1)!! }{ j!! } b^j.
\end{align*}
Using Eq.(\ref{eq:exp-first}), we get
\begin{align}
\label{eq:odd-exp-without-cr}
\nonumber &\E_{ \alpha, \beta \sim \N \rb{0, 1} } \sqb{ h_{i} \rb{ \frac{ \alpha x_1 + \beta \sqrt{C^2 - x_1^2}  }{C} } \ind{ \alpha \geq b } } \\ 
\nonumber &= \frac{x_1^i}{C^i} \sum_{ k=0, \text{even} }^i \binom{i}{k} B_{i-k, b}  \frac{k!}{ (k/2)! } \rb{ -2 }^{-k/2} \\
\nonumber &= \frac{x_1^i}{C^i} \phi \rb{0, 1; b} \sum_{ k=0, \text{even} }^i \binom{i}{k} \rb{ \sum_{j=0, \text{even}}^{i - k -1} \frac{ \rb{i - k - 1}!! }{ j!! } b^j }  \frac{k!}{ (k/2)! } \rb{ -2 }^{-k/2}  \\
&= \frac{x_1^i}{C^i} \phi \rb{0, 1; b} (i-1)!! \sum_{ r=0, r \text{even} }^{ i-1 } c_r b^r
\end{align}
where $c_r$ is given by
\begin{align*}
c_r = \frac{1}{ \rb{i=1}!! } \sum_{ k=0, \text{even} }^{i-r-1} \frac{ ( i-k-1 )!! }{ r!! } \frac{ k! }{ (k/2)! } \rb{ -2 }^{-k/2}.
\end{align*}
By a similar calculation given in part (a), we get
\begin{align*}
c_r = \frac{ \rb{-1}^{ \rb{i-r-1}/2 } }{r!!} \binom{ i/2 -1 }{ \rb{ r - 1}/2  } .
\end{align*}
Using value of $c_r$ in Eq.(\ref{eq:odd-exp-without-cr}), we get the required result.
\end{enumerate}
\end{proof}

\begin{lemma}
\label{lemma:induction-lemma}
Define $B_{i, b}$ as
\begin{align*}
B_{i, b} := \E_{ \alpha \sim \N \rb{0, 1} } \sqb{ \alpha^i \ind{ \alpha \geq b } }.
\end{align*}
 and define $\Phi \rb{0, 1; b}$ and $\phi \rb{ 0, 1; b}$ as 
\begin{align*}
\Phi \rb{0, 1; b} &= \Pr_{ \alpha \sim \N \rb{0, 1} } \sqb{ \alpha \geq b } \\
\phi \rb{0, 1; b} &= \frac{1}{ \sqrt{2 \pi} } \exp \rb{ \frac{- b^2}{2} }
\end{align*}
For any $b$, we have
\begin{align}
&\text{for even } i \geq 0: \hspace{1cm} B_{i, b} = \rb{ i-1 }!! \Phi \rb{0, 1; b} + \phi \rb{0, 1; b} \sum_{ j=1, \text{odd} }^{i-1} \frac{ (i-1)!! }{j!!} b^j \\
&\text{for odd } i > 0: \hspace{1cm} B_{i, b} = \phi \rb{0, 1; b} \sum_{ j=1, \text{even} }^{i-1} \frac{ (i-1)!! }{j!!} b^j
\end{align}
\end{lemma}
\begin{proof}
The lemma follows from Lemma A.7 of \cite{allen2019learning}.
\end{proof}

We will use two different view of the randomness. Define $w_0$ as $w_0 = \rb{ \alpha_1, \beta_1 }$ and $x = \rb{ x_1, \sqrt{C^2 - x_1^2} }$ where $\alpha_1$ and $\beta_1$ are standard normal random variables and $C$ is any positive constant. In alternative view of randomness, we write $w_0$ as
\begin{align*}
w_0 = \frac{ \dotp{ w_0 }{ x }  }{ \| x \|^2 } x + \frac{ \dotp{ w_0 }{ x^\perp } }{  \| x^\perp \|^2  } x^\perp
\end{align*}
where $x^\perp = \rb{ \sqrt{C^2 - x_1^2}, -x_1 }$. Define $\alpha'=\dotp{w_0}{ x } $ and $\beta'=\dotp{ w_0 }{ x^\perp } $ where $\alpha'$ and $\beta'$ are normal random variables with $0$ mean and $C^2$ variance. Using definitions of $\alpha'$ and $\beta'$, we get
\begin{align*}
w_0 = \frac{\alpha'}{C^2} x + \frac{ \beta' }{C^2} x^\perp = \frac{\alpha}{C} x + \frac{ \beta }{C} x^\perp
\end{align*}
where $\alpha$ and $\beta$ are standard normal random variable.

\begin{lemma} \label{lemma:polynomial-x-i} For every integer $i \geq 1$, there exists a constant $q_i'$ with $ \abs{ q_i' } \geq \frac{ ( i-1 )!! }{ 200 i^2 C^i } $ such that
\begin{align*}
&\text{for even }i: \hspace{1cm} x_1^i = \frac{1}{q_i'} \E_{ w_0 \sim \N \rb{0, \mathbf{I}}, b_0 \sim \N \rb{0, 1} } \sqb{ h_i \rb{ \alpha_1 } \ind{ 0 \leq -b_0 \leq 1/(2 i) } \ind{ \frac{ \dotp{ w_0 }{ x } }{ C } + b_0 \geq 0 } } \\
&\text{for odd }i: \hspace{1cm} x_1^i = \frac{1}{q_i'} \E_{ w_0 \sim \N \rb{0, \mathbf{I}}, b_0 \sim \N \rb{0, 1} } \sqb{ h_i \rb{ \alpha_1 } \ind{ \; \abs{ b_0 } \leq 1/(2 i) } \ind{ \frac{ \dotp{ w_0 }{ x } }{ C } + b_0 \geq 0 } } 
\end{align*}
\end{lemma}

\begin{proof}
First, we will prove for even $i$. 
By Lemma \ref{lemma:hermite-polynomial-property}, we get
\begin{align}
\label{eq:exp-x-i}
\nonumber &\E_{ w_0 \sim \N \rb{0, \mathbf{1}}, b_{0} \sim \N \rb{0, 1} } \sqb{ h_i \rb{ \alpha_1 } \ind{ 0 \leq - b_0 \leq 1/(2i) } \ind{    \frac{ \dotp{w_0}{x} }{ C } + b_0 \geq 0 } } \\
\nonumber &= \E_{ b_0 \sim \N \rb{0, 1} } \sqb{ \E_{ \alpha, \beta \sim \N \rb{0, 1} } \sqb{ h_i \rb{ \frac{ \alpha x_1 + \beta \sqrt{C^2 - x_1^2} }{C} } \ind{ \alpha \geq -b_0 } }  \ind{ 0 \leq - b_0 \leq 1/(2i) } } \\
&= \E_{b_0 \sim \N \rb{0, 1}} \sqb{ q_i \ind{ 0 \leq - b_0 \leq 1/(2i) } } x_1^i 
\end{align}
where 
\begin{align*}
q_i = \frac{  \rb{ i-1 }!! \exp \rb{ -\frac{b^2 }{2} }  }{ C^i \sqrt{2\pi}  } \sum_{ r=0, \text{even} }^{ (i-1) } \frac{ \rb{ -1 }^{ \frac{i-r-1}{2} } }{r!!}  \binom{i/2 - 1}{ (r-1)/2 } \rb{ - b_0 }^r.
\end{align*}
Now, we try to bound the coefficient $\E_{b_0 \sim \N \rb{0, 1}} \sqb{ q_i \ind{ 0 \leq - b_0 \leq 1/(2i) } }$. Define $c_r$ as
\begin{align*}
c_r := \frac{ \rb{ -1 }^{ \frac{i-r-1}{2} } }{r!!}  \binom{i/2 - 1}{ (r-1)/2 }.
\end{align*}
For $0 \leq - b_0 \leq 1/(2i)$ and for all odd $r$ with $1 < r \leq i-1$, 
\begin{align*}
\abs{ c_r  \rb{ -b_0 }^r } = \abs{ \frac{ \rb{ -1 }^{ \frac{i-r-1}{2} } }{r!!}  \binom{i/2 - 1}{ (r-1)/2 } \rb{ -b_0 }^r } \leq \abs{ \frac{ \rb{ -1 }^{ \frac{i-r+1}{2} } }{ (r-2)!!}  \binom{i/2 - 1}{ (r-3)/2 } \rb{ -b_0 }^r } \leq \frac{1}{4} \abs{ c_{r-2} \rb{ -b_0 }^{r-2} }.
\end{align*}
Using above relation, we get

\begin{align*}
\abs{ \sum_{r = 1, \text{odd}}^{i-1} c_r \rb{ -b_0 }^r } &\geq \abs{ \abs{ c_1 b_0 } - \abs{ \sum_{r=3, \text{odd}} c_r \rb{-b_0}^r } } \\
&\geq \abs{ \abs{ c_1 b_0 } - \abs{ \sum_{r=1}^{\infty}  \frac{1}{4^r} \abs{ c_{1} \rb{ b_0 } } } } \\
&\geq \abs{ \abs{c_1 b_0}  - \frac{1}{3}\abs{ \abs{c_1 b_0} } } \\
&\geq \frac{2}{3} \abs{ c_1 b_0 },
\end{align*}
and
\begin{align*}
\text{sign} \rb{ \sum_{r = 1, \text{odd}}^{i-1} c_r \rb{ -b_0 }^r } = \text{sign} \rb{ c_1 \rb{ -b_0 } } = \text{sign} \rb{ c_1 }.
\end{align*}
Using Eq.(\ref{eq:exp-x-i}), we get
\begin{align*}
&\abs{  \E_{b_0 \sim \N \rb{0, 1}} \sqb{ q_i \ind{ 0 \leq -b_0 \leq 1/(2i) } } } \\
&= \abs{ \E_{b_0 \sim \N \rb{0, 1}} \sqb{ \frac{  \rb{ i-1 }!! \exp \rb{ -\frac{b^2 }{2} }  }{ C^i \sqrt{2\pi}  } \sum_{ r=0, \text{even} }^{ (i-1) } c_r  \rb{ - b_0 }^r \ind{ 0 \leq -b_0 \leq 1/(2i) } } } \\
&= \abs{ \E_{b_0 \sim \N \rb{0, 1}} \sqb{ \frac{  \rb{ i-1 }!! \exp \rb{ -\frac{b^2  }{2} }  }{ C^i \sqrt{2\pi}  } \text{sign} \rb{ \sum_{ r=0, \text{even} }^{ (i-1) } c_r  \rb{ - b_0 }^r } \abs{ \sum_{ r=0, \text{even} }^{ (i-1) } c_r  \rb{ - b_0 }^r } \ind{ 0 \leq -b_0 \leq 1/(2i) } } } \\
&\geq  \abs{ \E_{b_0 \sim \N \rb{0, 1}} \sqb{ \frac{  \rb{ i-1 }!! \exp \rb{ -\frac{b^2 }{2} }  }{ C^i \sqrt{2\pi}  } \text{sign} \rb{ c_1 } \frac{2}{3} \abs{ c_1  \ b_0  } \ind{ 0 \leq -b_0 \leq 1/(2i) } } } \\
&\geq \frac{ (i-1)!! }{ 100 i^2 C^i }.
\end{align*}
This completes the proof for even $i$. Similarly for odd $i$, using Lemma \ref{lemma:hermite-polynomial-property}, we get
\begin{align}
\label{eq:odd-i}
\nonumber &\E_{ w_0 \sim \N \rb{0, \mathbf{1}}, b_{0} \sim \N \rb{0, 1} } \sqb{ h_i \rb{ \alpha_1 } \ind{ \abs{ b_0 } \leq 1/(2i) } \ind{    \frac{ \dotp{w_0}{x} }{ C } + b_0 \geq 0 } } \\
\nonumber &= \E_{ b_0 \sim \N \rb{0, 1} } \sqb{ \E_{ \alpha, \beta \sim \N \rb{0, 1} } \sqb{ h_i \rb{ \frac{ \alpha x_1 + \beta \sqrt{C^2 - x_1^2} }{C} } \ind{  \alpha  \geq -b_0 } }  \ind{ \abs{b_0} \leq 1/(2i) } } \\
&= \E_{b_0 \sim \N \rb{0, 1}} \sqb{ q_i \ind{ \abs{ b_0 } \leq 1/(2i) } } x_1^i 
\end{align}
where 
\begin{align*}
q_i = \frac{  \rb{ i-1 }!! \exp \rb{ -\frac{b^2 }{2} }  }{ C^i \sqrt{2\pi}  } \sum_{ r=0, \text{even} }^{ (i-1) } \frac{ \rb{ -1 }^{ \frac{i-r-1}{2} } }{r!!}  \binom{i/2 - 1}{ (r-1)/2 } b^r.
\end{align*}
Now, we will try to bound $\E_{b_0 \sim \N \rb{0, 1}} \sqb{ q_i \ind{ \abs{b_0 } \leq 1/(2i) } }$. Define $c_r$ as
\begin{align*}
c_r := \frac{ \rb{ -1 }^{ \frac{i-r-1}{2} } }{r!!}  \binom{i/2 - 1}{ (r-1)/2 }.
\end{align*}
For $\abs{b_0} \leq 1/(2i)$ and for all even $r$ with $1 < r \leq i-1$, we get
\begin{align*}
\abs{ c_r  \rb{ -b_0 }^r } = \abs{ \frac{ \rb{ -1 }^{ \frac{i-r-1}{2} } }{r!!}  \binom{i/2 - 1}{ (r-1)/2 } \rb{ -b_0 }^r } \leq \abs{ \frac{ \rb{ -1 }^{ \frac{i-r+1}{2} } }{ (r-2)!!}  \binom{i/2 - 1}{ (r-3)/2 } \rb{ -b_0 }^r } \leq \frac{1}{4} \abs{ c_{r-2} \rb{ -b_0 }^{r-2} }.
\end{align*}
Using above relation, we get
\begin{align*}
\abs{ \sum_{r = 1, \text{odd}}^{i-1} c_r \rb{ -b_0 }^r } &\geq \abs{ \abs{ c_0 } - \abs{ \sum_{r=2, \text{even}} c_r \rb{ -b_0 }^r } } \geq \abs{ \abs{ c_0 } - \abs{ \sum_{r=1}^{\infty}  \frac{1}{4^r} \abs{ c_0 } } } \geq \abs{ \abs{c_0}  - \frac{1}{3}\abs{ \abs{c_0} } } = \frac{2}{3} \abs{ c_0 } = \frac{2}{3} \abs{ \binom{ i/2 - 1 }{ -1/2 } } > \frac{1}{2i},
\end{align*}
 and 
 \begin{align*}
 \text{ sign } \rb{ \sum_{r = 1, \text{odd}}^{i-1} c_r \rb{ -b_0 }^r } = \text{sign} \rb{ c_0 }.
\end{align*}
 Using the formula of $q_i$ in Eq. (\ref{eq:odd-i}), we have
 \begin{align*}
&\abs{ \E_{b_0 \sim \N \rb{0, 1}} \sqb{ q_i \ind{ \abs{b_0 } \leq 1/(2i) } } }\\
&= \abs{ \E_{b_0 \sim \N \rb{0, 1}} \sqb{ \frac{  \rb{ i-1 }!! \exp \rb{ -\frac{b^2  }{2} }  }{ C^i \sqrt{2\pi}  } \sum_{ r=0, \text{even} }^{ (i-1) } c_r b^r \ind{ \abs{ b_0 } \leq 1/(2i) } } }\\
&= \abs{  \E_{b_0 \sim \N \rb{0, 1}} \sqb{ \frac{  \rb{ i-1 }!! \exp \rb{ -\frac{b^2  }{2} }  }{ C^i \sqrt{2\pi}  } \text{sign} \rb{ \sum_{ r=0, \text{even} }^{ (i-1) } c_r b^r } \abs{ \sum_{ r=0, \text{even} }^{ (i-1) } c_r b^r } \ind{ \abs{ b_0 } \leq 1/(2i) } }  } \\
&\geq \abs{  \E_{b_0 \sim \N \rb{0, 1}} \sqb{ \frac{  \rb{ i-1 }!! \exp \rb{ -\frac{b^2  }{2} }  }{ C^i \sqrt{2\pi}  } \text{sign} \rb{ c_0 } \frac{1}{2i} \ind{ \abs{b_0 } \leq 1/(2i) } }  } \\
&\geq \frac{ (i-1)!! }{ 100 i^2 C^i }
 \end{align*}
 This completes the proof for odd $i$.
\end{proof}

\begin{lemma}
\label{lemma:write-psi-as-sum}
For any constant $C \leq 1$ and for any arbitary function $\psi: [-C, C] \mapsto \R$, we have
\begin{align*}
\psi \rb{ x_1 } = c_0 + \sum_{i=1}^{ \infty } c_i' \E_{ w_0 \sim \N \rb{ 0, \mathbf{1} }, b_0 \sim \N \rb{0, 1} } \sqb{ h_i \rb{ \alpha_1 } \ind{ G_i( b_0 ) } \ind{ \frac{ \dotp{w_0}{ x } }{ C } + b_0 \geq 0 } }
\end{align*}
where $w_0 = \rb{ \alpha_1, \beta_1 }$, $c_i= i^{\text{th}}$ coefficient of taylor series of $\psi$ function,
\begin{align*}
\abs{c_i'} \leq \frac{ 200 i^2 \abs{ c_i } }{ (i-1)!! }  \hspace{1cm} \text{and}  \hspace{1cm}  G_i\rb{ b_0 } = \begin{cases} \abs{ b_0} \leq 1/(2i) & \text{if } i \text{ is odd} \\
0 < -b_0 \leq 1/(2i) & \text{if } i \text{ is even} 
\end{cases}
\end{align*}
\end{lemma}

\begin{proof}
Using Taylor expansion of function $\psi (x_1)$, we get
\begin{align*}
\psi \rb{ x_1 } &= c_0 + \sum_{i=1, \text{odd}}^{\infty} c_i x_1^i + \sum_{i=2, \text{even}}^{\infty} c_i x_1^i \\
&= c_0 + \sum_{i=1}^{\infty} c_i' \E_{\alpha, \beta, b_0 \sim \N \rb{0, 1}} \sqb{ h_i \rb{ \alpha_1 } \ind{ G_i \rb{ b_0 } } \ind{ \frac{ \dotp{x}{w_0} }{ C } + b_0 \geq 0 } }
\end{align*}
where above relation follows from Lemma \ref{lemma:polynomial-x-i} and $c_i'$ is given by
\begin{align*}
c_i' = \frac{c_i}{ q_i' }, \hspace{1cm} \abs{ c_i' } \leq \frac{ 200 i^2 \abs{ c_i } C^i }{  (i-1)!!  } \hspace{0.5cm} \text{and} \hspace{0.5cm}  G_i\rb{ b_0 } = \begin{cases} \abs{b_0} \leq 1/(2i) & \text{if } i \text{ is odd} \\
0 < -b_0 \leq 1/(2i) & \text{if } i \text{ is even} 
\end{cases}
\end{align*}
\end{proof}

\begin{lemma} \label{lemma:bound-infty-sum} For any $\epsilon \in (0, 1)$ and any positive integer $i$, setting $B_i \stackrel{\deff}{=} 100 i^{1/2} + 10 \sqrt{\log \frac{1}{\epsilon}}$, we have
\begin{enumerate}
\item $\sum_{i=1}^{\infty} \E_{z \sim \N \rb{0, 1}} \sqb{ \abs{ h_i(z) } \ind{ \abs{z} \geq B_i } } \leq \epsilon/8$
\item $ \sum_{i=1}^{\infty} \E_{z \sim \N \rb{0, 1}} \sqb{ \abs{ h_i(B_i) } \ind{ \abs{z} \geq B_i } } \leq \epsilon/8 $
\item $  \sum_{i=1}^{\infty} \E_{z \sim \N \rb{0, 1}} \sqb{ \abs{ h_i(z) } \ind{ \abs{z} \leq B_i } } \leq \frac{1}{2}  \comp{\epsilon}{\psi} $
\end{enumerate}
\end{lemma}
The Lemma is same as Claim C.2 of \cite{allen2019learning}.

\begin{lemma} 
\label{lemma:pseudo-net-approx-smooth-func}
For any positive integer $d$, for any $\epsilon \in (0, 1)$, for every function $\psi$, every $\epsilon \in (0, 1)$, every $u^*, x \in \R^{d}$ with $\norm{u^*}_2 \leq 1$ and $\| x \|_2 = 1$, there exist a function $\omega: \R^3 \to [ -\comp{ \epsilon }{ \psi }, \comp{ \epsilon }{ \psi } ]$ such that
\begin{align}
\label{eq:multivariate-approx}
\abs{ \E_{ w \sim \N \rb{ 0, \mathbf{I} }, b_0 \sim \N \rb{0, 1} } \sqb{ \omega \rb{ \dotp{w}{u^*}, b_0, \| u^* \| } \ind{ \dotp{ w}{ x } + b_0 \geq 0 } } - \psi \rb{ \dotp{u^*}{x} }  } \leq \epsilon.
\end{align}
\end{lemma}

\begin{proof}
Define $\hat{h}_i ( \alpha_1 ) \stackrel{ \deff }{=} h_i ( \alpha_1 ) \ind{ \abs{ \alpha_1 } \leq B_i } + h_i \rb{ \text{sign} ( \alpha_1 ) B_i } \ind{ \abs{\alpha_1} > B_i } $. From Lemma \ref{lemma:write-psi-as-sum}, we get
\begin{align*}
\psi(x_1) &= c_0 + \sum_{i=1}^{\infty} c_i' \E_{\alpha, \beta, b_0 \sim \N \rb{0, 1}} \sqb{ h_i \rb{ \alpha_1 } \ind{ G_i \rb{ b_0 } } \ind{ \frac{ \dotp{x}{w_0} }{ C } + b_0 \geq 0 } } \\
&= c_0 + R'(x_1) + \sum_{i=1}^{\infty} c_i' \E_{\alpha, \beta, b_0 \sim \N \rb{0, 1}} \sqb{ \hat{h}_i \rb{ \alpha_1 } \ind{ G_i \rb{ b_0 } } \ind{ \frac{ \dotp{x}{w_0} }{ C } + b_0 \geq 0 } }.
\end{align*}
where
\begin{align*}
R'(x_1) = \sum_{i=1}^{\infty} c_i' \E_{\alpha, \beta, b_0 \sim \N \rb{0, 1}} \sqb{ \rb{ h_i \rb{ \alpha_1 } \ind{ \abs{\alpha_1} > B_i } - h_i \rb{ \text{sign} ( \alpha_1 ) B_i } \ind{ \abs{\alpha_1} > B_i } } \ind{ G_i \rb{ b_0 } } \ind{ \frac{ \dotp{x}{w_0} }{ C } + b_0 \geq 0 } }.
\end{align*}
Using Lemma \ref{lemma:bound-infty-sum}, we have $\abs{ R'(x_1) } \leq \epsilon/4$. Define $\omega ( \alpha_1, b_0, C )$ as
\begin{align*}
\omega ( \alpha_1, b_0, C ) = 2c_0 + \sum_{i=1}^{\infty} c_i' \hat{h}_i (\alpha_1) \ind{ G_i \rb{ b_0 } }.
\end{align*}
Using definition of $\omega ( \alpha_1, b_0, C )$, we get
\begin{align*}
\abs{ \E_{ \alpha_1, \beta_1, b_0 \sim \N \rb{0, 1} } \sqb{ \omega \rb{ \alpha_1, b_0, C } \ind{ \frac{\alpha_1 x_1 + \beta_1 \sqrt{ C^2 - x_1^2 } }{ C } + b_0 \geq 0 } } } \leq \epsilon / 4.
\end{align*}
Using Lemma \ref{lemma:bound-infty-sum}, we have
\begin{align*}
\abs{ \omega \rb{ \alpha_1, b_0, C } } \leq 2 c_0 + \frac{\epsilon}{8} + \frac{1}{2} \comp{ \epsilon }{ \psi } \leq \comp{ \epsilon }{ \psi }
\end{align*}
This proves that for every function $\psi$, every $\epsilon \in (0, 1)$, every constant $C \in \R$ and for every $x_1 \in [-C, C]$ , there exist a function $\omega: \R^3 \to [ -\comp{\epsilon}{ \psi }, \comp{\epsilon}{\psi} ]$ such that we have
\begin{align}
\label{eq:univariate-approx}
\abs{ \E_{\alpha_1, \beta_1, b_0 \sim \N \rb{0, 1} }  \sqb{ \omega \rb{ \alpha_1, b_0, C }  \ind{ \frac{ \alpha_1 x_1 + \beta_1 \sqrt{ C^2 - x_1^2 } }{ C } + b_0 \geq 0 }  } - \psi ( x_1 ) } \leq \epsilon.
\end{align} 
We denote $u^{*\perp}_{i}$ for $2 \leq  i \leq d$ as $d-1$ orthogonal vectors of $u^*$ with $\| u^{*\perp}_{i} \| = \| u^* \|$. Now, using projection of $w$ on $u^*$, we get
\begin{align}
\label{eq:w-project-existence-thm}
w = \frac{ \dotp{ w }{ u^* } }{ \| u^* \|^2 } u^* + \sum_{i=2}^{d} \frac{ \dotp{ w }{ u^{*\perp}_{i} } }{ \| u^{*\perp}_{i} \|^2 }   u^{*\perp}_{i} = \frac{ \alpha_1' }{ \| u^* \|^2 }  u^* + \sum_{i=2}^{d} \frac{ \alpha_i' }{ \| u^* \|^2 } u^{*\perp}_{i}
\end{align}
where  $\alpha_i'$ for any $i$ such that $1 \leq i \leq d$ is a normal random variable with 0 mean and $\| u^* \|^2$ variance.  Define $x_1'$ as $ x_1' = \dotp{u^*}{x}$. Similarly, define $x_i' = \dotp{ u_i^{* \perp } }{ x }$  for $2 \leq  i \leq d$. Now, dot product $\dotp{w}{x}$ can be written as
\begin{align}
\label{eq:w-x-prod}
\nonumber \dotp{w}{x} &= \frac{1}{ \| u^* \|^2 } \dotp{\alpha_1' u^* + \sum_{i=2}^{d} \alpha_i' u^{*\perp}_{i}  }{x} \\ 
\nonumber &=  \frac{1}{ \| u^* \|^2 } \rb{ \alpha_1' x_1' + \sum_{i=2}^{d} \alpha_i' x_i' } \\
\nonumber &=   \frac{1}{ \| u^* \|^2 } \rb{ \alpha_1' x_1' + \beta_1' \sqrt{ \| u^* \|^2 - x_1'^2  } } \\
&= \frac{1}{ \| u^* \| } \rb{ \alpha_1 x_1' + \beta_1 \sqrt{ \| u^* \|^2 - x_1'^2  } }
\end{align}
where last inequality follows from $ \rb{ \sum_{i=1}^d x_i'^2 }  = \| u^* \|^2 $. Here $\alpha_1$ and $\beta_1$ are standard normal random variables. Setting $C=\| u^* \|$ and using Eq.(\ref{eq:univariate-approx}), Eq. (\ref{eq:w-project-existence-thm}) and Eq.(\ref{eq:w-x-prod}), we get 
\begin{align*}
\abs{ \E_{ w \sim \N \rb{ 0, \mathbf{I} }, b_0 \sim \N \rb{0, 1} } \sqb{ \omega \rb{ \dotp{w}{u^*}, b_0, \| u^* \| } \ind{ \dotp{ w}{ x } + b_0 \geq 0 } } - \psi \rb{ \dotp{u^*}{x} }  } \leq \epsilon
\end{align*}
\end{proof}

\begin{lemma} 
\label{lemma:expect-pseudo-target} For all $i \in [d]$, for any $\epsilon \in (0, 1)$, for any derivative of target function $\frac{ \partial \tarfunci{i} }{ \partial \vecx{i} }$ and for any $x$ with $\norm{x} \leq 1$, there exist a set of parameters $\theta_i^*$ such that we have
\begin{align*}
\abs{ \E_{ \initwm{i}{r}, \initbm{i}{r} \sim \N \rb{0, \frac{1}{m}} } \sqb{ \pnilt{\ell}{i}{*} } - \phi^{-1} \rb{ \frac{ \partial \tarfunci{i} }{ \partial \vecx{i} } } } \leq p_i \epsilon.
\end{align*}
Moreover, $L_{\infty}$ norm of $\theta^*_i$ is given by
\begin{align*}
\norm{ \theta^*_i }_{2, \infty} \leq  \frac{ \sqrt{\pi} \rb{ \sum_{r=1}^{p_i} U_{\omega_{i, r}} } }{m \epsilon_a \sqrt{2}}.
\end{align*}
\end{lemma}

\begin{proof}
We denote pseudo network with parameters $\theta_i^*$ as:
\begin{align*}
\pnilt{\ell}{i}{*} = \sum_{r=1}^m \initam{i}{r} \rb{ \dotp{ w_{i, r}^* }{ \nrmxi{i} } + b_{i, r}^* } \ind{  \neuralvalzerom{i} \geq 0 }.
\end{align*}
Similarly, $\dgit{i}{*}$ is given by $\phi \rb{ \pnilt{\ell}{i}{*} }$. We will use function $\omega_{i, j}$ to approximate a neuron of target function $\psi_{i, j}$ for all $i \in [d], j \in [p_i]$.  Setting $w_{i, r}^*$ and $b_{i, r}^*$ as
\begin{align*}
w_{i, r}^* &= \frac{ \sqrt{\pi} \text{sign} \rb{ \initam{i}{r} } }{m \epsilon_a \sqrt{2}} \sum_{j=1}^{p_i} \tftp{i, j} \omega_{i, j} \rb{ \sqrt{m} \dotp{ \initwm{i}{r} }{ \tfhw{i, j} } , \sqrt{m} \initbm{i}{r}, \| \tfhw{i, j} \|  } \tfhmw{i, j}, \\ 
b_{i, r}^* &= 0,
\end{align*}
we get 
\begin{align*}
& \abs{ \E_{ \initam{i}{r} \sim \N \rb{0, \epsilon_a^2} , \initwm{i}{r}, \initbm{i}{r} \sim \N \rb{0, \frac{1}{m}} } \sqb{ \pnilt{\ell}{i}{*} } - \phi^{-1} \rb{ \frac{ \partial \tarfunci{i} }{ \partial \nrmxi{i} } } } \\ 
=& \abs{ m \E_{ \initam{i}{r} \sim \N \rb{0, \epsilon_a^2} , \initwm{i}{r}, \initbm{i}{r} \sim \N \rb{0, \frac{1}{m}} } \sqb{ \initam{i}{r} \rb{ \dotp{ w_{i, r}^* }{ \nrmxi{i} } + b_{i, r}^* } \ind{  \neuralvalzerom{i} \geq 0 } } - \phi^{-1} \rb{ \frac{ \partial \tarfunci{i} }{ \partial \nrmxi{i} } } } \\ 
=& \Bigg{|} \frac{ \sqrt{\pi} }{ \epsilon_a \sqrt{2} } \E_{ \initam{i}{r} \sim \N \rb{0, \epsilon_a^2} , \initwm{i}{r}, \initbm{i}{r} \sim \N \rb{0, \frac{1}{m}} } \Bigg[ \initam{i}{r} \text{sign} \rb{ \initam{i}{r} } \sum_{j=1}^{p_i} \tftp{i, j} \omega_{i, j} \rb{ \sqrt{m} \dotp{ \initwm{i}{r} }{ \tfhw{i, j} } , \sqrt{m} \initbm{i}{r}, \| \tfhw{i, j} \|  } \dotp{ \tfhmw{i, j} }{ \nrmxi{i} } \\ 
&\ind{  \neuralvalzerom{i} \geq 0 } \Bigg] - \phi^{-1} \rb{ \frac{ \partial \tarfunci{i} }{ \partial \nrmxi{i} } } \Bigg{|} \\
=& \Bigg{|} \E_{ \initwm{i}{r}, \initbm{i}{r} \sim \N \rb{0, \frac{1}{m}} } \sqb{ \sum_{j=1}^{p_i} \tftp{j, r} \omega_{i, j} \rb{ \sqrt{m} \dotp{ \initwm{i}{r} }{ \tfhw{j, r} } , \sqrt{m} \initbm{i}{r}, \| \tfhw{j, r} \|  } \dotp{ \tfhmw{j, r} }{ \nrmxi{i} } \ind{  \neuralvalzerom{i} \geq 0 } } \\
&-  \sum_{j=1}^{p_i} \tftp{i, j}  \psi_{i, j} \rb{  \dotp{ \tfhw{i, j} }{ \nrmxi{i} }  } \rb{ \dotp{ \tfhmw{i, j} }{ \nrmxi{i} }  } \Bigg{|} \\
\leq & \; p_i \epsilon
\end{align*}
Bounding $\| w_{i, r}^*  \|$, we get
\begin{align*}
\| w_{i, r}^*  \|_2 &= \Bigg{\|}  \frac{ \sqrt{\pi} \text{sign} \rb{ \initam{i}{r} } }{m \epsilon_a \sqrt{2}} \sum_{j=1}^{p_i} \tftp{i, j} \omega_{i, j} \rb{ \sqrt{m} \dotp{ \initwm{i}{r} }{ \tfhw{i, j} } , \sqrt{m} \initbm{i}{r}, \| \tfhw{i, j} \|  } \tfhmw{i, j}  \Bigg{\|}_2 \\ 
&\leq \frac{ \sqrt{\pi} \rb{ \sum_{r=1}^{p_i} U_{\omega_{i, r}} } }{m \epsilon_a \sqrt{2}}.
\end{align*}
\end{proof}
Define upper bound on $\| w_{i, r}^*  \|_2$ as 
\begin{align*}
U_{ w_{i}^* } = \frac{ \sqrt{\pi} \rb{ \sum_{r=1}^{p_i} U_{\omega_{i, r}} } }{m \epsilon_a \sqrt{2}}
\end{align*}

\begin{lemma} 
\label{lemma:existence-pseudo-network}
 For any $i \in [d]$, for any $\epsilon \in (0, 1)$, for any derivative of target function $\frac{ \partial \tarfunci{i} }{ \partial \vecx{i} }$, for any $m \geq \Omega \rb{ \frac{ d^{10} \rb{ \sum_{i=1}^d  \sum_{r=1}^{p_i} U_{h_{i, r}} }^{12} }{ \epsilon_a^2 \epsilon^8 } }$ and for any $x$ with $\norm{x} \leq \frac{1}{2}$, there exist a set of parameters $\theta_i^*$ such that, with atleast $1 - \frac{1}{c_1} - \frac{1}{c_2} - \frac{1}{c_3} - \exp \rb{ - \frac{ \epsilon^2}{2 m C_i^2} } - \exp \rb{ - \frac{ 32 (c_4 - 1)^2  m^2 U_{ w_{i}^* }^2 }{ \pi }  } $ probability, we have
\begin{align*} 
\abs{ \phi^{-1} \rb{ \frac{ \partial \tarfunci{i} }{ \partial \vecx{i} } }  - \pnit{i}{*} } &\leq \rb{ p_i + 1} \epsilon + \frac{ 192 c_1 c_4 \epsilon_a m^{1.5} U_{ w_{i}^* }^2 \sqrt{ 2 \log m } }{ \sqrt{\pi} }.
\end{align*}
\end{lemma}
\begin{proof} We divide $\abs{ \phi^{-1} \rb{ \frac{ \partial \tarfunci{i} }{ \partial \vecx{i} } }  - \pnit{i}{*} }$ into five parts as
\begin{align}
\label{eq:divide-approx-using-pseudo-network}
& \nonumber \abs{ \phi^{-1} \rb{ \frac{ \partial \tarfunci{i} }{ \partial \vecx{i} } }  - \pnit{i}{*} } \leq \underbrace{ \abs{ \phi^{-1} \rb{ \frac{ \partial \tarfunci{i} }{ \partial \vecx{i} } } -  \E_{ \initam{i}{r} \sim \N \rb{0, \epsilon_a^2} , \initwm{i}{r}, \initbm{i}{r} \sim \N \rb{0, \frac{1}{m}} } \sqb{ \pnilt{\ell}{i}{*} } } }_{\RN{1}} \\
\nonumber  &+ \underbrace{ \abs{  \E_{ \initam{i}{r} \sim \N \rb{0, \epsilon_a^2} , \initwm{i}{r}, \initbm{i}{r} \sim \N \rb{0, \frac{1}{m}} } \sqb{ \pnilt{\ell}{i}{*} } - \E_{ \initam{i}{r} \sim \N \rb{0, \epsilon_a^2} , \initwm{i}{r}, \initbm{i}{r} \sim \N \rb{0, \frac{1}{m}} } \sqb{ \pnit{i}{*} } } }_{\RN{2}} \\
\nonumber  &+ \underbrace{ \abs{ \E_{ \initam{i}{r} \sim \N \rb{0, \epsilon_a^2} , \initwm{i}{r}, \initbm{i}{r} \sim \N \rb{0, \frac{1}{m}} } \sqb{ \pnit{i}{*} } - \E_{ \initam{i}{r} \sim \N \rb{0, \epsilon_a^2} , \initwm{i}{r}, \initbm{i}{r} \sim \N \rb{0, \frac{1}{m}} } \sqb{ \nnit{i}{*} } } }_{\RN{3}} \\
\nonumber &+ \underbrace{ \abs{ \E_{ \initam{i}{r} \sim \N \rb{0, \epsilon_a^2} , \initwm{i}{r}, \initbm{i}{r} \sim \N \rb{0, \frac{1}{m}} } \sqb{ \nnit{i}{*} } - \nnit{i}{*} } }_{\RN{4}} \\
&+ \underbrace{ \abs{ \nnit{i}{*} - \pnit{i}{*} } }_{\RN{5}}.
\end{align}
We know that the first part $\RN{1} \leq p_i \epsilon$ from Lemma \ref{lemma:expect-pseudo-target}. Since $\E_{ \initam{i}{r} \sim \N \rb{0, \epsilon_a^2} , \initwm{i}{r}, \initbm{i}{r} \sim \N \rb{0, \frac{1}{m}} } \sqb{ \pnilt{c}{i}{*} } = 0$, the second term $\RN{2} = 0$. Using Lemma \ref{lemma:bound-change-in-act-patterns} and Lemma \ref{lemma:coupling-f-prime-and-g-prime} for bounding the third term $\RN{3}$, we get
\begin{align}
\label{eq:bound-on-three}
\nonumber \RN{3} = & \; \abs{ \E_{ \initam{i}{r} \sim \N \rb{0, \epsilon_a^2} , \initwm{i}{r}, \initbm{i}{r} \sim \N \rb{0, \frac{1}{m}} } \sqb{ \pnit{i}{*} } - \E_{ \initam{i}{r} \sim \N \rb{0, \epsilon_a^2} , \initwm{i}{r}, \initbm{i}{r} \sim \N \rb{0, \frac{1}{m}} } \sqb{ \nnit{i}{*} } } \\  
\nonumber =& \; \abs{ \E_{ \initam{i}{r} \sim \N \rb{0, \epsilon_a^2} , \initwm{i}{r}, \initbm{i}{r} \sim \N \rb{0, \frac{1}{m}} } \sqb{ \pnit{i}{*} - \nnit{i}{*} } } \\
\nonumber \leq &  \; \E_{ \initam{i}{r} \sim \N \rb{0, \epsilon_a^2} , \initwm{i}{r}, \initbm{i}{r} \sim \N \rb{0, \frac{1}{m}} } \sqb{ \abs{ \pnit{i}{*} - \nnit{i}{*} } } \\ 
\nonumber \leq & \; \E_{ \initam{i}{r} \sim \N \rb{0, \epsilon_a^2} , \initwm{i}{r}, \initbm{i}{r} \sim \N \rb{0, \frac{1}{m}} } \sqb{ 24 c_1 \epsilon_a U_{ w_{i}^* } \abs{ \overline{\mathcal{H}}_i } \sqrt{ 2 \log m } } \\
 \nonumber \leq & \; 24 c_1 \epsilon_a U_{ w_{i}^* } \rb{ c_4 m \frac{ 4 U_{ w_{i}^* } \sqrt{m} }{\sqrt{ \pi }} } \sqrt{ 2 \log m }  \\
= & \; \frac{ 96 c_1 c_4 \epsilon_a m^{1.5} U_{ w_{i}^* }^2 \sqrt{ 2 \log m } }{ \sqrt{\pi} }.
\end{align}

We will use technique from \cite{yehudai2019power} to bound the fourth term $\RN{4}$. Define a function $\mathbf{N}_i$ as
\begin{align*}
\mathbf{N}_i &= \mathbf{N}_i \rb{ \rb{ \initam{i}{1}, \initwm{i}{1}, \initbm{i}{1} }, \ldots, \rb{ \initam{i}{m}, \initwm{i}{m}, \initbm{i}{m} } } = \sup_x \abs{ \E_{ \initam{i}{r} \sim \N \rb{0, \epsilon_a^2} , \initwm{i}{r}, \initbm{i}{r} \sim \N \rb{0, \frac{1}{m}} } \sqb{ \nnit{i}{*} } - \nnit{i}{*} }.
\end{align*}
We will now bound the expectation of $\mathbf{N}_i$ using McDiarmid's inequality (Fact \ref{fact:mcdiarmid-inequality}). For every $1 \leq r \leq m$, we get
\begin{align*}
&\Bigg| \mathbf{N}_i \rb{ \rb{ \initam{i}{1}, \initwm{i}{1}, \initbm{i}{1} } \ldots \rb{ \initam{i}{r}, \initwm{i}{r}, \initbm{i}{r} } \ldots \rb{ \initam{i}{m}, \initwm{i}{m}, \initbm{i}{m} } } \\ 
&- \mathbf{N}_i \rb{ \rb{ \initam{i}{1}, \initwm{i}{1}, \initbm{i}{1} } \ldots \rb{ \initam{i}{r}', \initwm{i}{r}', \initbm{i}{r}' } \ldots  \rb{ \initam{i}{m}, \initwm{i}{m}, \initbm{i}{m} } } \Bigg| \\
=& \sup_x \abs{ \initam{i}{r} \sigma \rb{  \dotp{ \initwm{i}{r} + w_{i,r}^* }{\nrmxi{i} }  + \rb{ \initbm{i}{r} + b_{i,r}^* }  } - \initam{i}{r}' \sigma \rb{  \dotp{ \initwm{i}{r}' + w_{i,r}^*}{\nrmxi{i} }  + \rb{ \initbm{i}{r}' + b_{i,r}^* }  }  } \\
=& \rb{ 2 c_1 \epsilon_a \sqrt{2 \log m } } \rb{ \frac{2 \rb{c_2 + c_3 }\sqrt{2 \log m} }{m} + 2 U_{ w_{i}^* }  },
\end{align*}
where last inequality follows with atleast $1 - \frac{1}{c_1} - \frac{1}{c_2} - \frac{1}{c_3}$ probability by applying Lemma \ref{lemma:concentration-folded-normal} on $\sqb{ \initam{i}{r} }_{r=1}^m, \sqb{ \initwm{i}{r} }_{r=1}^m$ and $\sqb{ \initbm{i}{r} }_{r=1}^m$. Define $\mathbf{C}_i$ as 
\begin{align*}
\mathbf{C}_i = \rb{ 2 c_1 \epsilon_a \sqrt{2 \log m } } \rb{ \frac{2 \rb{c_2 + c_3 }\sqrt{2 \log m} }{m} + 2 U_{ w_{i}^* }  }
\end{align*}
Using Lemma 26.2 from \cite{shalev2014understanding}, we get
\begin{align*}
\E_{ \initam{i}{r}, \initwm{i}{r}, \initbm{i}{r} } \sqb{ \mathbf{N}_i } \leq \frac{2}{m} \E_{ \initam{i}{r}, \initwm{i}{r}, \initbm{i}{r} } \sqb{ \sup_x \abs{ \sum_{r=1}^m \xi_r \initam{i}{r} \sigma \rb{ \neurvalm{i}{*} } } }
\end{align*}
where $\xi_1, \xi_2, \ldots, \xi_m$ are independent Rademacher random variables. Using Lipschitz continuity of ReLU activation, we get 
\begin{align*}
\E_{ \initam{i}{r}, \initwm{i}{r}, \initbm{i}{r} } \sqb{ \mathbf{N}_i } \leq \frac{2}{m} \E_{ \initam{i}{r}, \initwm{i}{r}, \initbm{i}{r} } \sqb{ \sup_x \abs{ \sum_{r=1}^m \xi_r \initam{i}{r} \rb{ \neurvalm{i}{*}{\nrmxi{i}} } } }
\end{align*}
Using Lemma 26.10 from \cite{shalev2014understanding}, we get
\begin{align*}
\E_{ \initam{i}{r}, \initwm{i}{r}, \initbm{i}{r} } \sqb{ \mathbf{N}_i } \leq & \; \E_{ \initam{i}{r}, \initwm{i}{r}, \initbm{i}{r} } \sqb{ \frac{ \max_{r \in [m]} \| a_{i, r} \rb{ \initwm{i}{r} + w_{i, r}^* }  \|_2 }{ \sqrt{m} }  } \\ 
&+ 2 \E_{ \initam{i}{r}, \initwm{i}{r}, \initbm{i}{r} } \sqb{ \frac{ \max_{r \in [m]} \| a_{i, r} \rb{ \initbm{i}{r} + b_{i, r}^* }  \|_2 }{ \sqrt{m} }  } \\
\leq & \rb{ 2 \frac{ \rb{ 2 c_1 \epsilon_a \sqrt{2 \log m}} }{ \sqrt{m} } \rb{ \frac{2 c_2 \sqrt{2 \log m}}{ \sqrt{m} } + U_{w_{i, r}^*} }   } + 2 \frac{ \rb{ 2 c_1 \epsilon_a \sqrt{2 \log m}} }{ \sqrt{m} } \frac{2 c_3 \sqrt{2 \log m}}{ \sqrt{m} }. 
\end{align*}
For $m \geq \Omega \rb{ \frac{ d^{10} \rb{ \sum_{i=1}^d  \sum_{r=1}^{p_i} U_{h_{i, r}} }^{12} }{ \epsilon_a^2 \epsilon^8 } }$, we have $U_{w_{i, r}^*} \leq  \frac{\rb{c_2 + c_3} \sqrt{2 \log m}}{ \sqrt{m} }$ and therefore, we get
\begin{align*}
\E_{ \initam{i}{r}, \initwm{i}{r}, \initbm{i}{r} } \sqb{ \mathbf{N}_i } \leq \frac{24 c_1 \rb{c_2 + c_3} \epsilon_a \log m  }{\sqrt{m}}.
\end{align*}
Using McDiarmid's inequality (Fact \ref{fact:mcdiarmid-inequality}), we get
\begin{align*}
\Pr \rb{ \textbf{N}_i -  \frac{24 c_1 \rb{c_2 + c_3} \epsilon_a \log m  }{\sqrt{m}} \geq \frac{\epsilon}{2} } \leq \Pr \rb{ \textbf{N}_i -  \E \sqb{\textbf{N}_i} \geq \frac{\epsilon }{2} } \leq \exp \rb{ - \frac{ \epsilon^2}{2 m C_i^2} }
\end{align*}
For $m \geq \Omega \rb{ \frac{ d^{10} \rb{ \sum_{i=1}^d  \sum_{r=1}^{p_i} U_{h_{i, r}} }^{12} }{ \epsilon_a^2 \epsilon^8 } }$, with at least $1 - \frac{1}{c_1} - \frac{1}{c_2} - \frac{1}{c_3} - \exp \rb{ - \frac{ \epsilon^2}{2 m C_i^2} }$ probability, for all $x$ with $\norm{x}_2 \leq 1$, we have
\begin{align}
\label{eq:bound-eq-four}
\abs{ \E_{ \initam{i}{r}, \initwm{i}{r}, \initbm{i}{r} } \sqb{ \nnit{i}{*} } - \nnit{i}{*} } \leq \epsilon
\end{align}
To bound $\RN{5}$, by Eq. (\ref{eq:coupling-N-P}), we know
\begin{align}
\label{eq:bound-eq-five}
\nonumber \RN{5} =& \abs{ \nnit{i}{*} - \pnit{i}{*} } \\ 
\nonumber \stackrel{ (\RomanNumLow{1}) }{ \leq } & 24 c_1 \epsilon_a U_{ w_{i, r}^* } \abs{ \overline{\mathcal{H}}_i } \sqrt{ 2 \log m } \\ 
\nonumber \stackrel{ (\RomanNumLow{2}) }{ \leq } & \; 24 c_1 \epsilon_a U_{ w_{i, r}^* } \rb{ c_4 m \frac{ 4 U_{ w_{i, r}^* } \sqrt{m} }{\sqrt{ \pi }} } \sqrt{ 2 \log m }  \\
= & \; \frac{ 96 c_1 c_4 \epsilon_a m^{1.5} U_{ w_{i, r}^* }^2 \sqrt{ 2 \log m } }{ \sqrt{\pi} },
\end{align}
where inequality $(\RomanNumLow{1})$ follows from Eq. (\ref{eq:coupling-N-P}) with atleast $1 - \frac{1}{c_1}$ probability and inequality $(\RomanNumLow{2})$ follows from Lemma \ref{lemma:bound-change-in-act-patterns} with atleast $1 - \frac{1}{c_1} - \exp \rb{ - \frac{ 32 (c_4 - 1)^2  m^2 U_{ w_{i, r}^* }^2 }{ \pi }  }$. Using Lemma \ref{lemma:expect-pseudo-target}, Eq.(\ref{eq:divide-approx-using-pseudo-network}), Eq.(\ref{eq:bound-on-three}), Eq.(\ref{eq:bound-eq-four}) and Eq.(\ref{eq:bound-eq-five}), with atleast $1 - \frac{1}{c_1} - \frac{1}{c_2} - \frac{1}{c_3} - \exp \rb{ - \frac{ \epsilon^2}{2 m C_i^2} } - \exp \rb{ - \frac{ 32 (c_4 - 1)^2  m^2 U_{ w_{i, r}^* }^2 }{ \pi }  } $ probability, we get
\begin{align*}
\abs{ \phi^{-1} \rb{ \frac{ \partial \tarfunci{i} }{ \partial \vecx{i} } }  - \pnit{i}{*} } &\leq p_i \epsilon + \frac{ 96 c_1 c_4 \epsilon_a m^{1.5} U_{ w_{i, r}^* }^2 \sqrt{ 2 \log m } }{ \sqrt{\pi} } + \epsilon + \frac{ 96 c_1 c_4 \epsilon_a m^{1.5} U_{ w_{i, r}^* }^2 \sqrt{ 2 \log m } }{ \sqrt{\pi} } \\
&= \rb{ p_i + 1} \epsilon + \frac{ 192 c_1 c_4 \epsilon_a m^{1.5} U_{ w_{i, r}^* }^2 \sqrt{ 2 \log m } }{ \sqrt{\pi} }.
\end{align*}
\end{proof}

\begin{lemma} 
\label{lemma:optimal-loss-by-opt-pseudo-net}
For any $\epsilon \in (0, 1)$, for any target function $\tarfunc{*}$, for any $m \geq \Omega \rb{ \frac{ d^{10} \rb{ \sum_{i=1}^d  \sum_{r=1}^{p_i} U_{h_{i, r}} }^{12} }{ \epsilon_a^2 \epsilon^8 } }$ and for any $x$ with $\norm{x}_2 \leq 1$, there exist a set of parameters $\theta^* =\rb{ \theta_1^*, \theta_2^*, \ldots, \theta_d^* }$ such that, with atleast $1 - \frac{d}{c_1} - \frac{d}{c_2} - \frac{d}{c_3} - d \exp \rb{ - \frac{ \epsilon^2}{2 m C_i^2} } - d \exp \rb{ - \frac{ 32 (c_4 - 1)^2  m^2 U_{ w_{i}^* }^2 }{ \pi }  } $ probability, we get
\begin{align*}
\abs{ \alm{ g^* }{x} - \alm{ \tarfunc{*} }{x} } \leq 3 \rb{ \sum_{i=1}^d p_i + d} \epsilon + \frac{ 576 c_1 c_4 \epsilon_a m^{1.5} \sqrt{ 2 \log m } }{ \sqrt{\pi} } \rb{ \sum_{i=1}^d U_{ w_{i}^* }^2 }.
\end{align*}
\end{lemma}

\begin{proof} Using definition of $\al$, we get
\begin{align*}
    \abs{ \alm{ g^* }{x} - \alm{ \tarfunc{*} }{x} } & \leq  \;  \abs{  \sum_{i=1}^{d} \sum_{j=1}^Q \Delta_x \rb{ \dgqpit{j}{i}{*} } -   \sum_{i=1}^{d} \sum_{j=1}^Q \Delta_x \rb{ \nabla_i \tarfunc{*}_i \rb{ \qpi{j}{i} }  }  } \\
    & + \abs{ \sum_{i=1}^d \log \rb{ \dgit{i}{*} } - \sum_{i=1}^d \log \rb{ \nabla_i \tarfunc{*}_i \rb{ \vecx{i} } } } \\
    \leq & \; \sum_{i=1}^{d} \sum_{j=1}^Q \Delta_x \abs{ \phi \rb{ P \rb{ \qpi{j}{i}, \theta^*_i } } -  \rb{ \nabla_i \tarfunc{*}_i \rb{ \qpi{j}{i} }  } } \\
    & + \sum_{i=1}^d \abs{ \log \rb{ \dgit{i}{*} } - \log \rb{ \nabla_i \tarfunc{*}_i \rb{ \vecx{i} } } } \\
    \stackrel{ ( \RomanNumLow{1} ) }{ \leq }  &  \; \sum_{i=1}^{d} \sum_{j=1}^Q \Delta_x \abs{ P \rb{ \qpi{j}{i}, \theta^*_i } -  \phi^{-1} \rb{ \nabla_i \tarfunc{*}_i \rb{ \qpi{j}{i} }  } }\\ 
    &+ \sum_{i=1}^d \abs{ \pnit{i}{*} - \phi^{-1} \rb{ \nabla_i \tarfunc{*}_i \rb{ \vecx{i} } } } \\
    \leq & \;  2 \rb{ \sum_{i=1}^d p_i + d} \epsilon + \frac{ 384 c_1 c_4 \epsilon_a m^{1.5} \sqrt{ 2 \log m } }{ \sqrt{\pi} } \rb{ \sum_{i=1}^d U_{ w_{i}^* }^2 } \\
    &+ \rb{ \sum_{i=1}^d p_i + d} \epsilon + \frac{ 192 c_1 c_4 \epsilon_a m^{1.5} \sqrt{ 2 \log m } }{ \sqrt{\pi} } \rb{ \sum_{i=1}^d U_{ w_{i}^* }^2 } \\
     \leq & \; 3 \rb{ \sum_{i=1}^d p_i + d} \epsilon + \frac{ 576 c_1 c_4 \epsilon_a m^{1.5} \sqrt{ 2 \log m } }{ \sqrt{\pi} } \rb{ \sum_{i=1}^d U_{ w_{i}^* }^2 },
\end{align*}
where inequality $(\RomanNumLow{1})$ follows from 1-Lipschitz continuity of $\phi( \cdot )$ and $\log \rb{ \phi \rb{ \cdot }  }$. The upper bound on $\norm{ \theta^* }_{2, \infty}$ is given by
\begin{align*}
\norm{ \theta^* }_{2, \infty} \leq \sum_{i=1}^d \norm{ \theta^*_i }_{2, \infty} \leq \sum_{i=1}^d \frac{ \sqrt{\pi} \rb{ \sum_{r=1}^{p_i} U_{ \omega_{i, r}} } }{m \epsilon_a \sqrt{2}} = \frac{ \sqrt{\pi} \rb{ \sum_{i=1}^d  \sum_{r=1}^{p_i} U_{\omega_{i, r}} } }{m \epsilon_a \sqrt{2}}.
\end{align*}
\end{proof}
We define upper bound on $\norm{ \theta^* }_{2, \infty}$ as $U_{\theta^*}$:
\begin{align*}
U_{ \theta^* } = \frac{ \sqrt{\pi} \rb{ \sum_{i=1}^d  \sum_{r=1}^{p_i} U_{\omega_{i, r}} } }{m \epsilon_a \sqrt{2}}.
\end{align*}

\section{Optimization}
\label{sec:optimization}

This section shows that SGD on the loss of the neural network can be closely approximated by the SGD on the loss of the pseudo-network (Theorem \ref{theorem:optimization-empirical-approximated-L}). Since the loss function of the pseudo-network is convex in its parameters (Lemma \ref{lemma:convex-pseudo-network-loss}), we get global optimization of the pseudo network, and hence, global optimization of the neural network. Moreover, there exist a pseudo-network which can approximation the target function and achieve training loss close to the trainign loss of the target function (Section \ref{sec:existence}). Therefore, SGD on the loss of the neural network can achieve training loss comparable to training loss of the target function (Theorem \ref{theorem:optimization-empirical-approximated-L}).

First, we will start with proving convexity of the loss function of the pseudo-network. 
\begin{lemma} \label{lemma:convex-pseudo-network-loss} (Convexity of the loss function of the pseudo-network) The loss function of the pseudo-network is convex with respect to the parameters of the neural network, and therefore, loss $\al$ satisfies first order condition of convexity for all $t \in [T]$ and for all $x$ with $\norm{x}_2 \leq 1$:
\begin{align*}
\al ( \dfv{ \gt{t} }, \mcx) - \al ( \nabla g^*, \mcx) \leq & \; \langle \nabla_{\theta} \al ( \dfv{ \gt{t} } , \mcx), \tht{t} - \theta^* \rangle .
\end{align*}
\end{lemma}

\begin{proof}
We decompose the loss function of the pseudo-network for each dimension into two parts:
\begin{align*}
    \alm{ \gt{t} }{ x } = \sum_{i=1}^d \rb{ \sum_{j=1}^{Q} \Delta_x \dgqpit{j}{i}{(t)} - \log \rb{ \dgit{i}{(t)} } } = \sum_{i=1}^d \rb{ \al_{i, 1} ( \nabla \gt{t}, x ) + \al_{i, 2} ( \nabla \gt{t} , x ) },
\end{align*}
where 
\begin{align*}
    \al_{i, 1} ( \nabla \gt{t} , x ) &=  \sum_{j=1}^{Q} \Delta_x \dfqpit{j}{i}{(t)} \hspace{1cm}  \text{and}  \hspace{1cm}  \al_{i, 2} ( \nabla \gt{t}, x ) &= - \log \rb{ \dgit{i}{(t)} }.
\end{align*}
We prove convexity of both $\al_{i, 1}( \nabla \gt{t}, x )$ and $\al_{i, 2} ( \nabla \gt{t}, x )$. We can write $\al_{i, 1} \rb{ \nabla \gt{t}, x }$ as
\begin{align*}
\al_{i, 1} \rb{ \dfv{\gt{t}}, x } = \sum_{j=1}^Q \Delta_x \phi \rb{ \pnit{i}{(t)} }.
\end{align*}
Note that $\phi \rb{ \pnit{i}{(t)} }$ is convex in $\pnit{i}{(t)}$ and $\pnit{i}{(t)}$ is linear in $\tht{t}_i$. As composition of any convex and linear function is convex, $\phi \rb{ \pnit{i}{(t)} }$ is convex. The first part of loss function $\al_{i, 1} \rb{ \dfv{\gt{t}}, x }$ is convex in $\tht{t}_i$ because sum of convex functions is also convex. By writing $\al_{i, 2} \rb{ \dfv{ \gt{t} }, x }$  in parts, we get
\begin{align*}
\al_{i, 2} \rb{ \dfv{ \gt{t} }, x } &= - \log \rb{ \phi \rb{ \pnit{i}{(t)} } } \\
&= -\log \rb{ \exp \rb{ \pnit{i}{(t)} } \ind{ \pnit{i}{(t)} \leq 0 } + \rb{ \pnit{i}{(t)} + 1 } \ind{ \pnit{i}{(t)} \geq 0 }   } \\ 
&= - \pnit{i}{(t)} \ind{ \pnit{i}{(t)} \leq 0 } - \log \rb{ \pnit{i}{(t)} + 1 }\ind{ \pnit{i}{(t)} \geq 0 }.
\end{align*}
Using last equality in the above equation, we can see that $\al_{i, 2}$ is convex in $\pnit{i}{(t)}$ and we know that $\pnit{i}{(t)}$ is linear in $\tht{t}_i$. Therefore, $\al_{i, 2}$ is convex in $\tht{t}_i$ because composition of any convex and linear function is a convex function. As $\al_{i, 1}$ and $\al_{i, 2}$ are convex, $\al$ is also convex in $\tht{t}_i$ because sum of convex functions is a convex function.
\end{proof}

\begin{remark} \label{rem:gaussian_convex} When we use the standard Gaussian for the base distribution, then the loss function will be:
\begin{align*}
\alm{ \gt{t} }{ x } &= \sum_{i=1}^d \rb{ \rb{ \sum_{j=1}^{Q} \Delta_x \dgqpit{j}{i}{(t)} }^2 - \log \rb{ \dgit{i}{(t)} } } \\ 
&= \sum_{i=1}^d \rb{ \al_{i, 1} ( \nabla \gt{t}, x ) + \al_{i, 2} ( \nabla \gt{t} , x ) }.
\end{align*}
Note that the second term in the decomposition $\al_{i, 2}$ is convex with same argument given in Lemma \ref{lemma:convex-pseudo-network-loss} and the first term $\al_{i, 1}$ is given by
\begin{align*}
    \al_{i, 1} \rb{ \nabla \gt{t}, x } = \rb{\sum_{j=1}^{Q} \Delta_x  \dgqpit{j}{i}{(t)}}^2.
\end{align*}
Using the same argument given in Lemma \ref{lemma:convex-pseudo-network-loss}, we get that $\sum_{j=1}^{Q} \Delta_x  \dgqpit{j}{i}{(t)}$ is convex in $\tht{t}_i$ but each summand in $\al_{i, 1}$ is square of convex function, which may not be convex in $\tht{t}_i$. Therefore, $\al_{i, 1}$ can be non-convex in $\tht{t}_i$.
\end{remark}

Recall that average loss of function $\ft{t}$ on training set $\mcx$ is defined as $\al \rb{ \dfv{ \ft{t} }, \mcx }$:
\begin{align*}
\al \rb{ \dfv{ \ft{t} }, \mcx } = \frac{1}{ | \mcx | } \sum_{ x \in \mcx } \al \rb{ \dfv{ \ft{t} }, x }.
\end{align*}
Similarly, average loss for $\gt{t}$ and average loss for $\tarfunc{*}$ is denoted by $\al \rb{ \dfv{ \gt{t} }, \mcx }$ and $\al \rb{ \dfv{ \tarfunc{*} }, \mcx }$, respectively.

\begin{theorem}
\label{theorem:optimization-empirical-approximated-L} 
(SGD achieves near-optimal loss) For every $\epsilon \in \rb{ 0, 1 }$, for every $m > \mathrm{poly}\rb{U_{\theta^*}, d, \frac{1}{\epsilon}}, $ learning rate $\eta=\Tilde{O} \rb{ \frac{1}{m \epsilon } }$ and number of steps $T=O \rb{ \frac{ U_{ \theta^* }^2 \log m }{ \epsilon^2 } }$ such that, with at least $0.94$ probability, we get
\begin{align*}
    \frac{1}{T} \sum_{t=0}^{T-1} \E_{\mathrm{sgd}} & [ \al( \dfv{ \ft{t} } , \mcx) ] - \al( \dfv{ \tarfunc{*} } , \mcx) \leq O( \epsilon ).
\end{align*}

\end{theorem}
\begin{proof} Recall that $\nabla g^*$ is a pseudo network which approximates the target function $\nabla F^*$. From Lemma \ref{lemma:convex-pseudo-network-loss}, we know that $\al  (\dfv{ \gt{t} }, \mcx)$ is convex in parameters $\theta$, which gives \todo{was $g^*$ defined?}
\begin{align}
\label{eq:diff-loss-time-t-optimal-inital}
    \nonumber \al ( \dfv{ \gt{t} }, \mcx) - \al ( \nabla g^*, \mcx) \leq & \; \langle \nabla_{\theta} \al ( \dfv{ \gt{t} } , \mcx), \tht{t} - \theta^* \rangle \\
    \nonumber \leq & \; \| \nabla_{\theta} \al (\dfv{ \gt{t} }, \mcx) - \nabla_{\theta} \al (\dfv{ \ft{t} }, \mcx) \|_{2, 1} \| \tht{t} - \theta^* \|_{2, \infty} \\ 
    &+ \langle \nabla_{\theta} \al ( \dfv{ \ft{t} } , \mcx), \tht{t} - \theta^* \rangle.
\end{align}
Recall that SGD update at time $t$ is given by 
\begin{align*}
\tht{t+1} = \tht{t} - \eta \nabla_{\theta} \al (\dfv{ \ft{t} }, \xt{t}).
\end{align*}
Using SGD update at time $t$, We have
\begin{align*}
    \| \tht{t+1} - \theta^* \|_{2, 2}^2 &= \| \tht{t} - \eta \nabla_{\theta} \al (\dfv{ \ft{t} }, \xt{t}) - \theta^* \|_{2, 2}^2 \\
    &= \| \tht{t} - \theta^* \|_{2, 2}^2 + \eta^2 \| \nabla_{\theta} \al (\dfv{ \ft{t} }, \xt{t}) \|_{2, 2}^2 - 2 \eta \langle \tht{t} - \theta^*, \nabla_{\theta} \al ( \dfv{ \ft{t} } , \xt{t}) \rangle.
\end{align*}
By taking expectation wrt $x_t$, we get
\begin{align}
\label{eq:exp-diff-theta-t-theta-star}
    \E_{\xt{t}} \left[ \| \tht{t+1} - \theta^* \|_{2, 2}^2 \right] = \| \tht{t} - \theta^* \|_{2, 2}^2 + \eta^2 \E_{\xt{t}} \left[ \| \nabla_{\theta} \al ( \dfv{ \ft{t} } , \xt{t}) \|_{2, 2}^2 \right] - 2 \eta \langle \nabla_{\theta} \al ( \dfv{ \ft{t} } , \mcx), \tht{t} - \theta^* \rangle.
\end{align}
Putting value of $\langle \nabla_{\theta} \al ( \dfv{ \ft{t} } , \mcx), \tht{t} - \theta^* \rangle$ from Eq.(\ref{eq:exp-diff-theta-t-theta-star}) to \eqref{eq:diff-loss-time-t-optimal-inital}, we get
\begin{align*}
    \al (\dfv{ \gt{t} }, \mcx) - \al (\nabla g^*, \mcx) \leq & \; \big{\|} \nabla_{\theta} \al (\dfv{ \gt{t} }, \mcx) - \nabla_{\theta} \al ( \dfv{ \ft{t} } , x) \big{\|}_{2, 1} \| \tht{t} - \theta^* \|_{2, \infty} \\  
    &+ \frac{ \| \tht{t} - \theta^*  \|_{2, 2}^2 - \E_{\xt{t}} \| \tht{t+1} - \theta^* \|_{2, 2}^2  }{ 2\eta } \\
    &+ \frac{ \eta }{ 2 } \E_{\xt{t}}  \| \nabla_{\theta} \al (\fpt{t}, \xt{t}) \|_{2, 2}^2 .
\end{align*}
By (\ref{eq:define-delta-bar}), (\ref{eq:bound-der-loss-wrt-w}) and (\ref{eq:bound-der-loss-wrt-b}), with atleast $1 - \frac{1}{c_1}$ probability, we have
\begin{align*}
    \Big{\|} \nabla_{\theta} \al ( \dfv{ \ft{t} } , \xt{t})  \Big{\|}_{2, 2}^2 \leq 2 m \LB^2.
\end{align*}
Averaging from $t=0$ to $T-1$, we get \todo{does this hold absolutely without any probability? Ans: Yes}
\begin{align*}
    \frac{1}{T} \sum_{t=0}^{T-1} \E_{\mathrm{sgd}} \sqb{ \al (\dfv{ \gt{t} }, \mcx) } - \al (\nabla g^*, \mcx) \leq & \; \frac{1}{T} \sum_{t=0}^{T-1} \sqb{ \big{\|} \nabla_{\theta} \al (\dfv{ \gt{t} }, \mcx) - \nabla_{\theta} \al ( \dfv{ \ft{t} } , x) \big{\|}_{2, 1} \| \tht{t} - \theta^* \|_{2, \infty} } \\  
    &+ \frac{ \| \tht{0} - \theta^* \|_{2, 2}^2 }{ 2 \eta T } \\
    &+ \frac{ \eta }{ 2 } \frac{1}{T} \sum_{t=0}^{T-1} \sqb{ \E_{\xt{t}}  \| \nabla_{\theta} \al (\fpt{t}, \xt{t}) \|_{2, 2}^2 } .
\end{align*}
\begin{align}
\label{eq:first-optim-eq}
    \nonumber \frac{1}{T} \sum_{t=0}^{T-1} \E_{\mathrm{sgd}} [ \al (\dfv{ \gt{t} }, \mcx) ] - \al ( \nabla g^* , \mcx) & \leq  \; \cgl \rb{  \sup_{t \in [T]} \| \tht{t} \|_{2, \infty} + \|  \theta^* \|_{2, \infty} } + \frac{ \| \tht{0} - \theta^* \|_{2, 2}^2 }{ 2 \eta T } + \eta  m \LB^2 \\
    & = \cgl \rb{  \sup_{t \in [T]} \| \tht{t} \|_{2, \infty} + \|  \theta^* \|_{2, \infty} } + \frac{ \|  \theta^* \|_{2, 2}^2 }{ 2 \eta T } + \eta  m \LB^2, 
\end{align}
where last inequality follows with atleast $1 - \frac{d}{c_1} - d \exp \rb{ - \frac{ 32 (c_4 - 1)^2 \eta^2 m^2 \LB^2 t^2 }{ \pi } }$. Recall that $\Gamma$ was defined in \eqref{eq:upper-bound-gradient-of-loss}. The last equality also uses the fact that initial change in weights $\tht{0}$ is equal to $\rb{0, 0, \ldots, 0}$. \todo{This may need to be clarified where $\tht{0}$ is defined. A similar point caused confusion to a reviewers also.} Using Lemmas \ref{lemma:coupling-loss} and \ref{lemma:optimal-loss-by-opt-pseudo-net} respectively, with 
probability at least $1 - \frac{d}{c_1}  - \frac{d}{c_2} - \frac{d}{c_3} - \sum_{t=1}^T d \exp \rb{ - \frac{ 32 (c_4 - 1)^2 \eta^2 m^2 \LB^2 t^2 }{ \pi } }  - d \exp \rb{ - \frac{ \epsilon^2}{2 m C_i^2} } - d \exp \rb{ - \frac{ 32 (c_4 - 1)^2  m^2 U_{ w_{i}^* }^2 }{ \pi }  } $ we have
\begin{align*}
    \frac{1}{T} \sum_{t=0}^{T-1} \E_{\mathrm{sgd}} & [ \al (\dfv{ \ft{t} }, \mcx) ] - \al (\nabla g^*, \mcx) \leq  \; \Gamma \rb{ \sup_{t \in [T]} \| \tht{t} \|_{2, \infty} + \| \theta^* \|_{2, \infty} } + \frac{ \| \theta^* \|_{2, 2}^2 }{ 2 \eta T } + \eta  m \LB^2 + 3 \dnp{t},  \\
    \frac{1}{T} \sum_{t=0}^{T-1} \E_{\mathrm{sgd}} & [ \al ( \dfv{ \ft{t} } , \mcx) ] - \al ( \nabla F^* , \mcx) \leq  \; \Gamma \rb{ \sup_{t \in [T]} \| \tht{t} \|_{2, \infty} + \| \theta^* \|_{2, \infty} } + \frac{ \| \theta^* \|_{2, 2}^2 }{ 2 \eta T } + \eta  m \LB^2 \\
    & + 3 \dnp{t} +3 \rb{ \sum_{i=1}^d p_i + d} \epsilon + \frac{ 576 c_1 c_4 \epsilon_a m^{1.5} \sqrt{ 2 \log m } }{ \sqrt{\pi} } \rb{ \sum_{i=1}^d U_{ w_{i}^* }^2 }.  
 \end{align*}
 We now choose values of $\eta$ and $T$: \todo{the second equality in the second equation below should be an inequality}
 \begin{equation}
 \label{eq:values-eta-T-m-td}
\begin{aligned}
    \eta &= \frac{ \epsilon }{ m \LB^2  } \\ 
    &= \frac{ \epsilon }{ m \rb{ 6 c_1 \epsilon_a \sqrt{2 \log m}  }^2 } \\
    &= \frac{\epsilon }{ 72 c_1^2 m \epsilon_a^2 \log m  }, \\
    T &:= \frac{ \| \theta^* \|_{2,2}^2 }{ 2 \eta \epsilon } \\ 
    & \leq  m U_{ \theta^* }^2 \frac{ 72 c_1^2 m \epsilon_a^2 \log m }{2 \epsilon^2} \\ 
    &= \frac{ 72 c_1^2 m^2 U_{ \theta^* }^2 \epsilon_a^2 \log m }{2 \epsilon^2} ,
\end{aligned}
\end{equation}
where we use chosen value of $\eta$ to get upper bound on $T$.
\todo{how was the second inequality in the second equation above obtained?}
Using above inequalities, we get the following equalities: \todo{The following equation is not well-formatted.}

\begin{align*}
    \frac{ \| \theta^* \|_{2, 2}^2 }{2 \eta T} &= \frac{ \| \theta^* \|_{2, 2}^2 }{2 \eta} \frac{ 2 \eta \epsilon }{ \| \theta^* \|_{2, 2}^2 } = \epsilon, \\
    \eta  m \LB^2 &=  \frac{ \epsilon }{ m \LB^2  } m \LB^2  = \epsilon.
\end{align*}

Using Lemma~\ref{lemma:optimal-loss-by-opt-pseudo-net}, we get
\begin{align*}
    \| \theta^* \|_{2, \infty} &\leq U_{ \theta^* }, \\
    \| \theta^* \|_{2, 2} &\leq \sqrt{m} \| \theta^* \|_{2, \infty} = \sqrt{m} U_{ \theta^* }.
\end{align*}
To get value of $m$, we will first upper bound $\sup_{t \in [T]} \| \tht{t} \|_{\infty}$, $\sup_{t \in [T]} \| \tht{t} \|_{\infty} + \| \theta^* \|_{\infty} $  and $\Gamma$:
\begin{align*}
    \sup_{t \in [T]} \| \tht{t} \|_{\infty} &= \sup_{t \in [T]} \eta \LB t = \eta \LB T = \frac{ \| \theta^* \|_2^2 \LB }{ 2 \epsilon } \leq m U_{\theta^*}^2 \frac{ \rb{ 6 c_1 \epsilon_a \sqrt{2 \log m}  } }{2 \epsilon} \\
    &= \frac{ \rb{ 3 c_1 m U_{\theta^*}^2 \epsilon_a \sqrt{2 \log m}  } }{ \epsilon} \\
    \sup_{t \in [T]} \| \tht{t} \|_{\infty} + \| \theta^* \|_{\infty} &\leq \frac{ \rb{ 3 c_1 m U_{\theta^*}^2 \epsilon_a \sqrt{2 \log m}  } }{ \epsilon} + U_{\theta^*}^2  \leq \frac{ \rb{ \rb{1 + 3 c_1} m U_{\theta^*}^2 \epsilon_a \sqrt{2 \log m}  } }{ \epsilon} \\
    \Gamma =& \; \frac{ 192 d \eta m^{1.5} \LB c_1 c_4 \epsilon_a t \sqrt{ \log m } }{ \sqrt{\pi} } + 24 c_1 d \epsilon_a m \dnp{t} \sqrt{2 \log m} \\
     \leq & \; \frac{ 192 d \eta m^{1.5} \LB c_1 c_4 \epsilon_a t \sqrt{ \log m } }{ \sqrt{\pi} } + 24 c_1 d \epsilon_a m \sqrt{2 \log m} \rb{ \frac{ 192 \eta^2 m^{1.5} \LB^2 c_1 c_4 \epsilon_a t^2 \sqrt{ \log m } }{ \sqrt{\pi } } } \\
     \leq &\frac{ 192 d \eta m^{1.5} \LB c_1 c_4 \epsilon_a t \sqrt{ \log m } }{ \sqrt{\pi} } + \frac{ 4608 \sqrt{2} c_1^2 c_4 d \epsilon_a^2 \eta^2 t^2 m^{2.5} \log m \LB^2 }{ \sqrt{\pi} }\\
     \leq & \; \frac{ 192 d m^{1.5} c_1 c_4 \epsilon_a \sqrt{ \log m } }{ \sqrt{\pi} } \rb{ \frac{ m U_{\theta^*}^2 }{ 2 \epsilon } } \rb{ 6 c_1 \epsilon_a \sqrt{2 \log m} } \\ 
     &+ \frac{ 4608 \sqrt{2} c_1^2 c_4 d \epsilon_a^2 m^{2.5} \log m }{ \sqrt{\pi} } \rb{ \frac{ m U_{\theta^*}^2 }{ 2 \epsilon } }^2 \rb{ 6 c_1 \epsilon_a \sqrt{2 \log m} }^2 \\
     \leq & \; \frac{ 576 \sqrt{2} d m^{2.5} c_1^2 c_4 \epsilon_a^2 U_{\theta^*}^2 \log m  }{ \epsilon \sqrt{\pi} } + \frac{ 82944 \sqrt{2} c_1^4 c_4 d \epsilon_a^4 m^{4.5 } U_{\theta^*}^4 \rb{ \log m }^2 }{ \sqrt{\pi} \epsilon^2 } \\
     \leq & \; \frac{ 165888 \sqrt{2} c_1^4 c_4 d \epsilon_a^4 m^{4.5 } U_{\theta^*}^4 \rb{ \log m }^2 }{ \sqrt{\pi} \epsilon^2 }.
\end{align*}
Multiplication of $\Gamma$ and $\rb{ \sup_{t \in [T]} \| \tht{t} \|_{\infty} + \| \theta^* \|_{\infty} }$ will be
\begin{align*}
    \Gamma \rb{ \sup_{t \in [T]} \| \tht{t} \|_{\infty} + \| \theta^* \|_{\infty} } &\leq \frac{ 165888 \sqrt{2} c_1^4 c_4 d \epsilon_a^4 m^{4.5 } U_{\theta^*}^4 \rb{ \log m }^2 }{ \sqrt{\pi} \epsilon^2 } \rb{ \frac{ \rb{ \rb{1 + 3 c_1} m U_{\theta^*}^2 \epsilon_a \sqrt{2 \log m}  } }{ \epsilon}  } \\
    &= \frac{ 331776 c_1^4 \rb{1 + 3 c_1} c_4 d \epsilon_a^5 m^{5.5 } U_{\theta^*}^6 \rb{ \log m }^{2.5} }{ \sqrt{\pi} \epsilon^3 } \\
    &= \frac{ 331776 c_1^4 \rb{1 + 3 c_1} c_4 d \epsilon_a^5 m^{5.5 } \rb{ \log m }^{2.5} }{ \sqrt{\pi} \epsilon^3 } \rb{ \frac{ \sqrt{\pi} \rb{ \sum_{i=1}^d  \sum_{r=1}^{p_i} U_{h_{i, r}} } }{m \epsilon_a \sqrt{2}} }^6 \\ 
    &= \frac{ 41472 \pi^{2.5} c_1^4 \rb{1 + 3 c_1} c_4 d \rb{ \log m }^{2.5} \rb{ \sum_{i=1}^d  \sum_{r=1}^{p_i} U_{h_{i, r}} }^6 }{ \sqrt{m} \epsilon^3 \epsilon_a }.
\end{align*}
Taking $m$ as 
\begin{align}
\label{eq:lower-bound-on-m}
    m \geq \Omega \rb{ \frac{ c_1^8 c_4^2 d^2 \rb{ 1 + 3c_1 }^2 \rb{ \sum_{i=1}^d  \sum_{r=1}^{p_i} U_{h_{i, r}} }^{12} }{ \epsilon_a^2 \epsilon^8 } },
\end{align}
we get
\begin{align*}
    \Gamma \rb{ \sup_{t \in [T]} \| \tht{t} \|_{\infty} + \| \theta^* \|_{\infty} } \leq \epsilon.
\end{align*}
Using \eqref{eqn:dnpt}, we get
\begin{align}
\label{eq:upper-bound-on-delta-np-t}
    \nonumber \dnp{t} &= \rb{ \frac{ 192 \eta^2 m^{1.5} \LB^2 c_1 c_4 \epsilon_a t^2 \sqrt{ \log m } }{ \sqrt{\pi } } }  \\ 
    \nonumber &\leq \rb{ \frac{ 192 m^{1.5} c_1 c_4 \epsilon_a \sqrt{ \log m } }{ \sqrt{\pi } } } \rb{ \frac{ m U_{\theta^*}^2 }{ 2 \epsilon } }^2 \rb{ 6 c_1 \epsilon_a \sqrt{2 \log m} }^2 \\
    &= \frac{ 3456 m^{3.5} c_1^3 c_4 \epsilon_a^3 U_{\theta^*}^4 \rb{ \log m }^{1.5} }{ \epsilon^2 \sqrt{\pi } }.
\end{align}
Using given choice of $m$ from \eqref{eq:lower-bound-on-m}, we get 
\begin{align*}
    \dnp{t} & \leq \frac{ 3456 m^{3.5} c_1^3 c_4 \epsilon_a^3 \rb{ \log m }^{1.5} }{ \epsilon^2 \sqrt{\pi } } \rb{ \frac{ \sqrt{\pi} \rb{ \sum_{i=1}^d  \sum_{r=1}^{p_i} U_{h_{i, r}} } }{m \epsilon_a \sqrt{2}} }^4 \\
    & = \frac{ 864 \pi^{1.5}  c_1^3 c_4 \rb{ \log m }^{1.5} }{ \epsilon_a \epsilon^2 }  \rb{ \sum_{i=1}^d  \sum_{r=1}^{p_i} U_{h_{i, r}} }^4 \rb{ \frac{ \epsilon_a^2 \epsilon^8  }{ c_1^8 c_4^2 d^2 \rb{ 1 + 3c_1 }^2 \rb{ \sum_{i=1}^d  \sum_{r=1}^{p_i} U_{h_{i, r}} }^{12} } }^{0.5} \\
    & = O \rb{ \frac{ \epsilon^2 \rb{ \log m }^{1.5} }{ c_1 \rb{ 1 + 3 c_1 } \rb{ \sum_{i=1}^d  \sum_{r=1}^{p_i} U_{h_{i, r}} }^2 } } \\
    & \leq O \rb{ \epsilon }.
\end{align*}
Similarly, using given choice of $m$ from \eqref{eq:lower-bound-on-m}, we get
\begin{align*}
\frac{ 576 c_1 c_4 \epsilon_a m^{1.5} \sqrt{ 2 \log m } }{ \sqrt{\pi} } \rb{ \sum_{i=1}^d U_{ w_{i}^* }^2 } &= \frac{ 576 c_1 c_4 \epsilon_a m^{1.5} \sqrt{ 2 \log m } }{ \sqrt{\pi} } \rb{ \sum_{i=1}^d \rb{ \frac{ \sqrt{\pi} \rb{ \sum_{r=1}^{p_i} U_{h_{i, r}} } }{m \epsilon_a \sqrt{2}}  }^2 } \\ 
&\leq \frac{ 288 \sqrt{\pi} c_1 c_4 \sqrt{ 2 \log m } }{ m^{0.5} \epsilon_a } \rb{ \sum_{i=1}^d  \sum_{r=1}^{p_i} U_{h_{i, r}} }^2 \\
&\leq O \rb{ \epsilon }.
\end{align*}
Using Eq.(\ref{eq:values-eta-T-m-td}) and Eq.(\ref{eq:lower-bound-on-m}), with at least $1 - \frac{d}{c_1}  - \frac{d}{c_2} - \frac{d}{c_3} - \sum_{t=1}^T d \exp \rb{ - \frac{ 32 (c_4 - 1)^2 \eta^2 m^2 \LB^2 t^2 }{ \pi } }  - d \exp \rb{ - \frac{ \epsilon^2}{2 m C_i^2} } - d \exp \rb{ - \frac{ 32 (c_4 - 1)^2  m^2 U_{ w_{i}^* }^2 }{ \pi }  } $ probability, we get
\begin{align*}
    \frac{1}{T} \sum_{t=0}^{T-1} \E_{\mathrm{sgd}} & [ \al ( \dfv{ \ft{t} } , \mcx) ] - \al ( \nabla F^* , \mcx) \leq  \; \Gamma \rb{ \sup_{t \in [T]} \| \tht{t} \|_{2, \infty} + \| \theta^* \|_{2, \infty} } + \frac{ \| \theta^* \|_{2, 2}^2 }{ 2 \eta T } + \eta  m \LB^2 \\
    & + 3 \dnp{t} +3 \rb{ \sum_{i=1}^d p_i + d} \epsilon + \frac{ 576 c_1 c_4 \epsilon_a m^{1.5} \sqrt{ 2 \log m } }{ \sqrt{\pi} } \rb{ \sum_{i=1}^d U_{ w_{i}^* }^2 } \\ 
    \leq & O ( \epsilon ) + 3 \rb{ \sum_{i=1}^d p_i + d} \epsilon .
\end{align*}

Taking $c_1=100d, c_2=100d, c_3=100d, c_4=d+1, \epsilon_a = \frac{\epsilon}{ 6000 \log m   } \leq \epsilon$ and rescaling $\epsilon$ as $\epsilon / \rb{ \sum_{i=1}^d p_i + d}$, with at least $0.97 - \sum_{t=1}^T d \exp \rb{ - \frac{ 32 d^2 \eta^2 m^2 \LB^2 t^2 }{ \pi } }  - d \exp \rb{ - \frac{ \epsilon^2}{2 m C_i^2} } - d \exp \rb{ - \frac{ 32 d^2  m^2 U_{ w_{i}^* }^2 }{ \pi }  }$ probability, we get
\begin{align*}
    \frac{1}{T} \sum_{t=0}^{T-1} \E_{\mathrm{sgd}} & [ \al (\dfv{ \ft{t} }, \mcx) ] - \al ( \nabla F^* , \mcx) \leq \;  O \rb{ \epsilon } .
\end{align*}
To find the lower bound on probability, we use $\sum_{t=1}^T \frac{1}{t^2} \leq \sum_{t=1}^{\infty} \frac{1}{t^2} \leq 2$:
\begin{align*}
    \sum_{t=1}^T d \exp \rb{ - \frac{ 32 d^2 \eta^2 m^2 \LB^2 t^2 }{ \pi } } & \stackrel{(\RomanNumLow{1})}{\leq}  \sum_{t=1}^T \frac{ d \pi }{ 32 d^2 \eta^2 m^2 \LB^2 t^2 }  
    = \frac{ \pi \LB^4 }{ 16 d \epsilon^2 \LB^2 }
    \leq \frac{ \pi \LB^2  }{ 16 d \epsilon^2 } 
    \leq \frac{ \pi  }{ 3200 }
    \leq 0.01.
\end{align*}
where inequality $(\RomanNumLow{1})$ follows from $\exp \rb{ -x } \leq \frac{1}{x}$ for all $x \geq 0$. To find lower bound on $d \exp \rb{ - \frac{ \epsilon^2}{2 m C_i^2} }$, we use same inequality:  
\begin{align*}
d \exp \rb{ - \frac{ \epsilon^2}{2 m C_i^2} } \leq \frac{2 d m C_i^2 }{ \epsilon^2 } = \frac{2 d m}{ \epsilon^2 } \rb{ 2 c_1 \epsilon_a \sqrt{2 \log m } } \rb{ \frac{2 \rb{c_2 + c_3 }\sqrt{2 \log m} }{m} + 2 \frac{ \sqrt{\pi} \rb{ \sum_{r=1}^{p_i} U_{h_{i, r}} } }{m \epsilon_a \sqrt{2}}  }^2 \leq 0.01
\end{align*}
where last inequality follows from given choice (Eq. \eqref{eq:lower-bound-on-m}) of sufficiently high $m$. Now, we will lower bound $d \exp \rb{ - \frac{ 32 d^2  m^2 U_{ w_{i}^* }^2 }{ \pi }  }$ quantity:
\begin{align*}
d \exp \rb{ - \frac{ 32 d^2  m^2 U_{ w_{i}^* }^2 }{ \pi }  } \leq \frac{\pi d}{ 32 d^2  m^2 U_{ w_{i}^* }^2 } = \frac{\pi d}{ 32 d^2  m^2 } \frac{ 2  m^2 \epsilon_a^2 }{ \pi \rb{ \sum_{r=1}^{p_i} U_{h_{i, r}} }^2 } = \frac{ \epsilon_a^2 }{ 16 d \rb{ \sum_{r=1}^{p_i} U_{h_{i, r}} }^2 } \leq 0.01 
\end{align*}
where last inequality follows from the value of $\epsilon_a$. Finally, we can say that, with at least $0.94$ probability, we get
\begin{align*}
    \frac{1}{T} \sum_{t=0}^{T-1} \E_{\mathrm{sgd}} & [ \al (\dfv{ \ft{t} }, \mcx) ] - \al ( \nabla F^*, \mcx) \leq O(\epsilon).
\end{align*}
\end{proof}

\section{Generalization}
\label{sec:generalization}
In this section, we prove generalization guarantees to complement our optimization result, and complete the proof of our main theorem (Theorem \ref{theorem:final-theorem}) about efficiently learning distributions using univariate normalizing flows. Recall that $\al ( \dfv{ \ft{t} }, \mcx )$  denotes an empirical average of $\al ( \dfv{ \ft{t} }, \mcx )$ over training data and $\al ( \dfv{ \ft{t} }, \mathcal{D} )$ denotes expectation with respect to underlying data distribution. The proof in this section can be broadly divided two parts. First, we prove that empirical average $\al ( \dfv{ \ft{t} }, \mcx )$ and $\al ( \dfv{ \tarfunc{*} }, \mcx )$ are close to expectation $\al ( \dfv{ \ft{t} }, \mathcal{D} )$ and $\al ( \dfv{ \tarfunc{*} }, \mathcal{D} )$, respectively (Lemma \ref{lemma:generalization-neural-net-func} and Lemma \ref{lemma:generalization-target-func}). 
Second, we prove that $\al ( \dfv{ \ft{t} }, \mathcal{D} )$ and $\al ( \dfv{ \tarfunc{*} }, \mathcal{D} )$ are close to $L ( \ft{t}, \mathcal{D} )$ and $L ( \tarfunc{*}, \mathcal{D} )$, respectively (Theorem \ref{theorem:final-theorem}).  \\[3pt]
Recall that the approximate loss function $\al$ is given by
\begin{align*}
    \al \rb{ \dfv{ \ft{t} }, x } &=  \sum_{i=1}^d \rb{ \sum_{j=1}^Q \Delta_x \phi\rb{ N \rb{ \qpi{j}{i}; \tht{t}_i } }   - \log\rb{ \phi\rb{ N \rb{ \vecx{i}; \tht{t}_i } } } },
\end{align*}
where
\begin{align*}
    N( \vecx{i}, \tht{t}_i ) = \sum_{r=1}^m \initam{i}{r} \sigma \rb{ \neurvalm{i}{(t)}{ \nrmxi{i} } }. 
\end{align*}
Similarly, we define $\alm{ \tarfunc{*} }{x}$ for the target function $\tarfunc{*}$.
\begin{lemma}
\label{lemma:empirical-rademacher-complexity-two-layer}
(Empirical Rademacher complexity for two-layer neural network) For every constant $B > 0$, for any number of training samples $n \geq 1$, for any time $t \geq 1$, with probability at least $1 - \frac{1}{c_1}$ over random initialization, the empirical Rademacher complexity is bounded by
\begin{align*}
    \frac{1}{n} \E_{ \xi \in \{ \pm  1 \}^n } \left[ \sup_{ \max_{ r \in [m] } \norm{ \wt{i, r}{t} }, \abs{ \bt{i, r}{t} } \leq B } \sum_{j=1}^n \xi_j N \rb{ \rb{ \vecx{i} }_j, \tht{t}_i } \right] \leq \frac{ 8 c_1 \epsilon_a B m \sqrt{2 \log m} }{ \sqrt{n} },
\end{align*}
where $\rb{ \vecx{i} }_j$ denotes first $i$ dimension of $j^{\text{th}}$ training example.

\end{lemma}
\begin{proof} Using part (\ref{it:linear-rademacher-complexity}) of Lemma \ref{lemma:rademacher-complexity-properties}, we get that $\{ x \mapsto \dotp{ \wt{i, r}{t} }{ \nrmxi{i} } + \bt{i, r}{t} \; | \; \norm{ \wt{i, r}{t} }_2 \leq B, \abs{ \bt{i, r}{t} } \leq B \}$ has Rademacher complexity $\frac{2 B}{ \sqrt{n} }$. Using part (\ref{it:addition-property-rademacher-complexity}) of Lemma \ref{lemma:rademacher-complexity-properties}, we get that $\{ x \mapsto \neurvalm{i}{(t)}{ \nrmxi{i} } \; | \; \norm{ \wt{i, r}{t} }_2 \leq B, \abs{ \bt{i, r}{t} } \leq B, \initwm{i}{r} \sim \N \rb{0, \frac{1}{m} \mathbf{1} } , \initbm{i}{r} \sim \N \rb{0, \frac{1}{m}} \}$ has Rademacher complexity $\frac{2 B}{ \sqrt{n} }$. Using part (\ref{it:neural-network-type-rademacher-complexity}) of Lemma \ref{lemma:rademacher-complexity-properties}, we get that class of functions in $\mcf = \{ x \mapsto \nnit{i}{(t)} \;   | \; \max_{ r \in [m] } \norm{ \wt{i, r}{t} }_2 \leq B, \max_{ r \in [m] } \abs{ \bt{i, r}{t} } \leq B   \}$ has Rademacher complexity
\begin{align*}
    \empR \rb{ \mcx; \mcf } \leq 2 \| \mathbf{a} \|_1 \frac{2 B}{ \sqrt{n} } \stackrel{(\RomanNumLow{1})}{\leq} \frac{ 8 c_1 \epsilon_a B m \sqrt{2 \log m} }{ \sqrt{n} },
\end{align*}
where inequality (\RomanNumLow{1}) follows from Lemma \ref{lemma:concentration-folded-normal} with at least $1 - \frac{1}{c_1}$ probability over random initialization.

\end{proof} 
We denote $M_{\nabla \tarfunc{*}}$ and $m_{\nabla \tarfunc{*}}$ as maximum and minimum value of $\nabla \tarfunc{*}$: 
\begin{align*}
M_{\nabla \tarfunc{*}} &= \max_{i \in [d], x \in \R^d} \nabla_i \tarfunci{i} \rb{ \vecx{i} }, \\
m_{\nabla \tarfunc{*}} &= \min_{i \in [d], x \in \R^d} \nabla_i \tarfunci{i} \rb{ \vecx{i} }.
\end{align*}
We find upper bound on maximum and lower bound on minimum value of the loss $\al$ for the target function $\tarfunc{*}$ in terms of $M_{\nabla \tarfunc{*}}$ and $m_{\nabla \tarfunc{*}}$:
\begin{align*}
    \sup_x \al \rb{ \dfv{ \tarfunc{*} }, x } &= \max_x \sum_{i=1}^d \rb{ \sum_{j=1}^{ Q } \Delta_x \rb{ \nabla_i \tarfunci{i} \rb{ \qpi{j}{i} } }   - \log \nabla_i \tarfunc{*}_i \rb{ \vecx{i} } } \leq 2 d M_{\nabla \tarfunc{*}} - d \log \rb{ m_{\nabla \tarfunc{*}} }, \\
    \inf_x \al \rb{ \dfv{ \tarfunc{*} }, x } &= \min_x   \sum_{i=1}^d \rb{ \sum_{j=1}^{ Q } \Delta_x \rb{ \nabla_i \tarfunci{i} \rb{ \qpi{j}{i} } }   - \log \nabla_i \tarfunc{*}_i \rb{ \vecx{i} } } \geq 2 d m_{\nabla \tarfunc{*}} - d \log \rb{ M_{\nabla \tarfunc{*}} },
\end{align*}
and define them respectively as $M_{\al}$ and $m_{\al}$:
\begin{equation}
\begin{aligned}
M_{\al} &= 2 d M_{\nabla \tarfunc{*}} - d \log \rb{ m_{\nabla \tarfunc{*}} }, \\
m_{\al} &= 2 d m_{\nabla \tarfunc{*}} - d \log \rb{ M_{\nabla \tarfunc{*}} }.
\end{aligned}
\end{equation}

\begin{lemma} \label{lemma:small-value-at-init} (Small value of neural network at initialization) For any dimension $i \in [d]$, for any constant $c_1 > 10, c_2 > 10$ and $c_3 > 10$, with probability at least $0.99 - \frac{1}{c_1} - \frac{1}{c_2} - \frac{1}{c_3}$, we have 
\begin{align*}
   & \abs{ \sum_{r=1}^m \initam{i}{r} \sigma \rb{ \neuralvalzerom{i} } } \leq 16 \sqrt{ \rb{d + 1} \log \rb{d + 1}} c_1 c_2 \epsilon_a \rb{ \log m} + 16 c_1 c_3 \epsilon_a \rb{ \log m}.
\end{align*}
\end{lemma}
\begin{proof}
Suppose, for any given $x$, there are $m'$ indicators with value 1. Without loss of generality, we can assume that indicators from $r=1$ to $r=m'$ is 1. Then,
\begin{align*}
    \abs{ \sum_{r=1}^m \initam{i}{r} \rb{ \neuralvalzerom{i} } \ind{ \neuralvalzerom{i} \geq 0 } } &= \abs{ \sum_{r=1}^{m'} \initam{i}{r} \rb{ \neuralvalzerom{i} } } \\
    &= \abs{ \dotp{ x } { \sum_{r=1}^{m'} \initam{i}{r} \initwm{i}{r} } + \sum_{r=1}^{m'} \initam{i}{r} \initbm{i}{r} } \\
\end{align*}
Now, applying Hoeffding's inequality (Fact \ref{fact:hoeffding-inequality}) on any dimension $j \in [d+1]$ for the sum in first part of the above equation, with atleast $1- \frac{1}{c_1} - \frac{1}{c_2}$ probability, we get
\begin{align}
\label{eq:bound-sum-ar0-wr0j}
    \nonumber \text{ Pr } \rb{ \abs{ \sum_{r=1}^{m'} \initam{i}{r} \initwm{i}{r, j} } \geq t } &\leq \exp \rb{ -\frac{ 2 t^2 m }{ m' \rb{ 2 c_1 \epsilon_a \sqrt{2 \log m} }^2 \rb{ 2 c_2 \sqrt{2 \log m} }^2 } } \\
    &\leq \exp \rb{ -\frac{ t^2 }{  32 c_1^2 c_2^2 \epsilon_a^2 \rb{ \log m}^2  } }. 
\end{align}
Using union bound, we get
\begin{align*}
\Pr \rb{ \bigcup_{j \in [d+1]} \rb{ \abs{ \sum_{r=1}^{m'} \initam{i}{r} \initwm{i}{r, j} } \geq t }  } \leq \rb{ d + 1 } \exp \rb{ -\frac{ t^2 }{  32 c_1^2 c_2^2 \epsilon_a^2 \rb{ \log m}^2  } }
\end{align*}
Using definition of $L_{\infty}-$norm, we have
\begin{align*}
\Pr \rb{  \norm{ \sum_{r=1}^{m'} \initam{i}{r} \initwm{i}{r} }_{\infty} \geq t  } \leq \rb{ d + 1 } \exp \rb{ -\frac{ t^2 }{  32 c_1^2 c_2^2 \epsilon_a^2 \rb{ \log m}^2  } }
\end{align*}
Plugging $t = 16 \sqrt{\log \rb{d + 1}} c_1 c_2 \epsilon_a \rb{ \log m}$ in above equation, with probability at least $1 - \exp \rb{ - 8 } - \frac{1}{c_1} - \frac{1}{c_2}$, we have
\begin{align*}
    \norm{ \sum_{r=1}^{m'} \initam{i}{r} \initwm{i}{r} }_{\infty} \leq 16 \sqrt{\log \rb{d + 1}} c_1 c_2 \epsilon_a \rb{ \log m},
\end{align*}
and using relation between $L_{2}$ and $L_{\infty}$ norm, we have
\begin{align}
\label{eq:L2-norm-pseudo-net1}
\norm{ \sum_{r=1}^{m'} \initam{i}{r} \initwm{i}{r} }_{2} \leq \sqrt{d+1} \norm{ \sum_{r=1}^{m'} \initam{i}{r} \initwm{i}{r} }_{\infty} \leq 16 \sqrt{ \rb{d + 1} \log \rb{d + 1}} c_1 c_2 \epsilon_a \rb{ \log m}.
\end{align}
Similarly, using Hoeffding's inequality (Fact \ref{fact:hoeffding-inequality}), with at least $1 - \frac{1}{c_1} - \frac{1}{c_3}$ probability, we get
\begin{align*}
\Pr \rb{ \abs{ \sum_{r=1}^{m'} \initam{i}{r} \initbm{i}{r} } \geq t } &\leq \exp \rb{ -\frac{ 2 t^2 m }{ m' \rb{ 2 c_1 \epsilon_a \sqrt{2 \log m} }^2 \rb{ 2 c_3 \sqrt{2 \log m} }^2 } } \\
    &\leq \exp \rb{ -\frac{ t^2 }{  32 c_1^2 c_3^2 \epsilon_a^2 \rb{ \log m}^2  } }. 
\end{align*}
Plugging $t=16 c_1 c_3 \epsilon_a \rb{ \log m}$, with at least $1 - \exp \rb{ - 8 } - \frac{1}{c_1}  - \frac{1}{c_3}$ probability, we get
\begin{align}
\label{eq:pseudo-net-bound-sum-part2}
    \abs{ \sum_{r=1}^{m'} \initam{i}{r} \initbm{i}{r} } \leq 16 c_1 c_3 \epsilon_a \rb{ \log m}.
\end{align}
Using Eq.\eqref{eq:L2-norm-pseudo-net1} and Eq.\eqref{eq:pseudo-net-bound-sum-part2}, with probability at least $0.99 - \frac{1}{c_1} - \frac{1}{c_2} - \frac{1}{c_3}$, we have 
\begin{align}
\label{eq:upper-bound-on-sum-ar0-wr0}
    \abs{ \sum_{r=1}^m \initam{i}{r} \rb{ \neuralvalzerom{i} } \ind{ \neuralvalzerom{i} \geq 0 } } &= \abs{ \dotp{ x } { \sum_{r=1}^{m'} \initam{i}{r} \initwm{i}{r} } + \sum_{r=1}^{m'} \initam{i}{r} \initbm{i}{r} }  \\
    \nonumber &\leq 16 \sqrt{ \rb{d + 1} \log \rb{d + 1}} c_1 c_2 \epsilon_a \rb{ \log m} + 16 c_1 c_3 \epsilon_a \rb{ \log m}.
\end{align}
This completes the proof.
\end{proof}

\begin{lemma} \label{lemma:generalization-neural-net-func} For any constant $T$, for any dimension $i \in [d]$, any time $1 \leq t \leq T$, any $\epsilon \in \rb{0, 1}$, suppose that the number of samples $n$ satisfies
\begin{align}\label{eqn:n_lb}
    n \geq O \rb{ \frac{ \rb{ M_{\al} - m_{\al} }^2 \rb{ Q + 1 }^2 d^2 \log \rb{d} \epsilon_a^4 U_{\theta^*}^4 m^4 \rb{ \log m }^2 }{ \epsilon^4 } }.
\end{align}
Then, with at least $0.98$ probability over random initialization, the population loss of any functions of the set $\{ x \mapsto \nnit{i}{(t)} \;\; | \;\; \norm{ \wt{i, r}{t} }_2 \leq \eta \LB T, \; \abs{ \bt{i, r}{t} } \leq \eta \LB T \;\; \forall r \in [m] \}$ is close to the empirical loss, i.e.
\begin{align*}
     \abs{ \E_{x \in \mcd} \left[ \al \rb{ \dfv{ \ft{t} }, x } \right] -  \al \rb{ \dfv{ \ft{t} }, \samples } } \leq \epsilon.
\end{align*}
\end{lemma}

\begin{proof}
We know that the loss for $i^{\text{th}}$ dimension $\al_i \rb{ \dfv{  \ft{t} }, x }$ depends on neural network $\nnit{i}{(t)}$ through $\rb{ N\rb{ \qpi{1}{i}; \tht{t}_i }, N\rb{ \qpi{2}{i}; \tht{t}_i }, \ldots, N\rb{ \qpi{Q}{i}; \tht{t}_i }, N \rb{ \vecx{i} ; \tht{t}_i } }$ vector. Using Fact \ref{fact:population-empirical-loss-diff-rademacher}, with at least $1 - \delta$ probability, we get
\begin{align}
\label{eq:population-empirical-loss-diff-init}
    \sup_{ N \in \mcf }  \abs{ \E_{x \sim \mcd} \left[ \al_i \rb{ \dfv{ \ft{t} }, x } \right] - \frac{1}{n} \sum_{i=1}^n \al_i \rb{ \dfv{ \ft{t} }, x } } \leq 2 \sqrt{2} L_s \rb{ Q + 1 } \empR \rb{ \mcx; \mathcal{F} }  + b_i \sqrt{ \frac{\log \frac{1}{\delta}}{ 2n } }
\end{align}
where $\mcf = \{ x \mapsto \nnit{i}{(t)} \;\; | \; \; \norm{ \wt{i, r}{t} }_2 \leq \eta \LB T, \; \abs{ \bt{i, r}{t} } \leq \eta \LB T \;\; \forall r \in [m] \}$. In the above equation, constant $b_i$ denotes upper bound on the loss $\al_i$ and $L_{s, i}$ denote standard Lipschitz constant of $\al_i$ with respect to $\rb{ N\rb{ \qpi{1}{i}; \tht{t}_i }, N\rb{ \qpi{2}{i}; \tht{t}_i }, \ldots, N\rb{ \qpi{Q}{i}; \tht{t}_i }, N \rb{ \vecx{i} ; \tht{t}_i } }$.  We denote $L_{c, i, j}$ as $j^{\text{th}}$ coordinate-wise Lipschitz continuity of loss $\al_i$ function as following: 
\begin{align*}
    L_{c, i, j} &\leq \sup_{ N \in \mcf, \norm{x}_2 \leq 1} \abs{  \Delta_x \phi'\rb{ N \rb{ \qpi{j}{i}, \tht{t}_i } } } \\
    &\leq \sup_{N \in \mcf, \norm{x}_2 \leq 1}   \frac{2}{ Q } \abs{ \phi'\rb{ N \rb{ \qpi{j}{i}, \tht{t}_i } } }  \\
    &\leq \frac{ 2 }{Q} \quad \forall j \in [ Q ], \\
    L_{c, i, Q + 1 } &\leq \sup_{N \in \mcf, \norm{x}_2 \leq 1} \frac{ \phi' \rb{ \nnit{i}{(t)} } }{ \phi \rb{ \nnit{i}{(t)} } } \\ 
    &= \sup_{N \in \mcf, \norm{x}_2 \leq 1} \frac{ \exp \rb{ \nnit{i}{(t)} } \ind{ \nnit{i}{(t)} \leq 0 } + \ind{ \nnit{i}{(t)} \geq 0 } }{ \exp \rb{ \nnit{i}{(t)} } \ind{ \nnit{i}{(t)} \leq 0 } + \rb{ \nnit{i}{(t)} + 1 } \ind{ \nnit{i}{(t)} \geq 0 } } \\
    &= \sup_{N \in \mcf, \norm{x} \leq 1} \ind{ \nnit{i}{(t)} \leq 0 } + \frac{1}{ \nnit{i}{(t)} + 1 } \ind{ \nnit{i}{(t)} \geq 0 } \\
    &\leq 1
\end{align*}
Using Lemma \ref{lemma:relation-std-lipschitz-coordinate-lipschitz}, standard Lipschitz constant of $\al_i$ is given by
\begin{align}
\label{eq:standard-lipschitz-loss}
    L_{s, i} &\leq \sqrt{ \sum_{j=1}^{ Q + 1 } L_{c, i, j}^2 } \leq \sqrt{ \frac{ 4 }{ Q } + 1 } \leq 2
\end{align}
To get constant $b_i$ (i.e., upper bound on $\al_i$), we use Lipschitz property of $\al_i$. We construct $\tilde{f}_i$ such that $\rb{ \nabla_i \tilde{f}_i \rb{ \qpi{1}{i} }, \nabla_i \tilde{f}_i \rb{ \qpi{2}{i} } , \ldots, \nabla_i \tilde{f}_i \rb{ \qpi{Q}{i} }, \nabla_i \tilde{f}_i \rb{ \vecx{i} } } = \rb{1, 1, \ldots, 1, 1}$.
\begin{align}
\label{eq:upper-bound-on-loss}
    \nonumber    \abs{ \al_i  \rb{ \dfv{ \ft{t} }, x } - \al_i  \rb{ \dfv{ \tilde{f}^{(t)} }, x } } = & \abs{ \sum_{j=1}^Q \Delta_x \dfqpit{j}{i}{(t)} - \sum_{j=1}^Q \Delta_x \nabla_i \tilde{f}_i \rb{ \qpi{j}{i} } } \\  
    \nonumber    &+ \abs{ \log \rb{ \phi \rb{ \nnit{i}{(t)} } } - \log \rb{  \phi \rb{ N \rb{ \vecx{i}, \tilde{\theta}_i } }  } } \\
       \nonumber \leq & \sum_{j=1}^Q \Delta_x \abs{ \phi \rb{ N \rb{ \qpi{j}{i}, \tht{t}_i } } - \phi \rb{ N \rb{ \qpi{j}{i}, \tilde{\theta}_i } } } \\
         \nonumber    &+ \abs{ \log \rb{ \phi \rb{ \nnit{i}{(t)} } } } \\
    \leq &  \sum_{j=1}^{ Q } \Delta_x \abs{ N\rb{ \qpi{j}{i}, \tht{t}_i }  } + \abs{ \nnit{i}{(t)} } 
\end{align}
Note that $\al  \rb{ \dfv{ \ft{t} }, x }$ depends upon $\rb{ N\rb{ \qpi{1}{i}; \tht{t}_i }, N\rb{ \qpi{2}{i}; \tht{t}_i }, \ldots, N\rb{ \qpi{Q}{i}; \tht{t}_i }, N \rb{ \vecx{i} ; \tht{t}_i } }$ vector and similarly, $\al \rb{ \dfv{ \tilde{f}^{(t)} }, x }$ depends upon $\rb{0, 0, 0, \ldots, 0, 0}$. Finding upper bound $\nnit{i}{(t)}$ for all $x \in \R^d$ with $\norm{x}_2 \leq 1$, we get
\begin{align*}
    \sup_{ N \in \mcf, \norm{x} \leq 1 } \nnit{i}{(t)} &\leq \sup_{ \norm{ \wt{i, r}{t} }_2 \leq \eta \LB T , \; \abs{ \bt{i, r}{t} } \leq \eta \LB T, \norm{x}_2 \leq 1 } \pnit{i}{(t)} + \dnp{T} \\
    &\leq \sup_{ \norm{ \wt{i, r}{t} }_2 \leq \eta \LB T , \; \abs{ \bt{i, r}{t} } \leq \eta \LB T, \norm{x}_2 \leq 1 } \sum_{r=1}^m \initam{i}{r} \sigma \rb{ \neuralvalzerom{i} } \\ 
    &+ \sum_{r=1}^m \initam{i}{r} \rb{ \dotp{ \wt{i, r}{t}}{  \nrmxi{i} } + \bt{i, r}{t} } \sigma \rb{ \neuralvalzerom{i} } + \dnp{T} \\
    & \stackrel{ (\RomanNumLow{1}) }{ \leq } 16 \sqrt{ \rb{d + 1} \log \rb{d + 1}} c_1 c_2 \epsilon_a \rb{ \log m} + 16 c_1 c_3 \epsilon_a \rb{ \log m} \\
    & + m \rb{ 2 c_1 \epsilon_a \sqrt{2 \log m} } \rb{ 2 \eta \LB T } + \dnp{T} \\
    & \stackrel{ ( \RomanNumLow{2} ) }{ \leq } 16 \sqrt{ \rb{d + 1} \log \rb{d + 1}} c_1 c_2 \epsilon_a \rb{ \log m} + 16 c_1 c_3 \epsilon_a \rb{ \log m} \\ 
    &+ m \rb{ 2 c_1 \epsilon_a \sqrt{2 \log m} } \rb{ 12 c_1 \epsilon_a \sqrt{2 \log m} } \rb{ \frac{m U_{\theta^*}^2 }{ 2 \epsilon } } + \dnp{T} \\
    & =  16 \sqrt{ \rb{d + 1} \log \rb{d + 1}} c_1 c_2 \epsilon_a \rb{ \log m} + 16 c_1 c_3 \epsilon_a \rb{ \log m} \\ 
    &+ m \rb{24 c_1^2 \epsilon_a^2 \log m  \rb{ \frac{ m U_{\theta^*}^2 }{\epsilon } } } + \dnp{T} \\
    & \leq O \rb{ \frac{ m^2 \epsilon_a^2 U_{ \theta^* }^2 \log m  }{ \epsilon } }
\end{align*}
where inequality $(\RomanNumLow{1})$ follows from Lemma \ref{lemma:small-value-at-init}, Lemma \ref{lemma:concentration-folded-normal} and Eq.(\ref{eq:upper-bound-on-wt-bt}). The inequality $(\RomanNumLow{2})$ uses our choices of $\eta$ and $T$ from Eq.(\ref{eq:values-eta-T-m-td}). We define $K$ as upper bound on $\sup_{ N \in \mcf, \; \norm{x}_2 \leq 1 } \nnit{i}{(t)}$:
\begin{align}
\label{eq:define-K}
    K := O \rb{ \frac{ m^2 \epsilon_a^2 U_{ \theta^* }^2 \log m  }{ \epsilon } }.
\end{align}
Using value of $K$ and Eq.(\ref{eq:upper-bound-on-loss}), we get upper bound $b_i$ on $\al_i$:
\begin{align*}
    b_i = 2K + K + \al_i  \rb{ \dfv{ \tilde{f}^{(t)} }, x } = 3K + 2.
\end{align*}
Using value of $b_i$ in Eq.(\ref{eq:population-empirical-loss-diff-init}) and Lemma \ref{lemma:empirical-rademacher-complexity-two-layer}, with at least $0.99 - \delta - \frac{1}{c_1} - \frac{1}{c_2}- \frac{1}{c_3}$ probability, we get
\begin{align*}
    \sup_{ N \in \mcf } & \abs{ \E_{x \in \mcd} \left[ \al_i \rb{ \dfv{ \ft{t} }, x } \right] - \frac{1}{n} \sum_{i=1}^n \al_i \rb{ \dfv{ \ft{t} }, x_i } } \\ 
    &\leq 4 \sqrt{2} \rb{ Q + 1 }  \frac{ 8 c_1 \epsilon_a \eta \LB T m \sqrt{2 \log m} }{ \sqrt{n} } + \rb{ 3K + 2 }\sqrt{ \frac{\log \frac{1}{\delta}}{ 2n } }.
\end{align*}
By summing over all dimension $i \in [d]$, with atleast $0.99 - d \delta - \frac{d}{c_1} - \frac{d }{c_2} - \frac{d }{c_3}$ probability, we get
\begin{align*}
\sup_{ N \in \mcf } \abs{ \E_{x \in \mcd} \left[ \al \rb{ \dfv{ \ft{t} }, x } \right] - \frac{1}{n} \sum_{i=1}^n \al \rb{ \dfv{ \ft{t} }, x_i } } &\leq \sum_{i=1}^d \sup_{ N \in \mcf } \abs{ \E_{x \in \mcd} \left[ \al_i \rb{ \dfv{ \ft{t} }, x } \right] - \frac{1}{n} \sum_{i=1}^n \al_i \rb{ \dfv{ \ft{t} }, x_i } }, \\
&\leq 4 \sqrt{2} d \rb{ Q + 1 }  \frac{ 8 c_1 \epsilon_a \eta \LB T m \sqrt{2 \log m} }{ \sqrt{n} } + \rb{ 3K + 2 } d \sqrt{ \frac{\log \frac{1}{\delta}}{ 2n } }.
\end{align*}
Using $\delta= \frac{0.001}{d}$ and our choice of $n$ given in \eqref{eqn:n_lb}, with probability at least $0.989$, we have
\begin{align*}
    \sup_{ N \in \mcf } & \abs{ \E_{x \in \mcd} \left[ \al_i \rb{ \dfv{ \ft{t} }, x } \right] - \frac{1}{n} \sum_{i=1}^n \al_i \rb{ \dfv{ \ft{t} }, x_i } } \leq \epsilon.
\end{align*}
\end{proof}

\begin{lemma} \label{lemma:generalization-target-func} (Concentration on approximated loss of target function) 	
	Suppose $n$ is sufficiently high such that it satisfies 
\begin{align*}
    n \geq O \rb{ \frac{ \rb{ M_{\al} - m_{\al} }^2 \rb{ Q + 1 }^2 d^2 \log \rb{d} \epsilon_a^4 U_{\theta^*}^4 m^4 \rb{ \log m }^2 }{ \epsilon^4 } }.
\end{align*}
If $n$ satisfies above condition, then with at least $0.9999$ probability, population loss of target function $\td$ is close to empirical loss i.e.
\begin{align*}
    \abs{ \E_{ x \sim \mcd } \left[ \al \rb{ \dfv{ \tarfunc{*} }, x } \right] - \al \rb{ \dfv{ \tarfunc{*} }, \mcx } } \leq \epsilon.
\end{align*}
\end{lemma}
\begin{proof}
Using Hoeffding's inequality (Fact \ref{fact:hoeffding-inequality}), we have
 \begin{align*}
     \text{Pr} \rb{ \abs{ \E_{ x \sim \mcd } \left[ \al \rb{ \dfv{ \tarfunc{*} }, \mcx } \right] - \al \rb{ \dfv{ \tarfunc{*} }, \mcx } } \geq \epsilon  } &\leq \exp \rb{ -\frac{ 2 n \epsilon^2 }{ \rb{ M_{\al} - m_{\al} }^2 } }
 \end{align*}
 Taking $n$ as
 \begin{align*}
     n \geq O \rb{ \frac{ \rb{ M_{\al} - m_{\al} }^2 \rb{ Q + 1 }^2 d^2 \log \rb{d} \epsilon_a^4 U_{\theta^*}^4 m^4 \rb{ \log m }^2 }{ \epsilon^4 } },
 \end{align*}
 with at least probability $0.9999$, we get
 \begin{align}
 \label{eq:pop-emp-loss-diff-bound-for-target}
      \abs{ \E_{ x \sim \mcd } \left[ \al \rb{ \dfv{ \tarfunc{*} }, x } \right] - \al \rb{ \dfv{ \tarfunc{*} }, \mcx } } \leq \epsilon
 \end{align}
\end{proof} 

\begin{corollary}
    Under same setting as Theorem \ref{theorem:optimization-empirical-approximated-L} and 
    \begin{align*}
        n \geq O \rb{ \frac{ \rb{ M_{\al} - m_{\al} }^2 \rb{ Q + 1 }^2 d^2 \log \rb{d} \epsilon_a^4 U_{\theta^*}^4 m^4 \rb{ \log m }^2 }{ \epsilon^4 } }
    \end{align*}
    then with at least $0.94$ probability, we get
    \begin{align*}
        \E_{\mathrm{sgd}} \left[ \frac{1}{T} \sum_{t=0}^{T-1}   \E_{ x \sim \mcd } \left[ \al (\fpt{t}, x) \right]  \right] - \E_{ x \sim \mcd } \left[ \al ( \dfv{ \tarfunc{*} }, x) \right] \leq O(\epsilon).   
    \end{align*}
\end{corollary}
\begin{proof}
 The corollary follows from Theorem \ref{theorem:optimization-empirical-approximated-L}, Lemma \ref{lemma:generalization-neural-net-func} and Lemma \ref{lemma:generalization-target-func}.
\end{proof}

Before stating our main theorem, we recall and define necessary terms used in stating the theorem. Recall that 
\begin{align*}
	M_{\nabla F} &= \max_{i \in [d], x \in \R^d} \nabla_i F_i \rb{ \vecx{i} } = \max_{i \in [d], x \in \R^d} \frac{ \partial F_i \rb{ \vecx{i} } }{ \partial x_i } , \\
	m_{\nabla F} &= \min_{i \in [d], x \in \R^d} \nabla_i F_i \rb{ \vecx{i} } = \min_{i \in [d], x \in \R^d} \frac{ \partial F_i \rb{ \vecx{i} } }{ \partial x_i }, \\
	M_{\nabla^2 F} &= \max_{i \in [d], x \in \R^d} \nabla_i^2 F_i \rb{ \vecx{i} } = \max_{i \in [d], x \in \R^d} \frac{ \partial^2 F_i \rb{ \vecx{i} } }{ \partial x_i^2 } , \\
	m_{\nabla^2 F} &= \min_{i \in [d], x \in \R^d} \nabla_i^2 F_i \rb{ \vecx{i} } = \min_{i \in [d], x \in \R^d} \frac{ \partial^2 F_i \rb{ \vecx{i} } }{ \partial x_i^2 }, \\
    M_{\al} &= \sup_x \al \rb{ \dfv{ \tarfunc{*} }, x } = 2 M_{\nabla F} - \log \rb{ m_{\nabla F} },  \\
    m_{\al} &= \inf_x \al \rb{ \dfv{ \tarfunc{*} }, x } = 2 m_{\nabla F} - \log \rb{ M_{\nabla F} }. \\
\end{align*}
Recall that for any function $\psi: \R \to \R$ with Taylor expansion $\psi(y) = \sum_{j=0}^{\infty} c_j y^j$, then its complexity $C_0( \psi, \epsilon )$ for any $\epsilon > 0$ is given by 
\begin{align*}
C_0( \psi, \epsilon ) = O( (\sum_{i=0}^{\infty} (i + 1)^{1.75} |c_i|) \mathrm{poly} ( \inlinefrac{1}{\epsilon} ) ), 
\end{align*}
which is a weighted norm of the Taylor coefficients.  Recall that we define upper bound on complexity of learning any $\psi$ function as
\begin{align*}
U_{ \psi } &= \max_{i \in [d], j \in [p_i] } C_0 \rb{ \psi_{i, j}, \epsilon }.
\end{align*}
Now, we will state our main theorem.
\begin{theorem} \label{theorem:final-theorem} (loss function is close to optimal) For every $\epsilon \in \rb{ 0, 1 }$, for every $m > \mathrm{poly}\rb{U_{\psi}, d, \rb{ \max_{i \in [d]} p_i }, \frac{1}{\epsilon}}, \eta=\Tilde{O} \rb{ \frac{1}{m \epsilon } }$ and $T=O \rb{ \frac{ d^2 \rb{ \max_{i \in [d]} p_i }^2 U_{ \psi }^2 \log m }{ \epsilon^2 } }$, for any target function $\td$ with finite second order derivative and number of quadrature points $Q \geq \frac{2 d M_{\nabla^2 \tarfunc{*} } + 2 d K_2 }{\epsilon}$ and number of training points $n \geq O \rb{ \frac{ \rb{ M_{\al} - m_{\al} }^2 \rb{ Q + 1 }^2 d^6 \log \rb{d} \rb{ \max_{i \in [d]} p_i }^4 U_{\psi}^4 m^4 \rb{ \log m }^2 }{ \epsilon^4 } }$, with at least $0.94$ probability, we have
\begin{align*}
    \E_{\mathrm{sgd}} \left[ \frac{1}{T} \sum_{t=0}^{T-1}   \E_{ x \sim \mcd } \left[ L (\ft{t}, x) \right] \right] - \E_{ x \sim \mcd } \left[ L (\tarfunc{*}, x) \right] \leq O(\epsilon),
\end{align*}
where $K_2$ is given by
\begin{align*}
    K_2 &= O \rb{ \frac{ m^2 U_{\psi}^2 d^6 \rb{ \max_{i \in [d]} p_i } }{ \epsilon } }.
\end{align*}
\end{theorem}
\begin{proof}
First, we will try to bound for all $x \in \R^d$ with $\norm{x}_2 \leq \frac{1}{2}$: \todo{Quantify $x$ throughout. Is it all $x \in [-1, 1]$?}
\begin{align*}
    \abs{ \al (\dfv{ \tarfunc{*} }, x) - L (\tarfunc{*},  x) } \leq & \abs{ \sum_{i=1}^d \sum_{j=1}^{ Q } \Delta_x \nabla_i \tarfunci{i} \rb{ \qpi{j}{i} } - \tarfunci{i}  } \\
    \leq &  \sum_{i=1}^d \abs{ \sum_{j=1}^{ Q } \Delta_x \nabla_i \tarfunci{i} \rb{ \qpi{j}{i} } - \tarfunci{i}  } \\
    \leq & \frac{ 2 d M_{\nabla^2 F} }{ Q }.
\end{align*}
Similarly, bounding error for $\fpt{t}$ for all $x \in \R^d$ with $\norm{x}_2 \leq \frac{1}{2}$, we will get
\begin{align*}
    \abs{ \al (\dfv{ \ft{t} }, x) - L (\ft{t}, x) } &\leq \abs{ \sum_{i=1}^d \rb{ \sum_{j=1}^{ Q } \Delta_x  \dfqpit{j}{i}{(t)}   - \ft{t}_i ( \vecx{i} )   } } \\
    &\leq   \frac{2 d \rb{ \sup_{x, i \in [d], t \in [T]} \nabla_i^2 \ft{t}_i ( \vecx{i} ) } }{ Q }. 
\end{align*}
To get $ \sup_{x, i \in [d], t \in [T]} \nabla_i^2 \ft{t}_i ( \vecx{i} )$, we will use Eq.(\ref{eq:define-K}).
\begin{align*}
    \sup_{x, i \in [d], t \in [T]} & \nabla_i^2 \ft{t}_i ( \vecx{i} ) = \sup_{x, i \in [d], t \in [T]} \abs{ \frac{ \partial^2 \ft{t}_i \rb{ \vecx{i} }  }{ \partial x_i^2 } }  \\
    &= \sup_{x, i \in [d], t \in [T]} \abs{ \frac{ \partial   }{ \partial x_i } \rb{ \phi \rb{ \nnit{i}{(t)} } } } \\
    &\leq \sup_{x, i \in [d], t \in [T]} \abs{ \frac{\partial }{ \partial x_i } \nnit{i}{(t)} } \\
    &\leq \sup_{x, i \in [d], t \in [T]} \Bigg|  \sum_{r=1}^m \initam{i}{r} \sigma' \rb{ \neurvalm{i}{(t)}{ \nrmxi{i} } } \Bigg( \rb{\initwm{i}{r, i} + \wt{i, r, i}{t} } \\ 
    &+ \rb{ \initwm{i}{r, {i+1}} + \wt{i, r, i+1}{t} } \frac{ x_i }{ \sqrt{ 1 - \| \vecx{i} \|^2 } } \Bigg) \Bigg| \\
    &= \sup_{x, i \in [d], t \in [T]}  \sum_{r \in \mathcal{H}_i } \initam{i}{r} \ind{ \neuralvalzerom{i} \geq 0 } \Bigg( \rb{\initwm{i}{r, i} + \wt{i, r, i}{t} } \\ 
    &+ \rb{ \initwm{i}{r, {i+1}} + \wt{i, r, i+1}{t} } \frac{ x_i }{ \sqrt{ 1 - \| \vecx{i} \|^2 } } \Bigg) + \sum_{r \in \mHB{t}_i } \initam{i}{r} \ind{ \neurvalm{i}{(t)}{ \nrmxi{i} } \geq 0 } \\ 
    &\Bigg( \rb{\initwm{i}{r, i} + \wt{i, r, i}{t} } + \rb{ \initwm{i}{r, {i+1}} + \wt{i, r, i+1}{t} } \frac{ x_i }{ \sqrt{ 1 - \| \vecx{i} \|^2 } } \Bigg) \\
    &\stackrel{( \RomanNumLow{1} )}{ \leq } 32 c_1 c_2 \epsilon_a \rb{ \log m} + 2 m \rb{ 2c_1 \epsilon_a \sqrt{2 \log m} } \rb{ \eta \LB T } \\ 
    &+ \rb{ c_4 m \frac{ 4 \eta \LB T \sqrt{m} }{\sqrt{ \pi }} } \rb{ 2 c_1 \epsilon_a \sqrt{2 \log m} } \rb{ \frac{2 c_2 \sqrt{ 2 \log m }}{ \sqrt{m} } } \\ 
    &+ \rb{ c_4 m \frac{ 4 \sqrt{m} }{\sqrt{ \pi }} } \rb{ 2 c_1 \epsilon_a \sqrt{2 \log m} } \rb{ \eta \LB T }^2 \\
    &\leq 32 c_1 c_2 \epsilon_a \rb{ \log m} + 2 m \rb{ 2c_1 \epsilon_a \sqrt{2 \log m} } \rb{ \frac{ \rb{ 3 c_1 m U_{\theta^*}^2 \epsilon_a \sqrt{2 \log m}  } }{ \epsilon} } \\ 
    &+ \rb{ c_4 m \frac{ 4 \sqrt{m} }{\sqrt{ \pi }} } \rb{ 2 c_1 \epsilon_a \sqrt{2 \log m} } \rb{ \frac{2 c_2 \sqrt{ 2 \log m }}{ \sqrt{m} } } \rb{ \frac{ \rb{ 3 c_1 m U_{\theta^*}^2 \epsilon_a \sqrt{2 \log m}  } }{ \epsilon} } \\ 
    &+ \rb{ c_4 m \frac{ 4 \sqrt{m} }{\sqrt{ \pi }} } \rb{ 2 c_1 \epsilon_a \sqrt{2 \log m} } \rb{ \frac{ \rb{ 3 c_1 m U_{\theta^*}^2 \epsilon_a \sqrt{2 \log m}  } }{ \epsilon} }^2  \\
    &\stackrel{ (\RomanNumLow{2}) }{ \leq }  O \rb{ d^2 \epsilon } + O \rb{ d^2 m^2 U_{\theta^*}^2 \epsilon } + O \rb{ m^2 U_{\theta^*}^2 d^4 \epsilon } + O \rb{ m^{3.5} U_{\theta^*}^4 \epsilon^2 }\\
    &\leq O \rb{ m^2 U_{\theta^*}^2 d^4 \epsilon },
\end{align*}
where inequality (i) follows by plugging $t=16c_1c_2\epsilon_a \log m$ in Eq.\eqref{eq:bound-sum-ar0-wr0j}, with . Define $K_2$ as upper bound on $\nabla_i^2 \ft{t}_i ( \vecx{i} )$,
\begin{align*}
    K_2 = O \rb{ m^2 U_{\theta^*}^2 d^4 \epsilon } =  O \rb{ \frac{ m^2 U_{\psi}^2 d^6 \rb{ \max_{i \in [d]} p_i } }{ \epsilon } }.
\end{align*}
Taking $Q$ as
\begin{align}
    Q \geq \frac{2 d M_{\nabla^2 \tarfunc{*} } + 2 d K_2 }{\epsilon}
\end{align}
Using given value of $Q$, we get that \todo{what is the quantification on $x$ here?}
\begin{align}
    \abs{ \al ( \dfv{ \tarfunc{*} } , x) - L (\tarfunc{*}, x) } &\leq \epsilon, \\
    \abs{ \al ( \dfv{ \ft{t} } , x) - L (\ft{t},  x) } &\leq \epsilon.
\end{align}
Using these relations, we get 
\begin{align*}
    \E_{\mathrm{sgd}} \left[ \frac{1}{T} \sum_{t=0}^{T-1}   \E_{ x \sim \mcd } \left[ L (\ft{t}, x) \right] \right] - \E_{ x \sim \mcd } \left[ L ( \tarfunc{*}, x) \right] \leq O(\epsilon).
\end{align*}
By the definition of KL divergence, we get
\begin{align*}
    \E_{\mathrm{sgd}} \left[ \frac{1}{T} \sum_{t=0}^{T-1}   \text{KL} \rb{ p_{\tarfunc{*}, Z} || p_{\ft{t}, Z} } \right] \leq O(\epsilon).
\end{align*}
\end{proof}

\section{Problem in Training of Constrained Normalizing Flow}
\label{sec:problem-training-cnf}
In this section, we provide details of why different initializations cause problems (described in section \ref{section:constrained-normalizing-flow-main}) in the training of Constrained Normalizing Flows. Recall that the loss function of normalizing flow with Gaussian distribution as base distribution is given by 
\begin{align*}
L_{G} \rb{ f, x } = \frac{f(x)^T f(x)}{2} - \log \rb{ \abs{ \det \rb{ \frac{ \partial f(x) }{ \partial x } } } },
\end{align*}
where function $f(x) : \R^d \to \R^d$ is parameterized using $d$ neural networks $N_1, N_2, \ldots, N_d$. The $i^{\text{th}}$ dimension of the function $\funci{i} = \nni{i}$. The neural network in CNF is defined as 
\begin{align*}
&\nni{i} = \tau \sum_{r=1}^m \initam{i}{r} \, \tanh \left( \dotp{  \initwm{i}{r} + w_{i, r} }{ \vecx{i} } + \left( \initbm{i}{r} + b_{i, r}  \right) \right), \\ 
&\text{with constraints $\initwm{i}{r, i} + w_{i, r, i} \geq \epsilon$, for all $r \in [m]$ and $i \in [d]$}.
\end{align*} 
Here, $\epsilon>0$ is a small constant and $\tau$ is a normalization constant which only depends on $m$. We use $\theta_i$ to denote parameters of $\nni{i}$ and $\theta$ to denote parameters of all neural networks. Initial weights $\initam{i}{r}$ and $\initwm{i}{r, i}$ are sampled from \emph{half-normal} distribution with parameters $\rb{0, \epsilon_a^2}$ and $\rb{0, \sigma_{wb}^2}$, resp. The half-normal random variable $Y$ with parameters $\rb{\mu, \sigma^2}$ is given by simply $\abs{X}$ where $X \sim \N \rb{\mu, \sigma^2}$. Here $\N \rb{\mu, \sigma^2}$ denote the Gaussian distribution with mean $\mu$ and variance $\sigma^2$.  Other weights ($\initbm{i}{r}, \initwm{i}{r, j}$ for $j \neq i$) are sampled from $\N \rb{0, \sigma_{wb}^2 }$.  We optimize the objective using \emph{projected SGD}.  Note that in this case, the constraints are very simple and projected SGD incurs very little overhead. 

The pseudo network function is given by $g(x) = \rb{ g_1 ( \vecx{1} ), g_2 ( \vecx{2} ), \ldots, g_d ( \vecx{d} ) }$, where $g_i ( \vecx{i} ) = \pni{i}$ is given by
\begin{align*}
\pni{i} = \tau \sum_{r=1}^m \initam{i}{r} \rb{ \tanh \rb{ \neuralvalzerom{i} } + \tanh' \rb{ \neuralvalzerom{i} } \rb{ \dotp{ w_{i, r} }{ \vecx{i} } + b_{i, r} } }
\end{align*}
 with constraints $\initwm{i}{r, i} + w_{i, r, i} \geq \epsilon$ for all $r$. We decompose pseudo network in two parts:
\[
\pni{i} = P_c (\vecx{i})  + \pnilt{\ell}{i}{},
\]
where $P_c (\vecx{i})$ and $\pnilt{\ell}{i}{}$ is given by
\begin{align*}
P_c (\vecx{i}) &= \tau \sum_{r=1}^m \initam{i}{r}  \tanh ( \dotp{ \initwm{i}{r}  }{ \vecx{i} }  + \initbm{i}{r} \\ 
\pnilt{\ell}{i}{}  &= \tau \sum_{r=1}^m \initam{i}{r} \tanh' ( \dotp{ \initwm{i}{r}  }{ \vecx{i} } + \initbm{i}{r} ) \rb{ \dotp{ w_{i, r} }{ \vecx{i} } + b_{i, r}  }.
\end{align*}
The loss function for pseudo network is given b
\begin{align*}
L_{G} \rb{ g, x } = \frac{g(x)^T g(x)}{2} - \log \rb{ \abs{ \det \rb{ \frac{ \partial g(x) }{ \partial x } } } } = \sum_{i=1}^d g_i \rb{ \vecx{i} } - \sum_{i=1}^d \log \rb{ \frac{ \partial g_i \rb{ \vecx{i} } }{ \partial x_i } }
\end{align*}
where $g(x) = \rb{g_i \rb{ \vecx{1} }, g_i \rb{ \vecx{1} }, \ldots, g_i \rb{ \vecx{n} } }$. The pseudo network $\pni{i}$, which approximates the neural network $\nni{i}$, will be
\[
\pni{i} = \tau \sum_{r=1}^m \initam{i}{r} \rb{ \tanh ( \dotp{ \initwm{i}{r} }{ \vecx{i} } + \initbm{i}{r} ) + \tanh' ( \dotp{ \initwm{i}{r} }{ \vecx{i} } + \initbm{i}{r} ) \rb{ \dotp{ w_r }{ \vecx{i} } + b_r  } },
\]
with constraints $\initwm{i}{r, i} + w_{i, r, i} \geq \epsilon$, for all $r \in [m]$. We decompose $\pni{i}$ into two parts: $\pni{i} = P_c \rb{ \vecx{i} } + \pnil{\ell}{i}$, where
\[
    P_c \rb{ \vecx{i} } = \tau \sum_{r=1}^m \initam{i}{r} \tanh ( \dotp{ \initwm{i}{r} }{ \vecx{i} } + \initbm{i}{r} )  \quad \text{and} \quad \pnil{\ell}{i} = \tau \sum_{r=1}^m \initam{i}{r} \tanh' ( \dotp{ \initwm{i}{r} }{ \vecx{i} } + \initbm{i}{r} ) \rb{ \dotp{ w_r }{ \vecx{i} } + b_r  }.
\]
Note that $P_c \rb{ \vecx{i} }$ only depends upon initialization and does not depend on parameters $\theta_i$. 

Let $\tarfunc{*}$ denote the target function and $C(\tarfunc{*})$ denote some complexity measure of $\tarfunc{*}$. We devide our analysis into two cases based on variance of $\initwm{i}{r}$ and $\initbm{i}{r}$. (1) In the first case, standard deviation $\sigma_{wb}$ satisfies $ \frac{\epsilon^2}{ C ( \tarfunc{*} ) \sqrt{ \log (md) } }   \leq \sigma_{wb} \leq 1$. (2) In the second case, standard deviation $\sigma_{wb}$ satisfies $ \frac{1}{\sqrt{m}} \leq \sigma_{wb} \leq \frac{\epsilon^2}{ C ( \tarfunc{*} ) \sqrt{ \log \rb{ m d } } }  $. We call the first case \emph{larger variance initalization} case and the second one \emph{smaller variance intialization} case. Analysis for larger variance case is given in Section \ref{subsec:large-variance-cnf-problem} and analysis for smaller variance case is given in Section \ref{subsec:smaller-variance-cnf-problem}. \todo{Need to revisit this para}


\subsection{Problem in optimization for smaller variance initialization case}
\label{subsec:smaller-variance-cnf-problem}

In this section, we will provide details about the problem in smaller variance initialization case for Constrained Normalizing Flows (CNFs). We prove in Theorem \ref{theorem:smaller-variance-problem-cnf-appendix} that if we choose small learning rate $\eta$ and number of time steps $T$ according to the theorem statement, then function learned by sufficiently overparameterized CNFs is close to a linear function. To prove the theorem, we start by bounding maximum possible change in weights $\norm{ \wt{i, r}{t }}$ and biases $\abs{ \bt{r}{t} }$ during $t=T$ iterations in Lemma \ref{lemma:bound-wt-bt-cnf}. Using bound on change in weights, we establish closeness between function value given by neural networks and function value given by pseudo networks (Lemma \ref{lemma:coupling-function-values-cnf}). We, then, prove that for any $t \in [T]$, pseudo network at time $t$ is close to a linear function (Lemma \ref{lemma:linear-pseudo-linear}). Using closeness between neural network and pseudo network and linearity of pseudo network, we get that neural networks are close to a linear function for given small learning rate $\eta$ and number of time steps $T$. Note that choosing similar values of $\eta$ and $T$ in supervised learning enables the provable \emph{successful} training of neural network. The same issue in approximation arises for \emph{all} activations with continuous derivative.

Recall that neural network $\nnit{i}{(t)}$ is given by
\begin{align*}
\nnit{i}{(t)} = \sum_{r=1}^m \initam{i}{r} \tanh \rb{ \neurvalm{i}{(t)}{\vecx{i}} },
\end{align*}
and derivative $\frac{\partial \nnit{i}{(t)} }{ \partial x_i }$ is given by
\begin{align*}
\frac{\partial \nnit{i}{(t)} }{ \partial x_i } = \sum_{r=1}^m \initam{i}{r} \tanh' \rb{ \neurvalm{i}{(t)}{\vecx{i}} } \rb{ \initwm{i}{r, i} + \wt{i, r, i}{t} }. 
\end{align*}
We denote $\frac{\partial \nnit{i}{(t)} }{ \partial x_i } $ as $N^{\prime }( \vecx{i} ; \tht{t}_i )$. 

\begin{lemma} (Bound on change in weights and biases) \label{lemma:bound-wt-bt-cnf} For every $x$ with $\norm{x} \leq 1$, every $i \in [d]$ and time  step $t \geq 1$, upper bound on weights $\wt{i, r}{t}$ and biases $\bt{i, r}{t}$ is given by following with at least $1 - \frac{d}{c_1} - \frac{d}{c_2}$ probability for any constant $c_1 > 10$, $c_2 > 10$.
\begin{align*}
\norm{ \wt{i, r}{t} }_2 &\leq \left( \frac{ \Tilde{L}_1 + \Tilde{L}_2 +  2 c_2 \Tilde{L}_2 \sigma_{wb}  \sqrt{2 \log \rb{m d} } }{ \Tilde{L}_2 } \right) \left( \left(1 + 2 \eta c_1 \epsilon_a \tau \sqrt{2 \log m} \Tilde{L}_2  \right)^t - 1 \right)  \\
\abs{ \bt{i, r}{t} }  &\leq 2 \eta c_1 \epsilon_a \tau \sqrt{ 2 \log m } \left( \Tilde{L}_1 +  2 c_2 \Tilde{L}_2 \sigma_{wb} \sqrt{ 2 \log \rb{m d} }   \right) t \\ 
&+ \left( \frac{  \Tilde{L}_1 + \Tilde{L}_2 +  2 c_2 \Tilde{L}_2 \sigma_{wb}  \sqrt{2 \log \rb{m d} }  }{ \Tilde{L}_2  } \right) \left(  \left(1 + 2 \eta c_1 \epsilon_a \tau \sqrt{2 \log m} \Tilde{L}_2 \right)^t - 1 \right)
\end{align*}
\end{lemma}
\begin{proof}
We first find upper bound on the derivative of loss function and $w_{i, r}$ and $b_{i, r}$.  We denote $\zeroone{i} = \rb{0, 0, \ldots, 0, 1} \in \R^i$. By taking derivative of $L_G( \ft{t}, x)$ with respect to $w_{i, r}$, we get
\begin{align*}
    \frac{ \partial L_G( \ft{t}, x) }{ \partial w_{i, r} } =& { \tau \nnit{i}{(t)} } ( \initam{i}{r} \vecx{i} \tanh' ( \neurvalm{i}{(t)}{\vecx{i}} )  )  \\ 
    &- \frac{\tau}{ N^{\prime }( \vecx{i} ; \tht{t}_i ) }  \Bigg( \zeroone{i} \initam{i}{r} \Big( \tanh'( \neurvalm{i}{(t)}{\vecx{i}} ) \\
    &+ \rb{ \initwm{i}{r, i} + \wt{i, r, i}{t} } \vecx{i} \tanh''( \neurvalm{i}{(t)}{\vecx{i}} ) \Big) \Bigg).
\end{align*}
We assume that $L_G( \ft{t} , x)$ is $\Tilde{L}_1$-lipschitz continuous wrt $N$ and $\Tilde{L}_2$-lipschitz continuous wrt $N^{\prime }$. Assuming $|\tanh'(.)| \leq 1$ and $\norm{x}_2 \leq 1$, we have
\begin{align*}
    \norm{ \frac{ \partial L_G( \ft{t}, x) }{ \partial w_{i, r} } }_2 \leq \tau \Tilde{L}_1  \initam{i}{r}  + \tau \Tilde{L}_2 \initam{i}{r} \left( 1 + \abs{ \initwm{i}{r, i} + \wt{i, r, i}{t} } |\tanh''( \neurvalm{i}{(t)}{ \vecx{i} } )| \right).
\end{align*}
Assuming $|\tanh''(.)| \leq 1$, we get
\begin{align*}
    \norm{ \frac{ \partial L_G( \ft{t} , x) }{ \partial w_{i, r} } }_{2} \leq \tau \Tilde{L}_1  \initam{i}{r}   + \tau \Tilde{L}_2  \initam{i}{r} \left( 1 +  \abs{ \wt{i, r, i}{t} } + \abs{ \initwm{i}{r, i} } \right).
\end{align*}
Using Lemma~\ref{lemma:concentration-folded-normal} for $\initam{i}{r}$ and $\initwm{i}{r, i}$, with probability at least $1 - \frac{1}{c_1} - \frac{1}{c_2}$, we have
\begin{align}
\label{eq:bound-loss-derivative-wrt-w}
    \norm{ \frac{ \partial L_G( \ft{t}, x) }{ \partial w_{i, r} } }_2 \leq  \left( 2 c_1 \epsilon_a \tau \sqrt{2 \log m} \right)  \left( \Tilde{L}_1 + \Tilde{L}_2 \left( 1 +  \abs{ \wt{i, r, i}{t} } +  2 c_2 \sigma_{wb} \sqrt{2 \log \rb{ m d} } \right) \right).
\end{align}
For projected gradient descent, we get
\begin{align*}
    \norm{ \wt{i, r}{t} }_2 &\leq \eta \sum_{j=0}^{t-1} \norm{ \frac{ \partial L_G( \ft{j}, \xt{j}) }{ \partial w_{i, r} } }_2 \\
    &\leq \eta \sum_{j=0}^{t-1} \left( \left( 2 c_1 \epsilon_a \tau \sqrt{2 \log m} \right)  \left( \Tilde{L}_1 + \Tilde{L}_2 +  2 c_2  \sigma_{wb} \Tilde{L}_2  \sqrt{2 \log \rb{ m d } } \right) + \left( 2 c_1 \epsilon_a \tau \sqrt{2 \log m} \right)  \Tilde{L}_2 | \wt{i, r, i}{j} | \right) \\
    & \leq \left( 2 \eta c_1 \epsilon_a \tau \sqrt{2 \log m} \right)   \left( \Tilde{L}_1 + \Tilde{L}_2 + 2 c_2  \Tilde{L}_2  \sigma_{wb} \sqrt{2 \log \rb{ m d } } \right) t + \left( 2 \eta c_1 \epsilon_a \tau \sqrt{2 \log m} \Tilde{L}_2  \right)   \left( \sum_{j=0}^{t-1} \norm{ \wt{i, r}{j} }_2 \right) .
\end{align*}
By defining $\alpha$ and $\beta$ as
\begin{align*}
    \alpha &= \left( 2 \eta c_1 \tau \epsilon_a \sqrt{2 \log m} \right)   \left( \Tilde{L}_1 + \Tilde{L}_2 +  2 c_2  \Tilde{L}_2 \sigma_{wb} \sqrt{2 \log \rb{ m d } } \right) \\
    \beta &= \left( 2 \eta c_1 \epsilon_a \tau \sqrt{2 \log m} \Tilde{L}_2  \right),
\end{align*}
we get
\begin{align}
\label{eq:w-t-leq-sum-w-j}
    \norm{ \wt{i, r}{t} }_2 &\leq \alpha t + \beta \left( \sum_{j=0}^{t-1} \norm{ \wt{i, r}{j} }_2 \right), \\
\nonumber    \text{where } \quad \sum_{j=0}^{t-1} \norm{ \wt{i, r}{j} }_2 & \leq \alpha (t-1) + (1 + \beta) \left( \sum_{j=0}^{t-2} \norm{ \wt{i, r}{j} }_2 \right) \\
\nonumber    &\leq \alpha \left( (t-1) + (1 + \beta) (t-2) \right) + (1 + \beta)^2 \left( \sum_{j=0}^{t-3} \norm{ \wt{i, r}{j} }_2 \right) \\
\nonumber    &\leq \alpha \left( (t-1) + (1 + \beta) (t-2) + (1 + \beta)^2 (t-3)  \right) + (1 + \beta)^3 \left( \sum_{j=0}^{t-4} \norm{ \wt{i, r}{j} }_2 \right). \\
\end{align}
In general, for any $t' \in \{0, 1, \ldots, t-1\}$, we can write 
\begin{align*}
    \sum_{j=0}^{t-1} \norm{ \wt{i, r}{j} }_2 &\leq \alpha \left( \sum_{j=1}^{t-t'-1} (1+\beta)^{j-1} (t-j) \right) + (1 + \beta)^{(t-t'-1)} \left( \sum_{j=0}^{t'} \norm{ \wt{i, r}{j} }_2 \right).
\end{align*}
By taking $t'=0$, we get
\begin{align*}
    \sum_{j=0}^{t-1} \norm{ \wt{i, r}{j} }_2 &\leq \alpha \left( \sum_{j=1}^{t-1} (1+\beta)^{j-1} (t-j) \right) .
\end{align*}

Note that $\sum_{j=1}^{t-1} (1 + \beta)^{(j-1)} (t - j)$ is sum of an arithmetic-geometric progression (AGP). Using Fact~\ref{fact:sum-of-agp}, we can simplify the above sum as
\begin{align}
\label{eq:w-sum-bound}
    \nonumber \sum_{j=0}^{t-1} \norm{ \wt{i, r}{j} }_2 &\leq \alpha \left( \sum_{j=1}^{t-1} (1+\beta)^{j-1} (t-j) \right) \\
    \nonumber &= \alpha \left( \frac{ (t-1) - (1 + \beta)^{t-1} }{ -\beta }  - \frac{ (1 + \beta) \left( 1 - (1 + \beta)^{t-2} \right) }{ \beta^2 } \right) \\
    \nonumber &= \alpha \left( \frac{ \beta (1 + \beta)^{t-1} - \beta (t - 1) - (1 + \beta) + (1 + \beta)^{t-1}  }{ \beta^2 } \right) \\
    &= \alpha \left( \frac{ (1 + \beta)^t - (1 + \beta t) }{ \beta^2 } \right) 
\end{align}
Using Eq.(\ref{eq:w-sum-bound}) to bound $\norm{ \wt{i, r}{t} }_2$ in Eq. \eqref{eq:w-t-leq-sum-w-j}, we get
\begin{align*}
    \norm{ \wt{i, r}{t} }_2 &\leq \alpha \left( t + \beta \left( \frac{ (1 + \beta)^t - (1 + \beta t) }{ \beta^2 } \right) \right) \\
    &= \alpha \left( \frac{ (1 + \beta)^t - 1 }{ \beta }  \right) \\
    &= \left( \frac{ \Tilde{L}_1 + \Tilde{L}_2 +  2 c_2 \Tilde{L}_2 \sigma_{wb}  \sqrt{2 \log \rb{ m d} } }{ \Tilde{L}_2  } \right) \left( \left(1 + 2 \eta c_1 \epsilon_a \tau \sqrt{2 \log m} \Tilde{L}_2   \right)^t - 1 \right).
\end{align*}
This completes the proof of upper bounding $\norm{ \wt{i, r}{t} }_2$. We use a similar procedure for $\abs{ \bt{i, r}{t} }$. By taking derivative $\frac{ \partial L_G ( \ft{t}, x ) }{\partial b_{i, r}}$, we get
\begin{align*}
    \frac{ \partial L_G( \ft{t} , x) }{ \partial b_{i, r} } =&  \nnit{i}{(t)} \tau \left( \initam{i}{r} \tanh'( \neurvalm{i}{(t)}{\vecx{i}} ) \right) \\ 
    &- \frac{ \tau }{ N'( \vecx{i} ; \tht{t}_i )}  \left( \initam{i}{r} ( \initwm{i}{r, i} + \wt{i, r, i}{t} ) \tanh''( \neurvalm{i}{(t)}{\vecx{i}} ) \right) .
\end{align*}
We assume that $L_G( \ft{t} , x)$ is $\Tilde{L}_1$-lipschitz wrt $N$ and $\Tilde{L}_2$-lipschitz wrt $N'$. Additionaly, using $| \tanh'( \cdot ) | \leq 1$ and $| \tanh''( \cdot ) | \leq 1$, we get
\begin{align*}
    \left| \frac{ \partial L_G( \ft{t} , x) }{ \partial b_{i, r} }  \right| \leq \Tilde{L}_1 \initam{i}{r} \tau + \Tilde{L}_2 \initam{i}{r} \tau  \left( \initwm{i}{r, i}  + | \wt{i, r, i}{t} | \right).
\end{align*}
Using Lemma~\ref{lemma:concentration-folded-normal} for $\initam{i}{r}$ and $\initwm{i}{r}$, with probability at least $1 - \frac{1}{c_1} - \frac{1}{c_2}$, we get
\begin{align}
\label{eq:bound-loss-derivative-wrt-b}
    \left| \frac{ \partial L_G( \ft{t} , x) }{ \partial b_{i, r} } \right| \leq \left( 2c_1 \epsilon_a \sqrt{ 2 \log m } \right) \tau  \left( \Tilde{L}_1 + \Tilde{L}_2 | \wt{i, r, i}{t} | +  2 c_2 \Tilde{L}_2 \sigma_{wb} \sqrt{ 2 \log \rb{ m d } }  \right) .
\end{align}
For projected gradient descent, summing from time step $j=0$ to $j=t-1$, we get
\begin{align*}
    | \bt{i, r}{t} | \leq & \; \eta \sum_{j=0}^{t-1} \left| \frac{ \partial L_G(\ft{j}, \xt{j} ) }{ \partial b_r } \right| \\ 
    =& \; 2 \eta c_1 \epsilon_a \tau \sqrt{ 2 \log m } \left( \Tilde{L}_1 + 2 c_2 \Tilde{L}_2 \sigma_{wb} \sqrt{ 2 \log \rb{ m d} }   \right) t + 2 \eta c_1 \epsilon_a \tau \Tilde{L}_2 \sqrt{2 \log m}  \left( \sum_{j=0}^{t-1} | \wt{i, r, i}{j} | \right) \\
    \leq & \; 2 \eta c_1 \epsilon_a \tau \sqrt{ 2 \log m } \left( \Tilde{L}_1 + 2 c_2 \Tilde{L}_2 \sigma_{wb} \sqrt{ 2 \log \rb{ m d} }   \right) t + 2 \eta c_1 \epsilon_a \tau \Tilde{L}_2 \sqrt{2 \log m}  \left( \sum_{j=0}^{t-1} \norm{ \wt{i, r}{t} }_2 \right) .
\end{align*}
Using Eq.(\ref{eq:w-sum-bound}), we get
\begin{align*}
    | \bt{i, r}{t} | \leq & \; 2 \eta c_1 \epsilon_a \tau \sqrt{ 2 \log m }  \left( \Tilde{L}_1 +  2 c_2 \Tilde{L}_2  \sigma_{wb} \sqrt{ 2 \log \rb{ m d} }   \right) t \\ 
    +& \;  2 \eta c_1 \epsilon_a \Tilde{L}_2 \sqrt{2 \log m}  \left( \frac{  \Tilde{L}_1 + \Tilde{L}_2 +  2 c_2  \Tilde{L}_2 \sigma_{wb}  \sqrt{2 \log \rb{ m d}}  }{ 2 \eta c_1 \epsilon_a \sqrt{2 \log m} \Tilde{L}_2^2  } \right) \left(  \left(1 + 2 \eta c_1 \tau \epsilon_a \sqrt{2 \log m} \Tilde{L}_2    \right)^t - 1 \right) \\
    = & \; 2 \eta c_1 \epsilon_a \tau \sqrt{ 2 \log m } \left( \Tilde{L}_1 +  2 c_2 \Tilde{L}_2 \sigma_{wb} \sqrt{ 2 \log \rb{ m d} }   \right) t \\ 
    +& \; \left( \frac{  \Tilde{L}_1 + \Tilde{L}_2 +  2 c_2 \Tilde{L}_2 \sigma_{wb}  \sqrt{2 \log \rb{ m d} }  }{ \Tilde{L}_2 } \right) \left(  \left(1 + 2 \eta c_1 \epsilon_a \tau \sqrt{2 \log m} \Tilde{L}_2 \right)^t - 1 \right).
\end{align*}
This completes the proof.
\end{proof}

Define $\lbd{w}{t}$ and $\lbd{b}{t}$ as upper bound on $\norm{ \wt{i, r}{t} }_2$ and $\abs{ \bt{i, r}{t} }$:
\begin{align*}
\lbd{w}{t} = & \; \left( \frac{ \Tilde{L}_1 + \Tilde{L}_2 +  2 c_2 \Tilde{L}_2 \sigma_{wb}  \sqrt{2 \log \rb{ m d} } }{ \Tilde{L}_2  } \right) \left( \left(1 + 2 \eta c_1 \epsilon_a \tau \sqrt{2 \log m} \Tilde{L}_2   \right)^t - 1 \right), \\
\lbd{b}{t} = & \; 2 \eta c_1 \epsilon_a \tau \sqrt{ 2 \log m } \left( \Tilde{L}_1 +  2 c_2 \Tilde{L}_2 \sigma_{wb} \sqrt{ 2 \log \rb{m d} }   \right) t \\ 
    &+ \; \left( \frac{  \Tilde{L}_1 + \Tilde{L}_2 +  2 c_2 \Tilde{L}_2 \sigma_{wb}  \sqrt{2 \log \rb{m d}}  }{ \Tilde{L}_2 } \right) \left(  \left(1 + 2 \eta c_1 \epsilon_a \tau \sqrt{2 \log m} \Tilde{L}_2 \right)^t - 1 \right).
\end{align*}

\begin{lemma}
\label{lemma:bound-on-lambda}
For any $\epsilon > 0$ , target function $\tarfunc{*}$ with some complexity measure $C( \tarfunc{*} )$, any $\sigma_{wb}$ which satisfy $  \frac{1}{ \sqrt{m} } \leq \sigma_{wb} \leq  \frac{\epsilon}{  C \rb{ \tarfunc{*} } \sqrt{ \log \rb{ m d } } } $, any hidden layer size $m \geq \Omega \rb{ \text{poly} \rb{ C( \tarfunc{*} ), d, \frac{1}{\epsilon}} }$, any learning rate $\eta \leq  \frac{ c_9 \epsilon}{ m \tau \epsilon_a^2 \log m } $ and $T \leq \frac{ c_{10} C( \tarfunc{*} ) }{ \epsilon^2 } $, with at least $1-\frac{d}{c_1} - \frac{d}{c_2}$ probability, we get
\begin{align*}
\lbd{w}{t} &\leq \left( \Tilde{L}_1 + \Tilde{L}_2 +  2 c_2  \Tilde{L}_2 \sigma_{wb}  \sqrt{2 \log \rb{ m d } }  \right) \rb{ \frac{ 4 \sqrt{2} c_1 c_9 c_{10} C( \tarfunc{*} )  }{ m \epsilon \epsilon_a \sqrt{\log m} } } \\
\lbd{b}{t} & \leq \left(  3 \Tilde{L}_1 + 2 \Tilde{L}_2 + 6 c_2 \Tilde{L}_2 \sigma_{wb}  \sqrt{2 \log \rb{ m d } } \right)  \rb{  \frac{ 2 \sqrt{2} c_1 c_9 c_{10} C( \tarfunc{*} ) }{ m \epsilon_a \epsilon \sqrt{ \log m } } }
\end{align*}
\end{lemma}

\begin{proof}
To simplify expression of $\lbd{w}{t}$, we will use Fact \ref{fact:bernouli-inequality}. First, we will check the condition for Fact \ref{fact:bernouli-inequality}:
\begin{align*}
2 \eta c_1 \epsilon_a \tau \sqrt{2 \log m} \Tilde{L}_2 \rb{t-1} &\leq 2 \eta c_1 \epsilon_a \tau \sqrt{2 \log m} \Tilde{L}_2 \rb{ T -1} \\
&\leq 2 \rb{ \frac{c_9 \epsilon}{ m \tau \epsilon_a^2 \log m }   } c_1 \epsilon_a \tau \sqrt{2 \log m} \rb{ \frac{ c_{10} C( \tarfunc{*} ) }{ \epsilon^2 }   - 1 } \\
&= \frac{ 2 \sqrt{2} c_1 c_9 c_{10} C( \tarfunc{*} ) }{ \epsilon_a \epsilon m \sqrt{ \log m } }.
\end{align*}
Choosing sufficiently high $m$ such that $m \geq \Omega \rb{ \text{poly} \rb{ C( \tarfunc{*} ), d, \frac{1}{\epsilon}} }$, we get 
\begin{align*}
2 \eta c_1 \epsilon_a \tau \sqrt{2 \log m} \Tilde{L}_2 \rb{ t - 1} \leq  0.5.
\end{align*}
By choosing sufficiently high $m$, the condition of Fact \ref{fact:bernouli-inequality} satisfies. Now, simplifying expression of $\lbd{w}{t}$ using Fact \ref{fact:bernouli-inequality}, we get
\begin{align*}
\lbd{w}{t} &= \left( \frac{ \Tilde{L}_1 + \Tilde{L}_2 +  2 c_2 \Tilde{L}_2 \sigma_{wb}  \sqrt{2 \log \rb{m d} } }{ \Tilde{L}_2 } \right) \left( \left(1 + 2 \eta c_1 \epsilon_a \tau \sqrt{2 \log m} \Tilde{L}_2 \right)^t - 1 \right) \\
&\leq \left( \frac{ \Tilde{L}_1 + \Tilde{L}_2 +  2 c_2 \Tilde{L}_2 \sigma_{wb}  \sqrt{2 \log \rb{m d} } }{ \Tilde{L}_2 } \right) \left(  4 \eta c_1 \epsilon_a \tau \sqrt{2 \log m} \Tilde{L}_2  t \right) \\
&\leq \left( \frac{ \Tilde{L}_1 + \Tilde{L}_2 +  2 c_2 \Tilde{L}_2 \sigma_{wb}  \sqrt{2 \log \rb{m d} } }{ \Tilde{L}_2 } \right) \left(  4 \eta c_1 \epsilon_a \tau \sqrt{2 \log m} \Tilde{L}_2  T \right) \\
&\leq \left( \frac{ \Tilde{L}_1 + \Tilde{L}_2 +  2 c_2 \Tilde{L}_2 \sigma_{wb}  \sqrt{2 \log \rb{m d} } }{ \Tilde{L}_2 } \right) \rb{ 4 c_1 \epsilon_a \tau \Tilde{L}_2 \sqrt{2 \log m} \rb{  \frac{c_9 \epsilon}{ m \tau \epsilon_a^2 \log m } } \rb{ \frac{ c_{10} C( \tarfunc{*} ) }{ \epsilon^2 } } } \\ 
&= \left( \frac{ \Tilde{L}_1 + \Tilde{L}_2 +  2 c_2  \Tilde{L}_2 \sigma_{wb}  \sqrt{2 \log \rb{m d} } }{ \Tilde{L}_2 } \right) \rb{ 4 c_1 \epsilon_a \tau \Tilde{L}_2 \sqrt{2 \log m} \rb{  \frac{c_9 \epsilon}{ m \tau \epsilon_a^2 \log m } } \rb{ \frac{ c_{10} C( \tarfunc{*} ) }{ \epsilon^2 } } } \\
&= \left( \Tilde{L}_1 + \Tilde{L}_2 +  2 c_2 \Tilde{L}_2 \sigma_{wb}  \sqrt{2 \log \rb{m d} }  \right) \rb{ \frac{ 4 \sqrt{2} c_1 c_9 c_{10} C( \tarfunc{*} )  }{ m \epsilon \epsilon_a \sqrt{\log m} } }.
\end{align*}
Simplifying expression of $\lbd{b}{t}$ in simillar manner as $\lbd{w}{t}$, we get
\begin{align*}
\lbd{b}{t} = & \; 2 \eta c_1 \epsilon_a \tau \sqrt{ 2 \log m } \left( \Tilde{L}_1 +  2 c_2 \Tilde{L}_2 \sigma_{wb} \sqrt{2 \log \rb{m d} }   \right) t \\ 
    +& \; \left( \frac{  \Tilde{L}_1 + \Tilde{L}_2 +  2 c_2 \Tilde{L}_2 \sigma_{wb}  \sqrt{2 \log \rb{m d} }  }{ \Tilde{L}_2 } \right) \left(  \left(1 + 2 \eta c_1 \epsilon_a \tau \sqrt{2 \log m} \Tilde{L}_2 \right)^t - 1 \right) \\
    \leq & \; 2 c_1 \epsilon_a \tau \sqrt{ 2 \log m } \left( \Tilde{L}_1 +  2 c_2 \Tilde{L}_2 \sigma_{wb} \sqrt{2 \log \rb{m d} } \right) \rb{  \frac{c_9 \epsilon}{ m \tau \epsilon_a^2 \log m } } \rb{ \frac{ c_{10} C( \tarfunc{*} ) }{ \epsilon^2 } }  \\ 
    & + \; \left( \frac{  \Tilde{L}_1 + \Tilde{L}_2 +  2 c_2 \Tilde{L}_2 \sigma_{wb}  \sqrt{2 \log \rb{m d} }  }{ \Tilde{L}_2 } \right) \left(  4 c_1 \epsilon_a \tau \sqrt{2 \log m} \Tilde{L}_2  \rb{  \frac{c_9 \epsilon}{ m \tau \epsilon_a^2 \log m } } \rb{ \frac{ c_{10} C( \tarfunc{*} ) }{ \epsilon^2 } } \right) \\
    = & \;  \left( \Tilde{L}_1 +  2 c_2 \Tilde{L}_2 \sigma_{wb} \sqrt{2 \log \rb{m d} }  \right) \rb{  \frac{ 2 \sqrt{2} c_1 c_9 c_{10} C( \tarfunc{*} ) }{ m \epsilon_a \epsilon \sqrt{ \log m } } } \\ 
    &+ \; \left(   \Tilde{L}_1 + \Tilde{L}_2 +  2 c_2 \Tilde{L}_2 \sigma_{wb}  \sqrt{2 \log \rb{m d} } \right)  \rb{  \frac{ 4 \sqrt{2} c_1 c_9 c_{10} C( \tarfunc{*} ) }{ m \epsilon_a \epsilon \sqrt{ \log m } } }   \\
    = & \; \rb{  \frac{ 2 \sqrt{2} c_1 c_9 c_{10} C( \tarfunc{*} ) }{ m \epsilon_a \epsilon \sqrt{ \log m } } } \left(  3 \Tilde{L}_1 + 2 \Tilde{L}_2 + 6 c_2 \Tilde{L}_2 \sigma_{wb}  \sqrt{2 \log \rb{m d} } \right) .
\end{align*}
\end{proof}

\begin{lemma} (Coupling between neural network and pseudo network)
\label{lemma:coupling-function-values-cnf}
For every $x$ with $\norm{x}_2 \leq 1$, every $i \in [d]$ and every time step $t \leq T$, with probability at least $1 - \frac{d}{c_1} - \frac{d}{c_2} $ over random initialization, we have
\begin{align*}
    \abs{ \nnit{i}{(t)}  - \pnit{i}{(t)} } \leq {2 c_1 \epsilon_a \tau m \sqrt{2 \log m}  } \left( \rb{ \lbd{w}{t} }^2 + { \lbd{b}{t} }^2 \right) 
\end{align*}{}
\end{lemma}

\begin{proof}
 Bounding difference between $\nnit{i}{(t)}$ and $\pnit{i}{(t)}$, we get
\begin{align*}
    \abs{\nnit{i}{(t)} - \pnit{i}{(t)} } & =  \Bigg|\tau \sum_{r=1}^m \initam{i}{r} \tanh ( \neurvalm{i}{(t)}{\vecx{i}} )  \\
     &- \tau  \sum_{r=1}^m \initam{i}{r} \left( \tanh ( \neuralvalzeromwv{i} ) + \initam{i}{r} \tanh'( \neuralvalzeromwv{i} ) ( \dotp{ \wt{i, r}{t} }{ \vecx{i} } + \bt{i, r}{t} ) \right) \Bigg| \\
   & = \left| \frac{\tau}{2} \sum_{r=1}^m \initam{i}{r} \tanh''(\xi_r) \left( \dotp{ \wt{i, r}{t} }{ \vecx{i} } + \bt{i, r}{t} \right)^2 \right|
\end{align*}
for some $\xi_r \in \R$. Using $|\tanh''( \xi_{r} )| \leq 1$ and $( \dotp{ \wt{i, r}{t} }{ \vecx{i} } + \bt{i, r}{t} )^2 \leq 2 \rb{ \dotp{ \wt{i, r}{t} }{ \vecx{i} }^2 + \rb{ \bt{i, r}{t} }^2 }$, we get
\begin{align*}
    \abs{\nnit{i}{(t)} - \pnit{i}{(t)} } & \leq  \tau \sum_{r=1}^m \initam{i}{r} \rb{ \dotp{ \wt{i, r}{t} }{ \vecx{i} }^2 + \rb{ \bt{i, r}{t} }^2 }.
\end{align*}{}
Using Lemma~\ref{lemma:concentration-folded-normal} and using $\norm{x}_2 \leq 1$, with at least $1 - \frac{1}{ c_1 }$ probability, we have
\begin{align*}
    \abs{\nnit{i}{(t)} - \pnit{i}{(t)} } &\leq {2 c_1 \epsilon_a \tau \sqrt{2 \log m}  } \sum_{r=1}^m \rb{ \norm{ \wt{i, r}{t} }_2^2 + \rb{ \bt{r}{t} }^2 } \\
     &\leq {2 c_1 \epsilon_a \tau \sqrt{2 \log m}  } \left( \| \Wt{t}_i \|_{2, 2}^2 + \| \Bt{t}_i \|_2^2 \right)  \\
     &\leq {2 c_1 \epsilon_a \tau m \sqrt{2 \log m}  } \left( \rb{ \lbd{w}{t} }^2 + \rb{ \lbd{b}{t} }^2 \right).
\end{align*}
Using union bound for all $i \in [d]$, we complete the proof.
\end{proof}

\begin{lemma}
\label{lemma:linear-pseudo-linear}
For any $\epsilon \in (0, \frac{1}{d^3})$, every $i \in [d]$ and every time step $t \leq T$, any target function $\tarfunc{*}$ with some complexity measure of target function $C( \tarfunc{*} )$, any variance $\sigma_{wb}$ with $\frac{1}{\sqrt{m}} \leq \sigma_{wb} \leq \frac{\epsilon^2}{ C(\tarfunc{*}) \sqrt{ \log \rb{ m d } } }$, any hidden layer size $m \geq \Omega \rb{ \text{poly} \rb{ C( \tarfunc{*} ), d, \frac{1}{\epsilon}} }$, choosing learning rate $\eta = O \rb{ \frac{\epsilon}{ m \tau \epsilon_a^2 \log m } } $ and $T= O \rb{ \frac{ C( \tarfunc{*} ) }{ \epsilon^2 } }$, with probability at least $1 - \frac{d}{c_1} - \frac{d}{c_2} - \frac{d}{c_3}$ over random initialization, we get
\begin{align*}
\abs{ \pnilt{\ell}{i}{(t)} - \tau \sum_{r=1}^m \initam{i}{r} \rb{ \dotp{ \wt{i, r}{t} }{ \vecx{i} } + \bt{i, r}{t} } } \leq O( \epsilon ).
\end{align*}
\end{lemma}
\begin{proof}
Recalling the definition of $\pnilt{\ell}{i}{(t)}$:
\begin{align*}
\pnilt{\ell}{i}{(t)}= \tau \sum_{r=1}^m \initam{i}{r} \rb{ \tanh' ( \neuralvalzeromwv{i} ) \rb{ \dotp{ \wt{r}{t} }{ \vecx{i}} + \bt{i, r}{t} } }.
\end{align*}
Subtracting the linear function from $\pnilt{\ell}{i}{(t)}$ will give us the following:
\begin{align}
\label{eq:linear-pseudo-linear}
 \nonumber \abs{ \pnilt{\ell}{i}{(t)} - \tau \sum_{r=1}^m \initam{i}{r} \rb{ \dotp{ \wt{i, r}{t} }{ \vecx{i} } + \bt{i, r}{t} } } &\leq \abs{ \tau \sum_{r=1}^m \initam{i}{r} \rb{ \rb{ \tanh' ( \neuralvalzeromwv{i} ) - 1 } \rb{ \dotp{ \wt{i, r}{t} }{ \vecx{i} } + \bt{i, r}{t} } } } \\ 
 &\leq  \tau \sum_{r=1}^m  \initam{i}{r} \abs{  \tanh' ( \neuralvalzeromwv{i} ) - 1  }  \abs{ \dotp{ \wt{i, r}{t} }{ \vecx{i} } + \bt{i, r}{t}  }  .
\end{align}
First, we will try to find upper bound on $\abs{  \tanh' ( \neuralvalzeromwv{i} ) - 1  }$:
\begin{align*}
\abs{  \tanh' ( \neuralvalzeromwv{i} ) - 1  }  &= \abs{  \tanh' ( \neuralvalzeromwv{i} )   - \tanh'(0)  } \stackrel{ ( \RomanNumLow{1} ) }{ \leq } \abs{ \neuralvalzeromwv{i}  }, 
\end{align*}
where inequality $(\RomanNumLow{1})$ follows from 1-Lipschitz continuity of $\tanh'(\cdot)$ function. Using Lemma \ref{lemma:concentration-folded-normal} on $\initwm{i}{r}$ and $\initbm{i}{r}$, with probability at least $1 - \frac{1}{c_2} - \frac{1}{c_3}$, we get that
\begin{align*}
\abs{ \tanh' ( \neuralvalzeromwv{i} )  - 1 } \leq 2 c_2 \sigma_{wb} \sqrt{2 \log \rb{ md } } + 2 c_3\sigma_{wb} \sqrt{2 \log m }.
\end{align*}
Using above inequality in Eq. (\ref{eq:linear-pseudo-linear}), with probability at least $1 - \frac{1}{c_1} - \frac{1}{c_3}$ over random initialization, we get
\begin{align*}
\abs{ \pnilt{\ell}{i}{(t)} - \tau \sum_{r=1}^m \initam{i}{r} \rb{ \dotp{ \wt{i, r}{t} }{ \vecx{i} } + \bt{i, r}{t} } } & \leq \tau \sum_{r=1}^m \initam{i}{r} \abs{ \tanh' ( \neuralvalzeromwv{i} )  - 1 } \abs{ \dotp{ \wt{i, r}{t} }{ \vecx{i} } + \bt{i, r}{t}  }    \\
&\leq \tau \sum_{r=1}^m \initam{i}{r} \rb{  2 c_2 \sigma_{wb} \sqrt{2 \log \rb{ md } } + 2 c_3\sigma_{wb} \sqrt{2 \log m } } \abs{ \dotp{ \wt{i, r}{t} }{ \vecx{i} } + \bt{i, r}{t}  }  .
\end{align*}
Using Lemma \ref{lemma:concentration-folded-normal} and Lemma \ref{lemma:bound-wt-bt-cnf}, with probability at least $1 - \frac{1}{c_1} - \frac{1}{c_2} - \frac{1}{c_3}$, we get
\begin{align*}
&\abs{ \pnilt{\ell}{i}{(t)} - \tau \sum_{r=1}^m \initam{i}{r} \rb{ \dotp{ \wt{i, r}{t} }{ \vecx{i} } + \bt{i, r}{t} } } \\ 
& \leq \tau \sum_{r=1}^m \initam{i}{r} \abs{ \tanh' ( \neuralvalzeromwv{i} )  - 1 } \abs{ \dotp{ \wt{i, r}{t} }{ \vecx{i} } + \bt{i, r}{t}  }    \\
& \leq m \tau \rb{ 2 c_1 \epsilon_a \sqrt{2 \log m } } \rb{ 2 c_2 \sigma_{wb} \sqrt{2 \log \rb{ md } } + 2 c_3\sigma_{wb} \sqrt{2 \log m } } \rb{ \lbd{w}{t} + \lbd{b}{t} } \\
& \leq 8 c_1 \rb{ c_2 \sqrt{ \log \rb{ md } } + c_3 \sqrt{ \log m } } \epsilon_a \sigma_{wb} m \tau \sqrt{ \log m } \rb{ \lbd{w}{t} + \lbd{b}{t} }.
\end{align*}
Using bound on $\lbd{w}{t}$ and $\lbd{b}{t}$ from Lemma \ref{lemma:bound-on-lambda}, we get
\begin{align*}
&\abs{ \pnilt{\ell}{i}{(t)} - \tau \sum_{r=1}^m \initam{i}{r} \rb{ \dotp{ \wt{i, r}{t} }{ \vecx{i} } + \bt{i, r}{t} } } \\ 
\leq & \; 8 c_1 \rb{ c_2 \sqrt{ \log \rb{ md } } + c_3 \sqrt{ \log m } } \epsilon_a \sigma_{wb} m \tau \sqrt{ \log m } \Bigg(  \rb{  \frac{ 2 \sqrt{2} c_1 c_9 c_{10} C( \tarfunc{*} ) }{ m \epsilon_a \epsilon \sqrt{ \log m } } } \Big(  5 \Tilde{L}_1 + 4 \Tilde{L}_2  \\
&+ 10 c_2 \Tilde{L}_2 \sigma_{wb}  \sqrt{2 \log \rb{ m d }  } \Big)  \Bigg) \\
\leq & \; \frac{ 16 \sqrt{2} c_1^2 c_9 c_{10} \rb{ c_2 \sqrt{ \log \rb{ md } } + c_3 \sqrt{ \log m } } \sigma_{wb} \tau C( \tarfunc{*} ) }{ \epsilon } \Big(  5 \Tilde{L}_1 + 4 \Tilde{L}_2  + 10 c_2 \Tilde{L}_2 \sigma_{wb}  \sqrt{2 \log \rb{ m d } } \Big).
\end{align*}
Using $\sigma_{wb} \leq \frac{ \epsilon^2 }{ C( \tarfunc{*} ) \sqrt{ \log \rb{ m d } }  } $ and re-scaling $\epsilon$ by $\frac{\epsilon}{d^3}$, we get
\begin{align*}
\abs{ \pnilt{\ell}{i}{(t)} - \tau \sum_{r=1}^m \initam{i}{r} \rb{ \dotp{ \wt{i, r}{t} }{ \vecx{i} } + \bt{i, r}{t} } } &\leq O \rb{ \tau \epsilon }.
\end{align*}
Using $\tau \leq 1$ for $\sigma_{wb} \geq \frac{1}{\sqrt{m}}$, we get
\begin{align*}
\abs{ \pnilt{\ell}{i}{(t)} - \tau \sum_{r=1}^m \initam{i}{r} \rb{ \dotp{ \wt{i, r}{t} }{ \vecx{i} } + \bt{i, r}{t} } } &\leq O \rb{ \epsilon }.
\end{align*}

\end{proof}

\begin{theorem}
\label{theorem:smaller-variance-problem-cnf-appendix}

For any $\epsilon \in (0, \frac{1}{d^3})$, any $i \in [d]$, any target function $\tarfunc{*}$ with some complexity measure of target function $C( \tarfunc{*} )$, any $\sigma_{wb}$ which satisfy $  \frac{1}{ \sqrt{m} } \leq \sigma_{wb} \leq  \frac{\epsilon}{  C \rb{ \tarfunc{*} } \sqrt{ \log \rb{ m d } } } $, any hidden layer size $m \geq \Omega \rb{ \text{poly} \rb{ C( \tarfunc{*} ), d, \frac{1}{\epsilon}} }$, choosing normalization constant $\tau$ such that $\abs{P_c(x)} \leq O( \epsilon )$,  learning rate $\eta = O \rb{ \frac{\epsilon}{ m \tau \epsilon_a^2 \log m } } $ and $T= O \rb{ \frac{ C( \tarfunc{*} ) }{ \epsilon^2 } }$, with probability at least 0.9 over random initialization, Projected SGD on neural network after $T$ iterations 
\begin{align}
\label{eq:nn-optimization-cnf}
\abs{ \nnit{i}{(T)} - \rb{ \dotp{ \alpha_i }{ \vecx{i} } + \beta_i } } \leq O \rb{ \epsilon },
\end{align}
where $\alpha_i$ and $\beta_i$ are given by
\begin{align*}
\alpha_i &= \tau \sum_{r=1}^m \initam{i}{r} \wt{i, r}{T} \quad	\text{and} \quad
\beta_i = \tau \sum_{r=1}^m \initam{i}{r} \bt{i, r}{T} .
\end{align*}
\end{theorem}
\begin{proof}
By decomposing the difference between $\nnit{i}{(T)}$ and $\rb{ \dotp{ \alpha_i }{ \vecx{i} } + \beta_i }$ into two parts, we get
\begin{align}
\label{eq:diff-N-P}
\abs{ \nnit{i}{(T)} - \rb{ \dotp{ \alpha }{ \vecx{i} } + \beta } } \leq \underbrace{ \abs{ \nnit{i}{(T)}  - \pnit{i}{(T)}  } }_{ \RomanNumUp{1} } + \underbrace{ \abs{ \pnit{i}{(T)} - \rb{ \dotp{ \alpha_i }{ \vecx{i} } + \beta_i } } }_{ \RomanNumUp{2} }.
\end{align}
Using Lemma \ref{lemma:coupling-function-values-cnf}, we can bound $\RomanNumUp{1}$:
\begin{align*}
\RomanNumUp{1} &\leq {2 c_1 \epsilon_a \tau m \sqrt{2 \log m}  } \left( \rb{ \lbd{w}{t} }^2 + \rb{ \lbd{b}{t} }^2 \right) \\
&\leq {2 c_1 \epsilon_a \tau m \sqrt{2 \log m}  } \left( \rb{ \lbd{w}{t} }^2 + \rb{ \lbd{b}{t} }^2 \right) \\
&\leq {2 c_1 \epsilon_a \tau m \sqrt{2 \log m}  } \rb{ \frac{ 2 \sqrt{2} c_1 c_9 c_{10} C( \tarfunc{*} )  }{ m \epsilon \epsilon_a \sqrt{\log m} } }^2 \Bigg( 4 \left( \Tilde{L}_1 + \Tilde{L}_2 +  2 c_2 \Tilde{L}_2 \sigma_{wb}  \sqrt{2 \log \rb{ m d } }  \right)^2 \\ 
&+ \left(  3 \Tilde{L}_1 + 2 \Tilde{L}_2 + 6 c_2 \Tilde{L}_2 \sigma_{wb}  \sqrt{2 \log \rb{ m d } } \right)^2  \Bigg).
\end{align*}
Choosing sufficienty high $m$ such that $m \geq \Omega \rb{ \text{poly} \rb{ C( \tarfunc{*} ), \frac{1}{\epsilon}} }$, we get
\begin{align*}
\RomanNumUp{1} \leq O \rb{ \epsilon }
\end{align*}
To bound $\RomanNumUp{2}$, we use Lemma \ref{lemma:linear-pseudo-linear}. 
\begin{align*}
\RomanNumUp{2} &\leq \abs{ P_c( \vecx{i} ) } + \abs{ \pnilt{\ell}{i}{(t)} - \rb{ \dotp{ \alpha_i }{ \vecx{i} } + \beta_i } } \leq O(\epsilon)
\end{align*}
Using Eq. (\ref{eq:diff-N-P}), we get
\begin{align*}
\abs{ \nnit{i}{(t)} - \rb{ \dotp{ \alpha_i }{ \vecx{i} } + \beta_i } } \leq O ( \epsilon )
\end{align*}
\end{proof}

\subsection{Problem in optimization for larger variance initialization case}
\label{subsec:large-variance-cnf-problem}
In this section, we will provide details about the problem for larger variance initialization case. Recall that $P_c \rb{ \vecx{i} }$ only depends upon initialization and does not depend on $\theta_i$. Hence, it can not approximate the target function after the training, therefore $\pnil{\ell}{i}$ needs to approximate target function with $P_c \rb{ \vecx{i} }$ subtracted but in this case, we prove in Theorem \ref{theorem:larger-variance-cnf-problem-appendix} that if norm of change in weights $\| \tht{T} \|_{2, 1}$ is small then $\abs{ \pnil{\ell}{i} }$ is very small for sufficiently large $m$; therefore, it can not approximate every target function. We also provide reasons and details in Lemma \ref{lemma:CNF-problem-coupling-problem-higher-var} about the requirement of small norm of change in weights $\| \tht{T} \|_{2, 1}$. In short, small norm of change in weights is required to maintain coupling between neural networks and pseudo networks. For large variance initialization case, we have \todo{Need to revisit this para}

\begin{theorem} (small value of $\pnilt{\ell}{i}{(T)}$)
\label{theorem:larger-variance-cnf-problem-appendix}
For any standard deviation $ \frac{\epsilon^2}{ C ( \tarfunc{*} ) \sqrt{ \log m } }   \leq \sigma_{wb} \leq 1$, for any $i \in [d]$, any constant $c_8 > 0$ and any $\eta > 0$, $T>1$, if upper bound on norm of change of parameters is given by
\begin{align*}
\| \tht{T}_i \|_{2, 1} &\leq  O \rb{ \frac{1}{ d^2 \epsilon_a \sigma_{wb} \tau m^{c_8} \log m } } ,
\end{align*}
then for all $x \in \R^d$ with $\norm{x}_2 \leq 1$, with probabillity at least $0.99$, we have
\begin{align*}
\abs{ \pnilt{\ell}{i}{(T)} } \leq \frac{1}{3 \sqrt{2} c_2 \sigma_{wb} m^{c_8} \sqrt{\log m} } = O \rb{ \frac{1}{d \sigma_{wb} m^{c_8} \sqrt{\log \rb{m d} } } }.
\end{align*}
Given upper bound on $\| \tht{T} \|_{2, 1}$ is necessary to ensure closeness between neural network and pseudo network (More details given in Lemma \ref{lemma:CNF-problem-coupling-problem-higher-var}).
\end{theorem}

\begin{proof}
Using the definition of $\pnilt{\ell}{i}{(t)}$, we get
\begin{align*}
\abs{ \pnilt{\ell}{i}{(t)} } &= \abs{ \tau \sum_{r=1}^m \initam{i}{r} \tanh' \rb{ \neuralvalzeromwv{i} } \rb{ \dotp{ \wt{i, r}{t} }{ x } + \bt{i, r}{t} } } \\
& \stackrel{( \RomanNumLow{1} )}{\leq} \tau \rb{ 2 c_1 \epsilon_a \sqrt{2 \log m} } \sum_{r=1}^m \rb{ \norm{ \wt{i, r}{t} }_2 + \abs{ \bt{i, r}{t} }  } \\
&\leq \tau  \rb{ 2 c_1 \epsilon_a \sqrt{2 \log m} }  \rb{ \norm{ \Wt{t}_i }_{2, 1} + \norm{ \Bt{t}_i }_1  },
\end{align*}
where inequality $(\RomanNumLow{1})$ follows from Lemma \ref{lemma:concentration-folded-normal} with at least $1-\frac{1}{c_1}$ probability. Using upper bound on $\| \tht{T}_i \|_{2, 1}$ from the theorem statement, with at least $1 - \frac{1}{c_1}$ probability, we get 
\begin{align*}
\abs{\pnilt{\ell}{i}{(t)} } &\leq  \tau \rb{ 2 c_1 \epsilon_a \sqrt{2 \log m} }  \rb{ \| \Wt{T}_i \|_{2, 1} + \| \Bt{T}_i \|_{1}  } \\
&\leq \tau \rb{ 2 c_1 \epsilon_a \sqrt{2 \log m} }  \rb{ 2 \| \tht{T}_i \|_{2, 1}  } \\
&\leq \tau \rb{ 2 c_1 \epsilon_a \sqrt{2 \log m} }  \rb{ \frac{1}{12 c_1 c_2 \epsilon_a \sigma_{wb} \tau m^{c_8} \sqrt{ \log m \log \rb{m d} } } } \\
& \leq \frac{1}{3 \sqrt{2} c_2 \sigma_{wb} m^{c_8} \sqrt{\log \rb{ m d} } }.
\end{align*}
This completes the proof.
\end{proof}

Recall that we denote derivative $\frac{\partial \funci{i} }{ \partial x_i }$ as $\dfi{i}$. Similarly, we use $\dgi{i}$ to derivative of $g_i \rb{ \vecx{i} }$.

\begin{lemma} \label{lemma:CNF-problem-coupling-problem-higher-var} (Requirement of having small $L_{2, 1}$-norm of change in weights $\| \tht{T}_i \|_{2, 1}$ ) For any constant $c_8 > 0$, for all $i \in [d]$, if following bound either on $ \| \tht{T}_i \|_{2, 1}$ holds,
\begin{align*}
\| \tht{T}_i \|_{2, 1} & = \omega \rb{ \frac{1}{ d^2 \epsilon_a \sigma_{wb} \tau m^{c_8} \sqrt{ \log \rb{ m } \log \rb{ m d } } } } 
\end{align*}
then, with at least $0.98$ probability, coupling between $\dfit{i}{(t)}$ and $\dgit{i}{(t)}$ can be lost. More precisely, 
\begin{align*}
 \abs{ \dfit{i}{(t)} - \dgit{i}{(t)} } \leq \omega \rb{ \frac{1}{m^{c_8}} }
\end{align*}
\end{lemma}

\begin{proof}
First, we will find upper bound on difference between $\dfit{i}{(t)}$ and $\dgit{i}{(t)}$:
\begin{align*}
     \abs{ \dfit{i}{(t)} - \dgit{i}{(t)} } &= \Bigg| \tau \sum_{r=1}^m \initam{i}{r} \big( ( \wt{i, r, i}{t} + \initwm{i}{r, i}) \Big( \tanh'( \neurvalm{i}{(t)}{ \vecx{i} } ) \\ 
     &- \tanh'( \neuralvalzeromwv{i} ) \Big) - \tanh''( \neuralvalzeromwv{i} ) \left( \initwm{i}{r, i} ( \dotp{ \wt{i, r}{t} }{ \vecx{i} } + \bt{i, r}{t} ) \big) \right)  \Bigg| \\
     &\leq  \tau \sum_{r=1}^m 2 \initam{i}{r} \abs{ \wt{i, r, i}{t} + \initwm{i}{r, i} } \abs{ \dotp{ \wt{i, r}{t} }{ \vecx{i} } + \bt{i, r}{t} }  + \tau \sum_{r=1}^m \initam{i}{r} \initwm{i}{r, i} \abs{ \dotp{ \wt{i, r}{t} }{ \vecx{i} } + \bt{i, r}{t} } \\
     &= \tau \sum_{r=1}^m \initam{i}{r} \abs{ \dotp{ \wt{i, r}{t} }{ \vecx{i} } + \bt{i, r}{t} } \rb{ 2\abs{ \wt{i, r, i}{t} + \initwm{i}{r, i} } +  \initwm{i}{r, i} } \\
	&= \tau \sum_{r=1}^m \initam{i}{r} \abs{ \dotp{ \wt{i, r}{t} }{ \vecx{i} } + \bt{i, r}{t} } \rb{ 2| \wt{i, r, i}{t} | + 3 \initwm{i}{r, i} } \\
	&\leq  \tau \sum_{r=1}^m \initam{i}{r} \rb{ \norm{ \wt{i, r}{t} }_2 + \abs{ \bt{i, r}{t} } } \rb{ 2 \norm{ \Wt{t}_i }_{\infty, \infty} + 6 c_2 \sigma_{wb} \sqrt{ 2 \log \rb{m d} } }  \\
     & \stackrel{(\RomanNumLow{1})}{\leq}  \tau \rb{ 2 c_1 \epsilon_a \sqrt{2 \log m } } \rb{ \norm{ \Wt{t}_i }_{2, 1} + \norm{ \Bt{t}_i }_1 } \rb{ 2 \norm{ \Wt{t}_i }_{\infty, \infty} + 6 c_2 \sigma_{wb} \sqrt{ 2 \log \rb{m d} } },
\end{align*}
where inequality $(\RomanNumLow{1})$ follows from Lemma \ref{lemma:concentration-folded-normal} with probability at least $1-\frac{1}{c_1} - \frac{1}{c_2}$. Using bounds on norm $\| \tht{T}_i \|_{2, 1}$, we get  
\begin{align*}
\abs{ \dfit{i}{(t)} - \dgit{i}{(t)} } \leq & \; \tau \rb{ 2 c_1 \epsilon_a \sqrt{2 \log m } } \rb{ \norm{ \Wt{t}_i }_{2, 1} + \norm{ \Bt{t}_i }_1 } \rb{ 2 \norm{ \Wt{t}_i }_{\infty, \infty} + 6 c_2 \sigma_{wb} \sqrt{ 2 \log \rb{m d} } } \\
 \leq & \; \tau \rb{ 2 c_1 \epsilon_a \sqrt{2 \log m } } \rb{ \norm{ \Wt{t}_i }_{2, 1} + \norm{ \Bt{t}_i }_1 } 2 \norm{ \Wt{t}_i }_{\infty, \infty}  \\
  &+ \tau \rb{ 2 c_1 \epsilon_a \sqrt{2 \log m } } \rb{ \norm{ \Wt{t}_i }_{2, 1} + \norm{ \Bt{t}_i }_1 } 6 c_2 \sigma_{wb} \sqrt{ 2 \log \rb{m d} }  \\
 \leq & \; \tau \rb{ 2 c_1 \epsilon_a \sqrt{2 \log m } } \rb{ \norm{ \Wt{t}_i }_{2, 1} + \norm{ \Bt{t}_i }_1 } 2 \norm{ \Wt{t}_i }_{\infty, \infty}  \\
  &+ \tau \rb{ 2 c_1 \epsilon_a \sqrt{2 \log m } } \rb{ 2 \norm{ \tht{T}_i }_{2, 1} } 6 c_2 \sigma_{wb} \sqrt{ 2 \log \rb{m d} }  \\
   \leq & \; \tau \rb{ 2 c_1 \epsilon_a \sqrt{2 \log m } } \rb{ \norm{ \Wt{t}_i }_{2, 1} + \norm{ \Bt{t}_i }_1 } 2 \norm{ \Wt{t}_i }_{\infty, \infty}  + \omega \rb{ \frac{1}{ m^{c_8} } } \\
   \leq & \; \omega \rb{ \frac{1}{ m^{c_8} } }.
\end{align*}
Using union bound on all $i \in [d]$, with probability atleast $1 - \frac{d}{c_1} - \frac{d}{c_2}$, for all $i \in [d]$, we get
\begin{align*}
\abs{ \dfit{i}{(t)} - \dgit{i}{(t)} } \leq & \; \omega \rb{ \frac{1}{ m^{c_8} } }.
\end{align*}
Taking $c_1 = 100d$ and $c_2 = 100 d$, with atleast 0.98 probability, for all $i \in [d]$, we get
\begin{align*}
 \abs{ \dfit{i}{(t)} - \dgit{i}{(t)} } \leq \omega \rb{ \frac{1}{ m^{c_8} } }.
\end{align*}
\end{proof}

\section{Additional experiments}
\label{sec:additional-experiments}
\todo{outline of the section}
In this section, we show experimental results for both CNF and UNF on different datasets. First, we describe experimental setup in Subsection \ref{subsec:experimental-setup}. Then, we discuss our main observations for constrained normalizing flow and unconstrained normalizing flow in Subsection \ref{subsec:CNF-experiments} and \ref{subsec:UNF-experiments}. In Subsection \ref{subsec:training-curves}, we plot training curves for both CNF and UNF for different learning rates and datasets. Codes for the experiments are available at \url{https://github.com/kulinshah98/overparam-NFs}.

\subsection{Experimental Setup}
\label{subsec:experimental-setup}
\paragraph{Datasets.} We use five synthetic datasets for our experiments. All datasets contain 10,000 data points. The details about the datasets are given below: 
\begin{itemize}
\item \emph{Mixture of Gaussian Dataset:} Data in this dataset lies in 1D and is generated from mixture of 2 Gaussians with means at 2.5 and -2.5. The standard deviation of both Gaussians is 1.
\item \emph{Mixture of Beta Dataset:} Data in this dataset lies in 1D and is generated from mixture of 3 Beta distribution. The parameters of Beta distributions are given by (5, 30), (30, 5) and (30, 30).
\item \emph{Grid Dataset:} Data in this dataset lies in 2D. Figure of the data is given in \ref{fig:grid-dataset}. Brightness at any point in this 2D plot represents the unnormalized probability density of that point.
\begin{figure}
\centering
\includegraphics[scale=0.5]{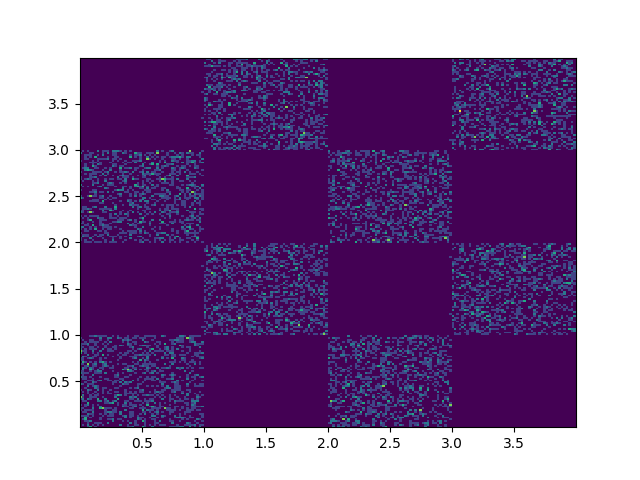}
\caption{Grid dataset}
\label{fig:grid-dataset}
\end{figure}
\item \emph{5D Mixture of Gaussian dataset:} Data in this dataset lies in 5D and is generated from mixture of 10 Gaussians. 
\end{itemize}

\paragraph{Architecture.} We use similar architecture as described in Section~\ref{section:constrained-normalizing-flow-main} and Section~\ref{sec:unconstrained-normalizing-flow} for both constrained and unconstrained normalizing flows. In all our experiments, we fix the weights of the output layer and train the weights and biases of the hidden layer. In UNFs, we use one-hidden layer network for all datasets while in CNFs, we use one-hidden layer network for 1D datasets and use two-hidden layer network for Grid dataset and three-hidden layer network for 5D Mixture of Gaussian dataset. We initialize weights of neural network as described in Section~\ref{section:constrained-normalizing-flow-main} and Section~\ref{sec:unconstrained-normalizing-flow}. We choose $\epsilon_a$ (standard deviation of top layer of neural networks in both UNF and CNF) from $\{0.15, 0.2, 0.25\}$ using the training error after a fixed number of iteration as a metric to evaluate.

\paragraph{Training Procedure.} We use same training procedure for both constrained and unconstrained normalizing flows as described in Section~\ref{section:constrained-normalizing-flow-main} and Section~\ref{sec:unconstrained-normalizing-flow}. We use same base distribution as used in theoretical results for both CNF and UNF (i.e., standard Gaussian for CNFs and standard exponential for UNFs). \todo{Rephrase} Although, we believe that our experimental result can hold for all common distributions as a base distribution. 
In all our experiments, we use mini-batch SGD with batch size 32 for the training. 

All our results are averaged over 5 different iterations. We used NVIDIA Tesla P100 GPU for approx 1000 hours to generate our final experimental results. Our experimental results validate the dichotomy between constrained and unconstrained normalizing flows which was established in Section~\ref{section:constrained-normalizing-flow-main} and Section~\ref{sec:unconstrained-normalizing-flow}. 

\subsection{Results for constrained normalizing flow} 
\label{subsec:CNF-experiments}
\begin{figure}
\begin{center}
 \begin{tabular}{cc}
         \includegraphics[width=0.45\columnwidth]{figures/mog-width-error.png} 
&         
         \includegraphics[width=0.45\columnwidth]{figures/mob-width-error.png}  \\
         \includegraphics[width=0.45\columnwidth]{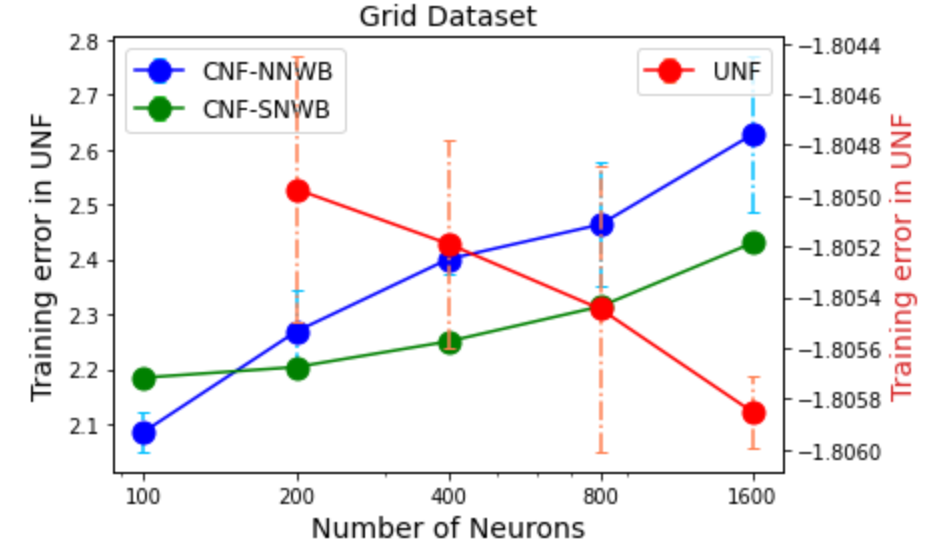} &
		\includegraphics[width=0.45\columnwidth]{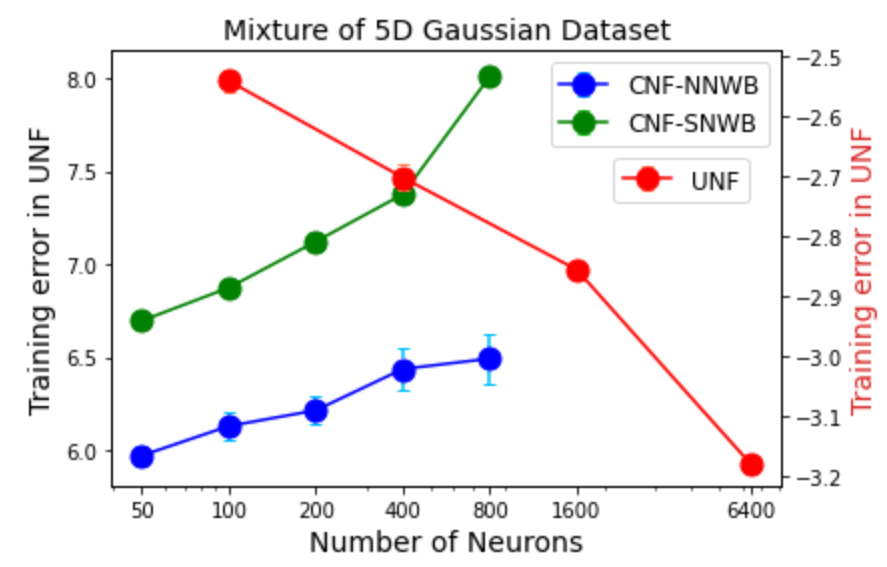} 
    \end{tabular}
\end{center}
		\caption{Comparison between CNF and UNF of training error after a fixed number of training iterations}
		\label{fig:width-vs-error-CNF-UNF}
\end{figure}

\begin{figure}
\begin{center}
\begin{tabular}{cc}
         \includegraphics[width=0.45\columnwidth]{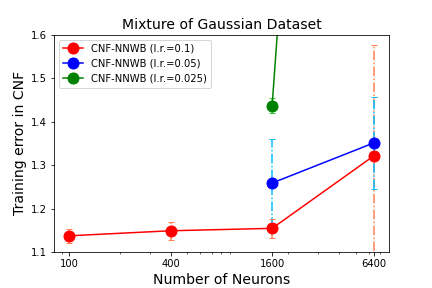} & 
         \includegraphics[width=0.45\columnwidth]{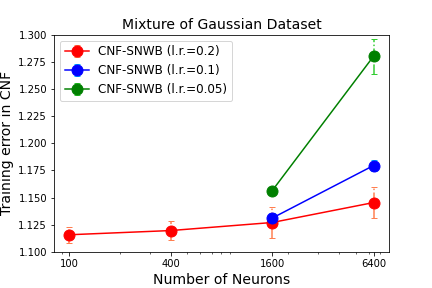} \\
		\includegraphics[width=0.45\columnwidth]{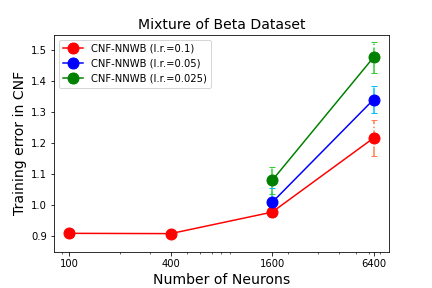} & 
         \includegraphics[width=0.45\columnwidth]{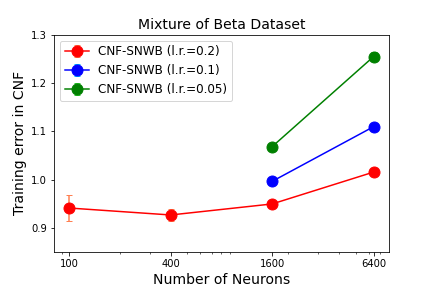} \\
\end{tabular}
\end{center}
\caption{Training error of CNF-NNWB and CNF-SNWB after a fixed number of training iterations for different learning rates}
\label{fig:width-vs-error-diff-learning-rates}
\end{figure}

In Section~\ref{section:constrained-normalizing-flow-main}, we suggested that high overparameterization may adversely affect training for constrained normalizing flows \todo{Make precise?}. In this section, we give empirical evidence for our claims. We use Gaussian distribution as a base distribution in all our experiments of constrained normalizing flow. We experiment with two different initialization for weights and biases of the hidden layer. 1) Gaussian distribution with zero-mean and $\sfrac{1}{m}$ variance ($\sigma_{wb}^2 = \sfrac{1}{m}$) where $m$ is number of neurons in hidden layer. We call CNF with this initialization as CNF-NNWB (CNF with Normalized Normal initialization for Weights and Biases) and 2) Standard Gaussian distribution ($\sigma_{wb}^2 = 1$). We denote CNF with this initialization as CNF-SNWB (CNF with Standard Normal initialization for Weights and Biases). We observe training error and $L_2$ distance of parameters from initialization after a fixed number of iterations for both CNF-NNWB and CNF-SNWB. \todo{write more about learning rates tried?} We made following two observations: 

\paragraph{Effect of overparameterization on training speed of CNF.} In Figure \ref{fig:width-vs-error-CNF-UNF} and Figure \ref{fig:width-vs-error-diff-learning-rates}, we plot width of neural networks versus training error after a fixed number of iterations and for a fixed learning rate. We see that training error for CNF models increases as we increase overparameterization of neural networks, which means that to reach a fixed training error, larger models take \emph{more} number of training updates. This shows that for any fixed learning rate, as we increase overparameterization in CNF, training speed \emph{decreases}. This phenomenon is consistent across different datasets, different learning rates and different initializations. This result is \emph{novel} and \emph{surprising} because in supervised learning, overparameterization helps in \emph{faster} convergence for a fixed learning rate \cite{neyshabur2015search} and we
are not aware of \emph{any other} settings where overparametrization has such strong negative effect.

\begin{figure}
\begin{center}
\begin{tabular}{ccc}
         \includegraphics[width=0.31\columnwidth]{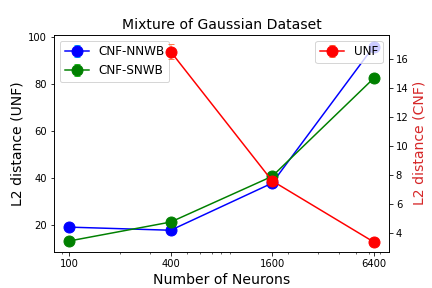} & 
         \includegraphics[width=0.31\columnwidth]{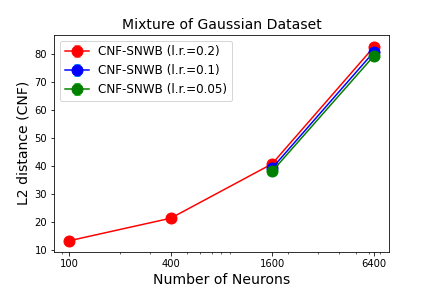} & 
         \includegraphics[width=0.31\columnwidth]{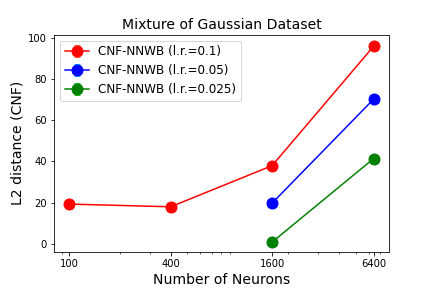} \\
		\includegraphics[width=0.31\columnwidth]{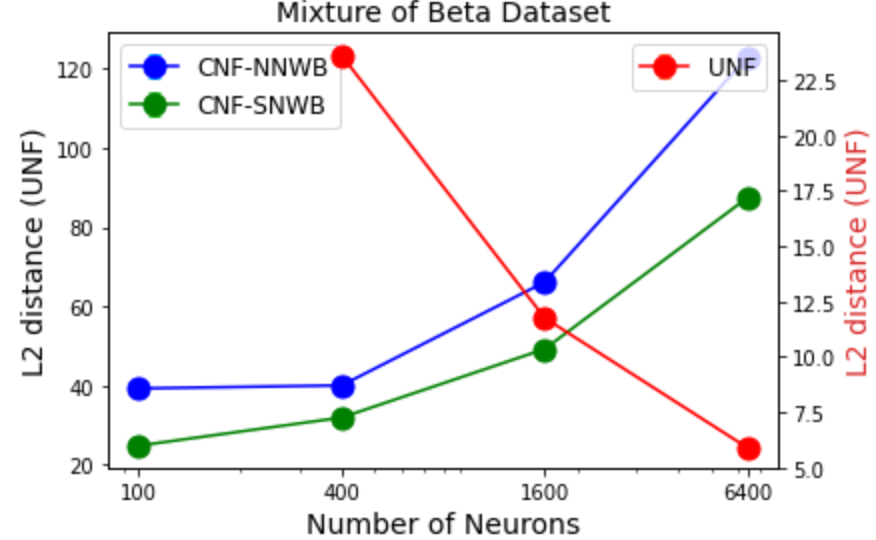} & 
		\includegraphics[width=0.31\columnwidth]{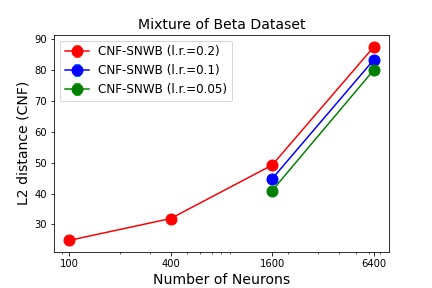} & 
         \includegraphics[width=0.31\columnwidth]{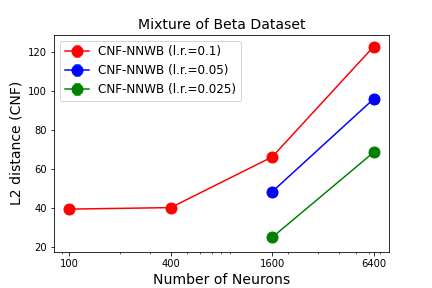}
\end{tabular}
\end{center}
\caption{Comparison of $L_2$ distance from initialization between UNF and CNF models}
\label{fig:width-vs-L2-distance-UNF-CNF}
\end{figure}

\paragraph{Effect of overparameterization on $L_2$ distance of parameters from initialization.} Figure \ref{fig:width-vs-L2-distance-UNF-CNF} has plots of width of neural networks versus $L_2$ distance of parameters from the initialization after a fixed number of training iterations. From the figure, we see that as we increase overparameterization in CNF models, $L_2$ distance from the initialization also increases. From our previous observation, we know that after a fixed number of training iterations, training error increases as overparameterization increases. Combining experiment on $L_2$ distance with our previous observation, we get that to achieve same training error, more overparameterized model have larger $L_2$ distance compared to their smaller counterparts. This result is \emph{surprising} because in supervised learning, it is known that more overparameterized model have smaller distance of parameters from the initialization \cite{nagarajan2019generalization}.

\subsection{Results for unconstrained normalizing flow}
\label{subsec:UNF-experiments}
In Section \ref{sec:unconstrained-normalizing-flow}, we prove that overparameterized neural network can efficiently learn the data distribution. In this section, we will provide empirical evidence that overparameterization helps in training of UNF.
 Similar to CNF, we study training error and $L_2$ distance of parameters from initialization after a fixed number of training iterations. We made following two observations: 
 
\paragraph{Effect of overparameterization on training speed of UNF.} In Figure \ref{fig:width-vs-error-CNF-UNF}, we see that training error after a fixed number of iterations decreases with increasing width of neural networks in UNF, which means that to reach a fixed training error, larger models need \emph{smaller} number of training updates. This implies that for any fixed learning rate, increasing overparameterization in UNF \emph{increases} training speed. This trend is consistent with supervised learning, where it is known that overparameterization helps in \emph{faster} convergence for a fixed learning rate \cite{neyshabur2015search}. 
 
\paragraph{Effect of overparameterization on $L_2$ distance of parameters from initialization.} Figure \ref{fig:width-vs-L2-distance-UNF-CNF} shows that as we increase overparameterization in UNF models, $L_2$ distance of parameters from the initialization decreases. Our previous observation was that after a fixed number of training iterations, training error decreases or remains almost same as overparameterization increases. Combining our observation on $L_2$ distance with our previous observation, we get that to achieve a fixed training error, more overparameterized model require smaller $L_2$ distance compared to their less overparameterized counterparts. This result is \emph{consistent} with supervised learning, where it is known that more overparameterized model have smaller distance of parameters from the initialization \cite{nagarajan2019generalization}.

\subsection{Results on Miniboone dataset}

\begin{figure}
\begin{center} 
\includegraphics[width=0.45\columnwidth]{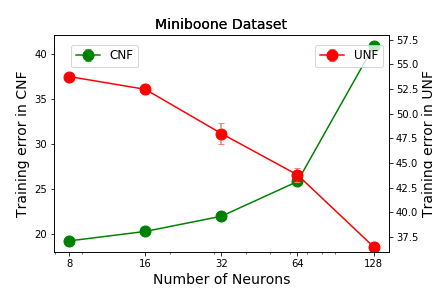} 
\end{center}
\caption{Training error of CNF and UNF after a fixed number of training epochs. }
\label{fig:width-vs-error-miniboone}
\end{figure}

To show experimental results on a real-world dataset, we use miniboone dataset \citep{uci}. The dataset contains examples of electron neutrino and muon neutrino. This dataset contains around 30K examples and lies in 43 dimensions. To test our phenomenon, we modify the official implementation of block neural autoregressive flow (BNAF) \citep{BlockAutoRegressive} for CNF and Unconstrained Monotonic Neural Network Flow \citep{FrenchPaper} for UNF. We use 3 hidden layers for CNF and 3 hidden layers for both embedding network and derivative network. We use one flow model for both of them and use a mini-batch SGD optimizer with a learning rate of 0.001. The figure to illustrate the change in training error by changing the width of the network for each dimension is plotted in \ref{fig:width-vs-error-miniboone}. From the figure, we see that the training error for CNFs increases with an increase in width of the network whereas the training error for UNFs decreases with an increase in width of the network. This observation supports our theoretical results. 

\begin{figure}
\begin{center}
    \begin{tabular}{ccc}
    \includegraphics[width=0.3\columnwidth]{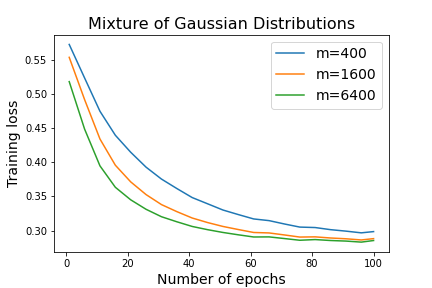} &
        \includegraphics[width=0.3\columnwidth]{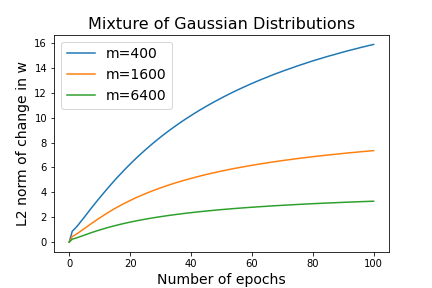}  & \includegraphics[width=0.3\columnwidth]{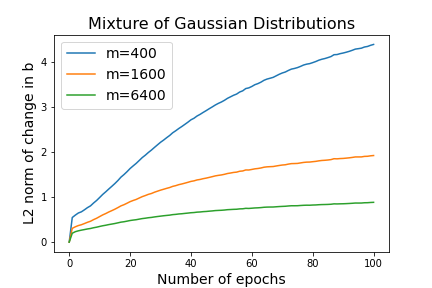} \\
        \includegraphics[width=0.3\columnwidth]{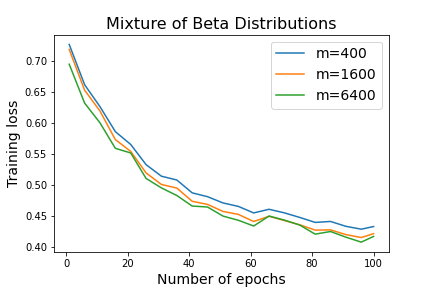} &
        \includegraphics[width=0.3\columnwidth]{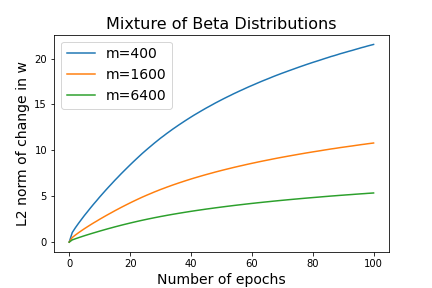}  & 
        \includegraphics[width=0.3\columnwidth]{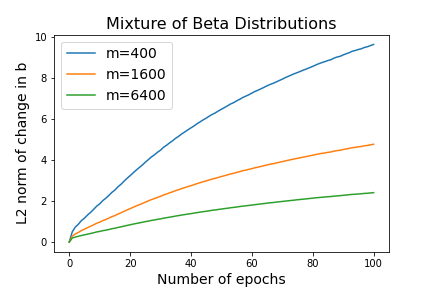} 
    \end{tabular}
\end{center}
\caption{Effect of over-parameterization on training of unconstrained normalizing flow on mixture of Gaussian and mixture of beta distributions}
\label{fig:unconstrained-flow-results}
\end{figure}

\subsection{Training curves for Constrained and Unconstrained Normalizing Flow}
\label{subsec:training-curves}
To provide a complete picture, we provide training error and $L_2$ distance of weights $\Wt{t}$ and biases $\Bt{t}$ from the initialization for all time step $t$ during the training. We first discuss results for CNFs and then move our discussion to UNFs.

\begin{figure}
\begin{center}
    \begin{tabular}{ccc}
         \includegraphics[width=0.31\columnwidth]{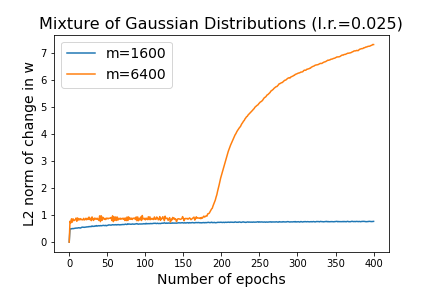} & \includegraphics[width=0.31\columnwidth]{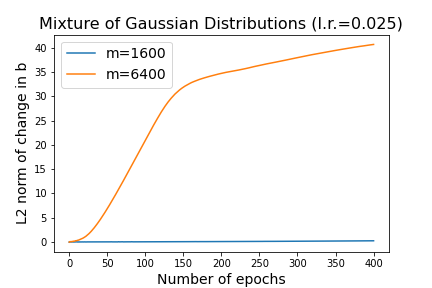} &
         \includegraphics[width=0.31\columnwidth]{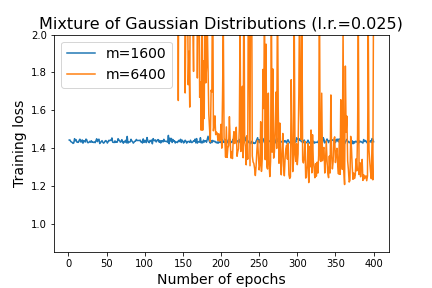} \\
         \includegraphics[width=0.31\columnwidth]{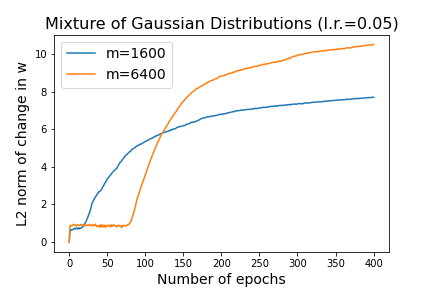} & \includegraphics[width=0.31\columnwidth]{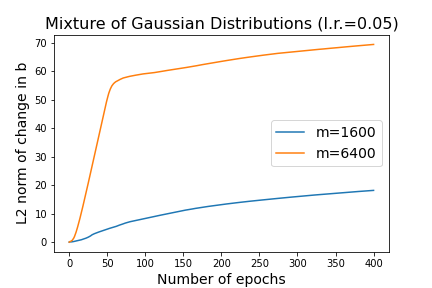} &
         \includegraphics[width=0.31\columnwidth]{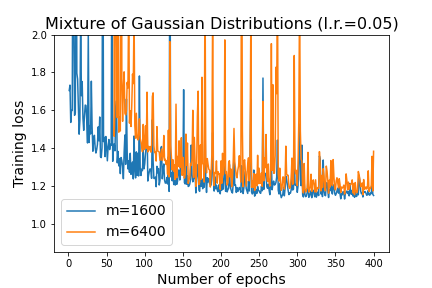} \\
         \includegraphics[width=0.31\columnwidth]{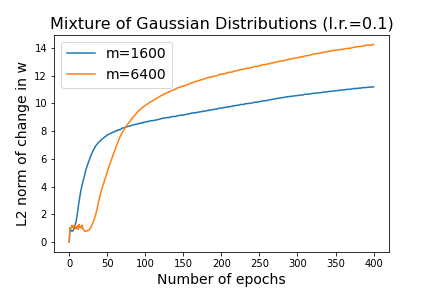} & \includegraphics[width=0.31\columnwidth]{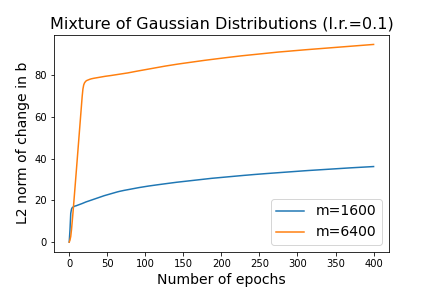} &
         \includegraphics[width=0.31\columnwidth]{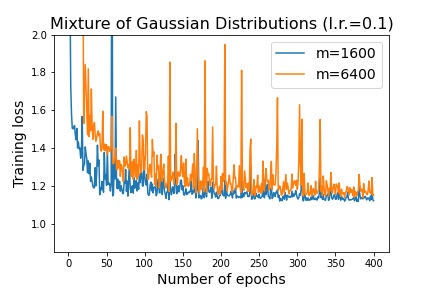}
    \end{tabular}
\end{center}
\caption{Effect of over-parameterization on training of CNF-NNWB on mixture of Gaussian dataset for number of hidden nodes $m=1600, 6400$ }
\label{fig:constrained-flow-mog-high-m-results-appendix}
\end{figure}

\begin{figure}
\begin{center}
    \begin{tabular}{ccc}
         \includegraphics[width=0.31\columnwidth]{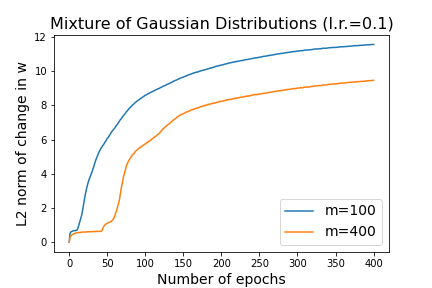} & 
         \includegraphics[width=0.31\columnwidth]{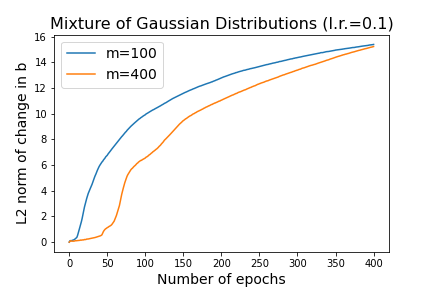} & 
         \includegraphics[width=0.31\columnwidth]{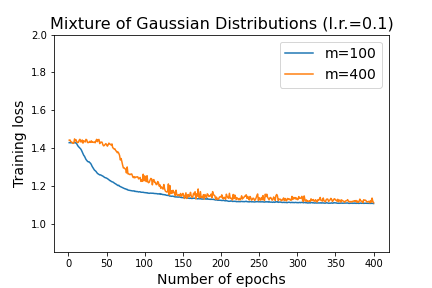} \\
         \includegraphics[width=0.31\columnwidth]{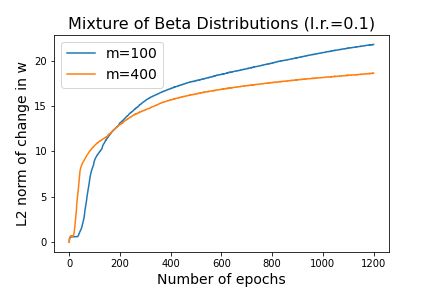} & 
         \includegraphics[width=0.31\columnwidth]{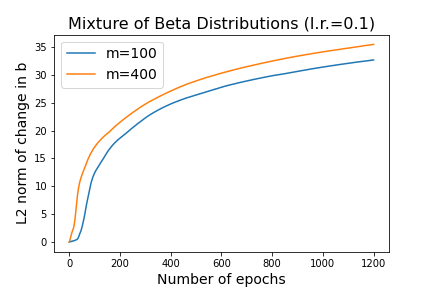} & 
         \includegraphics[width=0.31\columnwidth]{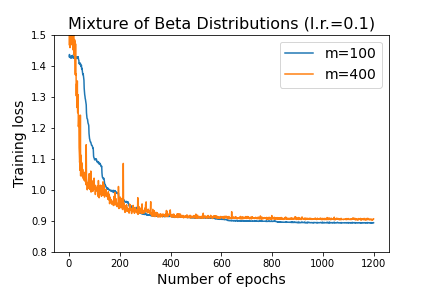}
    \end{tabular}
\end{center}
\caption{Effect of over-parameterization on training of small sized CNF-NNWB}
\label{fig:constrained-flow-small-m-results}
\end{figure}

\begin{figure}
\begin{center}
    \begin{tabular}{ccc}
         \includegraphics[width=0.31\columnwidth]{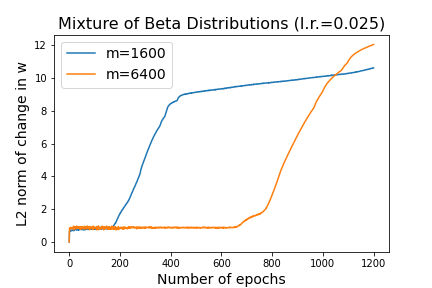} & \includegraphics[width=0.31\columnwidth]{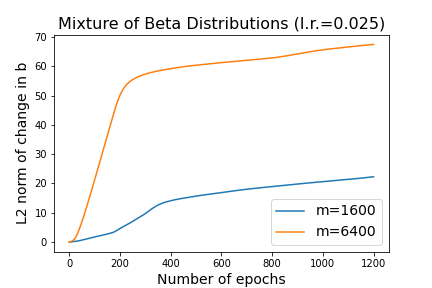} &
         \includegraphics[width=0.31\columnwidth]{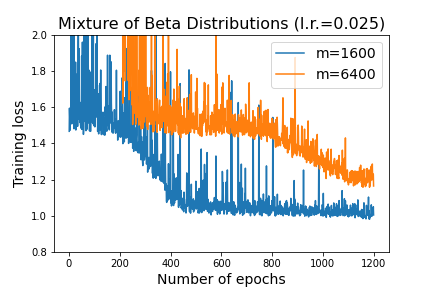} \\
         \includegraphics[width=0.31\columnwidth]{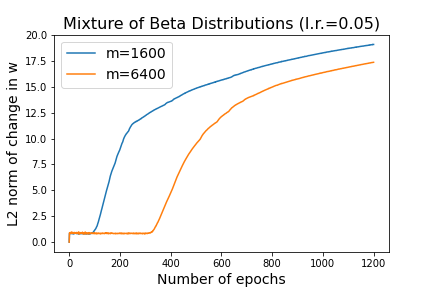} & \includegraphics[width=0.31\columnwidth]{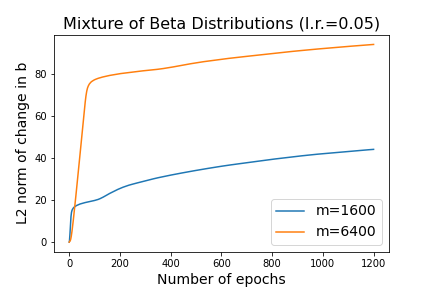} &
         \includegraphics[width=0.31\columnwidth]{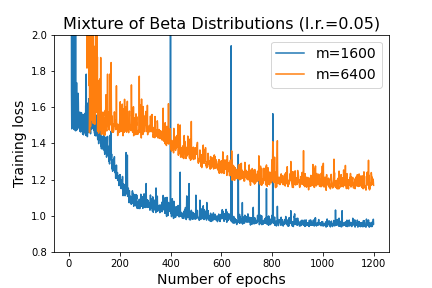} \\
         \includegraphics[width=0.31\columnwidth]{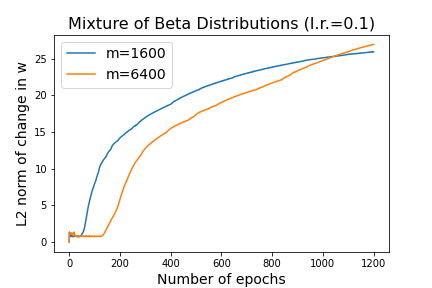} & \includegraphics[width=0.31\columnwidth]{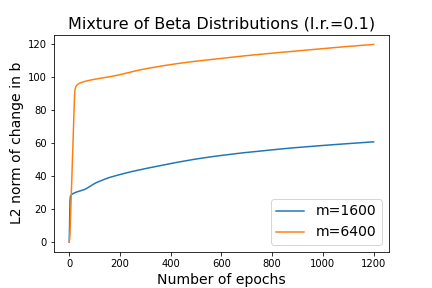} &
         \includegraphics[width=0.31\columnwidth]{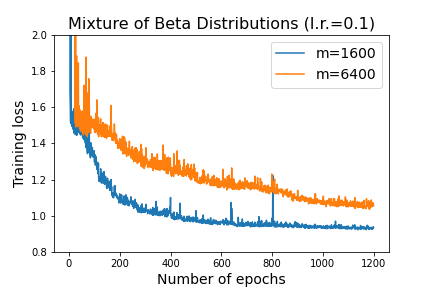}
    \end{tabular}
\end{center}
\caption{Effect of over-parameterization on training of CNF-NNWB on mixture of beta distribution dataset for number of hidden nodes $m=1600, 6400$}
\label{fig:constrained-flow-beta-high-m-results}
\end{figure}

\begin{figure}
\begin{center}
    \begin{tabular}{ccc}
         \includegraphics[width=0.31\columnwidth]{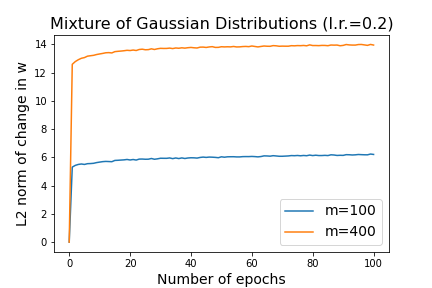} & 
         \includegraphics[width=0.31\columnwidth]{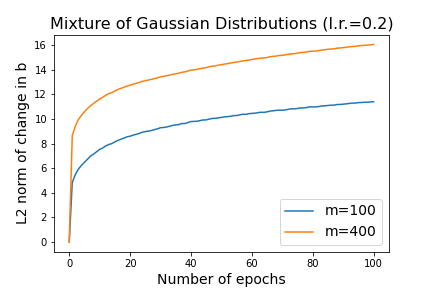} & 
         \includegraphics[width=0.31\columnwidth]{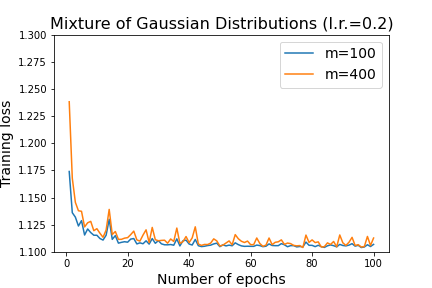} \\
         \includegraphics[width=0.31\columnwidth]{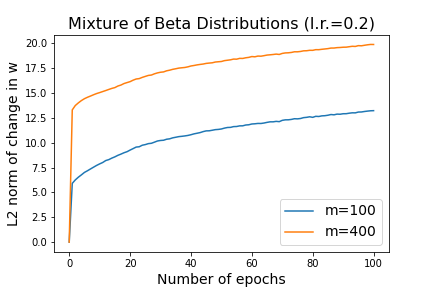} & 
         \includegraphics[width=0.31\columnwidth]{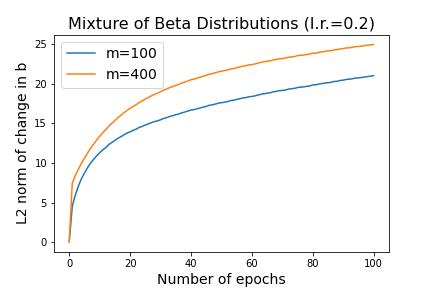} & 
         \includegraphics[width=0.31\columnwidth]{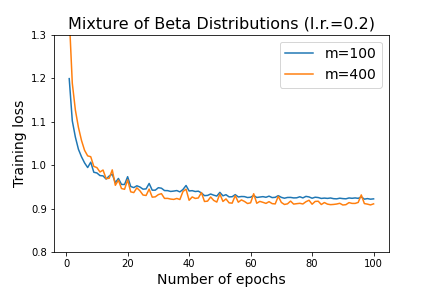}
    \end{tabular}
\end{center}
\caption{Effect of over-parameterization on training of small sized CNF-SNWB of weights and biases }  
\label{fig:constrained-flow-small-m-std-init-results}
\end{figure}

\begin{figure}
\begin{center}
    \begin{tabular}{ccc}
         \includegraphics[width=0.31\columnwidth]{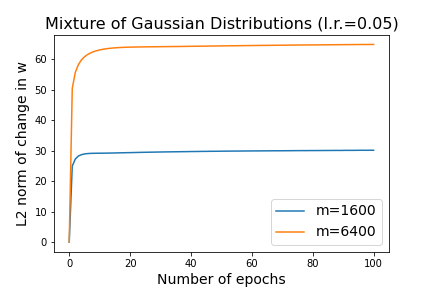} & \includegraphics[width=0.31\columnwidth]{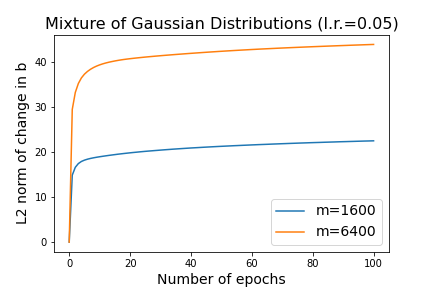} &
         \includegraphics[width=0.31\columnwidth]{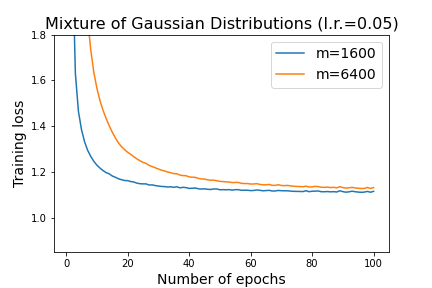} \\
        \includegraphics[width=0.31\columnwidth]{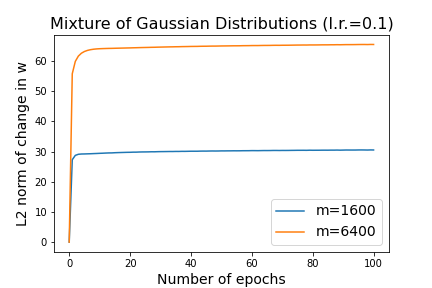} & \includegraphics[width=0.31\columnwidth]{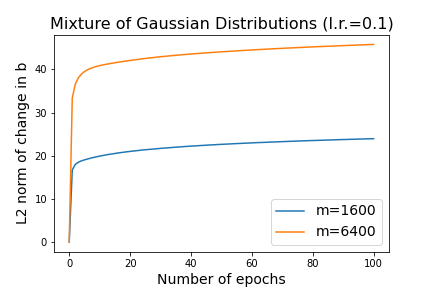} &
         \includegraphics[width=0.31\columnwidth]{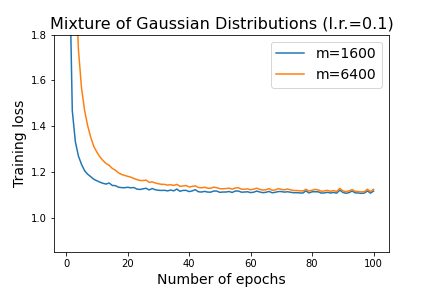} \\
             \includegraphics[width=0.31\columnwidth]{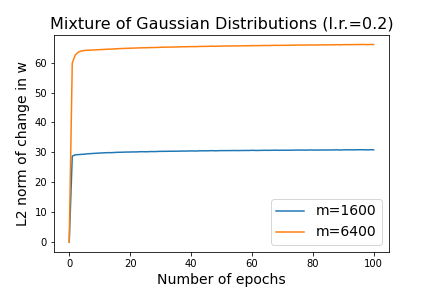} & \includegraphics[width=0.31\columnwidth]{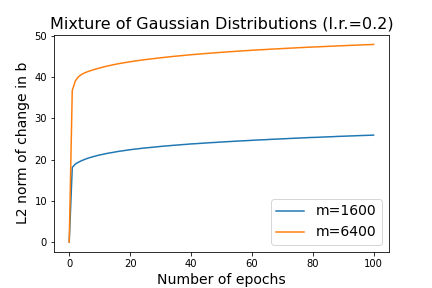} &
         \includegraphics[width=0.31\columnwidth]{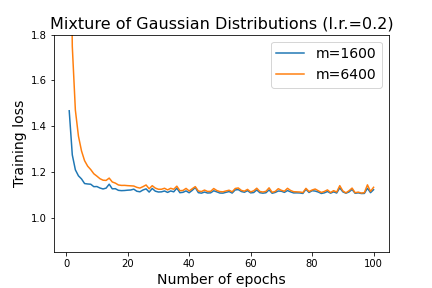}
    \end{tabular} 
\end{center}
\caption{Effect of over-parameterization on training of CNF-SNWB on mixture of Gaussian dataset for number of hidden nodes $m=1600, 6400$ }
\label{fig:constrained-flow-mog-high-m-high-variance-results-appendix}
\end{figure}

\begin{figure}
\begin{center}
    \begin{tabular}{ccc}
         \includegraphics[width=0.31\columnwidth]{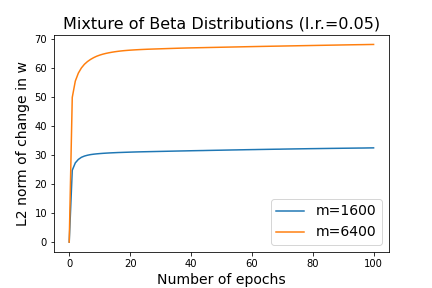} & \includegraphics[width=0.31\columnwidth]{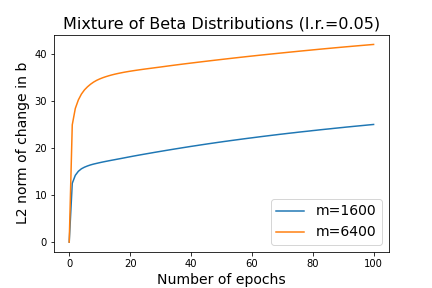} &
         \includegraphics[width=0.31\columnwidth]{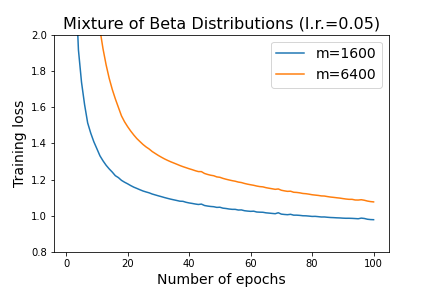} \\
        \includegraphics[width=0.31\columnwidth]{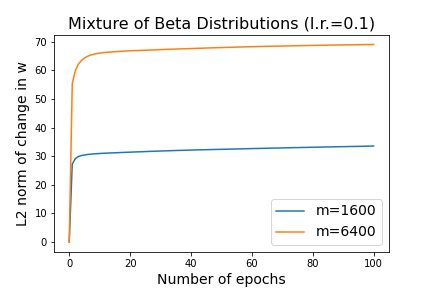} & \includegraphics[width=0.31\columnwidth]{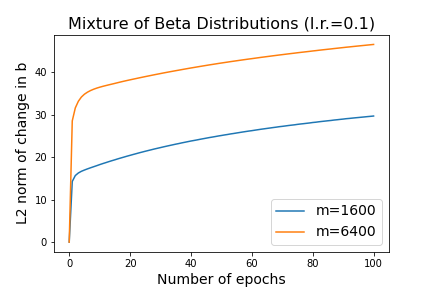} &
         \includegraphics[width=0.31\columnwidth]{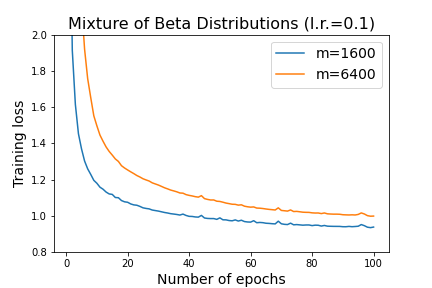} \\
             \includegraphics[width=0.31\columnwidth]{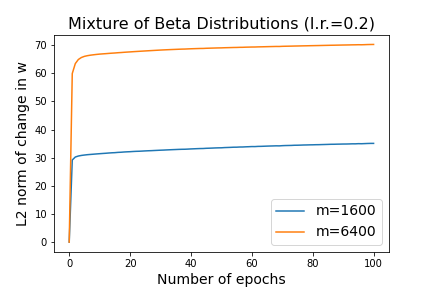} & \includegraphics[width=0.31\columnwidth]{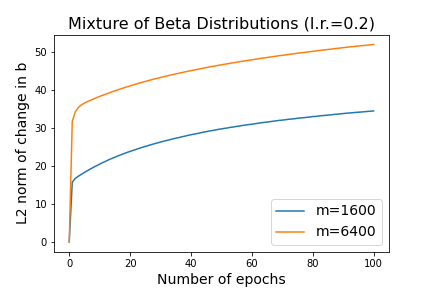} &
         \includegraphics[width=0.31\columnwidth]{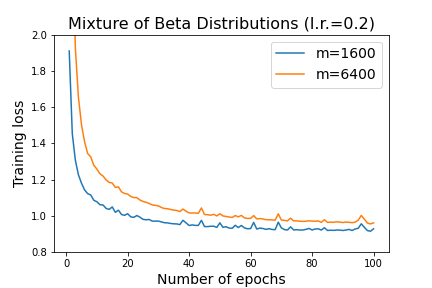}
    \end{tabular} 
\end{center}
\caption{Effect of over-parameterization on training of CNF-SNWB on mixture of Beta distribution dataset for number of hidden nodes $m=1600, 6400$ }
\label{fig:constrained-flow-beta-high-m-high-variance-results-appendix}
\end{figure}

\paragraph{Constrained Normalizing Flow.} In Figure \ref{fig:constrained-flow-mog-high-m-results-appendix}, \ref{fig:constrained-flow-small-m-results}, \ref{fig:constrained-flow-small-m-std-init-results},  \ref{fig:constrained-flow-beta-high-m-results}, \ref{fig:constrained-flow-mog-high-m-high-variance-results-appendix} and \ref{fig:constrained-flow-beta-high-m-high-variance-results-appendix}, we plot number of epochs on x-axis and y-axis can be training error, $L_2$ distance of weights or $L_2$ distance of biases from the initialization. 

In all figures, we see that for any fixed learning rate, curve of training error for smaller $m$ is always below than curve of training for larger $m$, which proves our claim that increasing overparameterization hurts the training speed of CNF models. This phenomenon is consistent for all datasets, different initializations and various learning rates. Only exception to this phenomenon is results on mixture of Gaussian dataset for $m=1600$ and $m=6400$ and learning rate equal to 0.025 but note that in this case, the training of CNF for $m=6400$ is very unstable and therefore, at some time steps, $m=6400$ curve has slightly smaller training error than $m=1600$ because of unstable training. 

Apart from training error, we see that $L_2$ distance for biases (that is, $L_2$ norm of $\Bt{t}$) is always larger for large $m$. The difference is clearly visible and significant in comparison figures of large hidden layer nodes ($m=1600$ and $m=6400$). This is consistent across different initializations, datasets and learning rates. Only exception to this trend is results on mixture of Gaussian dataset for $m=100$ and $m=400$. Even in this case, $L_2$ distance is comparable for $m=100$ and $m=400$.


\paragraph{Unconstrained Normalizing Flow.}

Similar to Constrained Normalizing Flows, we study the effect of overparameterization on convergence speed and $L_2$-norm of $\Wt{t}$ and $\Bt{t}$. The first row of Figure \ref{fig:unconstrained-flow-results} contains results for mixture of Gaussians dataset and the second row contains results for mixtures of beta distributions dataset. From the first column of Fig. \ref{fig:unconstrained-flow-results}, we see that the training speed for larger $m$ is better or comparable to smaller $m$. Additionally, we see that $L_2$-norm of $\Wt{t}$ and $\Bt{t}$ decreases significantly with increasing $m$. This results validate our theoretical finding that $L_2$ distance of parameters from the initialization decreases with increasing $m$.


\begin{figure}
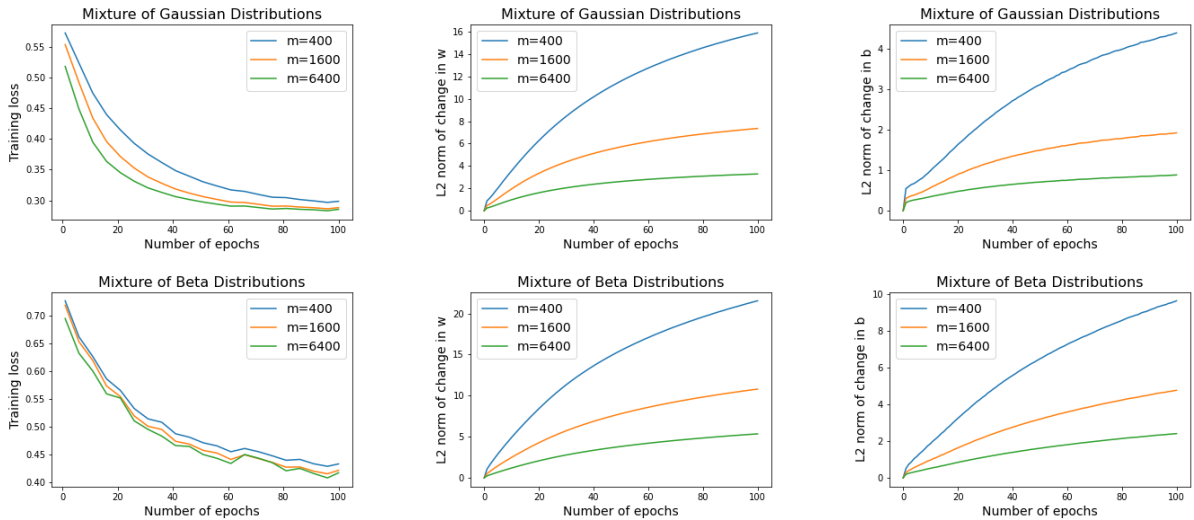

\begin{center}
    \begin{tabular}{ccc}
    \includegraphics[width=0.3\columnwidth]{figures/umnn/mog-training-loss.png} &
        \includegraphics[width=0.3\columnwidth]{figures/umnn/mog-change-in-w.png}  & \includegraphics[width=0.3\columnwidth]{figures/umnn/mog-change-in-b.png} \\
        \includegraphics[width=0.3\columnwidth]{figures/umnn/beta-training-loss.png} &
        \includegraphics[width=0.3\columnwidth]{figures/umnn/beta-change-in-w.png}  & 
        \includegraphics[width=0.3\columnwidth]{figures/umnn/beta-change-in-b.png} 
    \end{tabular}
\end{center}
\caption{Effect of over-parameterization on training of UNF on mixture of Gaussian and mixture of beta distributions}
\label{fig:unconstrained-flow-results}
\end{figure}

\section{Related Work}
\label{sec:related-work}
Previous work on normalizing flows has studied different variants such as planar and radial flows in \cite{RezendeMohamed2015}, Sylvester flow in \cite{vdberg2018sylvester}, Householder flow in \cite{tomczak2016improving}, masked autoregressive flow in \cite{papamakarios2017masked}. Most variants of normalizing flows are specific to certain applications, and the expressive power (i.e., which base and data distributions they can map between) and complexity of normalizing flow models have been studied recently, e.g. \cite{kong2020expressive} and \cite{teshima2020coupling}. Invertible transformations defined by monotonic neural networks can be combined into autoregressive flows that are universal density approximators of continuous probability distributions; see Masked Autoregressive Flows (MAF) \cite{papamakarios2017masked}, UNMM-MAF by \cite{FrenchPaper}, Neural Autoregressive Flows (NAF) by \cite{huang2018neural}, Block Neural Autoregressive Flow (B-NAF) by \cite{cao2019block}. Unconstrained Monotonic Neural Network (UMNN) models proposed by \cite{FrenchPaper} are particularly relevant to the technical part of our paper.

\cite{risteski2020representationalNF} theoretically study representation ability of affine couplings (a type of normalizing flow) and particularly analyze several aspects such as depth of normalizing flows. \cite{lei2020sgd, feizi2021understandingGAN} show that when the generator is a two-layer tanh, sigmoid or leaky ReLU network, Wasserstein GAN trained with stochastic gradient descent-ascent converges to a global solution with polynomial time and sample complexity. Using the moments method and a learning algorithm motivated by tensor decomposition, \cite{li2020making} show that GANs can efficiently learn a large class of distributions including those generated by two-layer networks. 
\cite{nguyen2019dynamics} show that two-layer autoencoders with ReLU or threshold activations can be trained with normalized gradient descent over the reconstruction loss to provably learn the parameters of any generative bilinear model (e.g., mixture of Gaussians, sparse coding model). \cite{nguyen2019benefits} extend the work of \cite{du2018gradient} on supervised learning mentioned earlier to study weakly-trained (i.e., only encoder is trained) and jointly-trained (i.e., both encoder and decoder are trained) two-layer autoencoders, and show joint training requires less overparameterization and converges to a global optimum. The effect of overparameterization in unsupervised learning has also been of recent interest. \cite{buhai2020benefits} do an empirical study to show that across a variety of latent variable models and training algorithms, overparameterization can significantly increase the number of recovered ground truth latent variables. \cite{radhakrishnan2020overparameterized} show that overparameterized autoencoders and sequence encoders essentially implement associative memory by storing training samples as attractors in a dynamical system.

\section{Useful facts}

\begin{fact} \label{fact:sum-prod-hermite} For any $i \geq 0$, let $h_i$ denote the degree$-i$ probabilists' Hermite polynomial
\begin{align*}
h_i(x) = i! \sum_{m=0}^{ \lfloor \frac{i}{2} \rfloor } \frac{ (-1)^m }{  m! (i - 2m)!  } \frac{x^{i - 2m}}{2^m}.
\end{align*}
The Hermite polynomials satisfy following summation and multiplication formulas.
\begin{align*}
h_i(x + y) &= \sum_{k=0}^i \binom{i}{k} x^{i-k} h_k(y), \\
h_i(xy) &= \sum_{k=0}^{ \lfloor \frac{i}{2}  \rfloor } y^{i-2k} ( y^2 - 1 )^k \binom{i}{2k} \frac{ (2k)! }{ k! } 2^{-k} h_{i-2k}(x).
\end{align*}
\end{fact}

\begin{fact} \label{fact:hermite-normal-expectation} Let $h_i$ denote the degree$-i$ probabilists' Hermite polynomial, then for $i>0$, we have
\begin{align*}
\E_{ \beta \sim \N\rb{0, 1} } \sqb{ h_i \rb{ \beta } } = 0.
\end{align*}
\end{fact}

\begin{lemma} \label{lemma:concentration-chi-square} Suppose $Z_k \sim \N(0, \sigma^2) $ and $Y = \sum_{k=1}^n Z_k^2 $ is chi-squared distribution with following property for all $t \in \rb{0, 1}$.
\begin{align*}
    \text{Pr} \left[ \abs{ \frac{1}{n} \sum_{k=1}^n Z_k^2 - \sigma^2 } \geq t \right] \leq 2 \exp \rb{ - \frac{n t^2}{8 \sigma^4 } }
\end{align*}
\end{lemma}
\begin{proof} From example 2.11 from \cite{wainwright_2019}, for $Z_k' \sim \N(0, 1) $ and $Y = \sum_{k=1}^n Z_k'^2 $ is chi-squared distribution with following property for all $t \in \rb{0, 1}$.
\begin{align*}
    \text{Pr} \left[ \abs{ \frac{1}{n} \sum_{k=1}^n Z_k'^2 - 1 } \geq t \right] \leq 2 \exp \rb{ - \frac{n t^2}{8 } }
\end{align*}
Using above equation for $\frac{Z_k}{ \sigma }$, 
\begin{align*}
    \text{Pr} \left[ \abs{ \frac{1}{n} \sum_{k=1}^n \frac{ Z_k^2 }{ \sigma^2 } - 1 } \geq \frac{t}{\sigma^2} \right] &\leq 2 \exp \rb{ - \frac{n t^2}{8 \sigma^4} } \\
    \text{Pr} \left[ \abs{ \frac{1}{n} \sum_{k=1}^n  Z_k^2  - \sigma^2 } \geq t \right] &\leq 2 \exp \rb{ - \frac{n t^2}{8 \sigma^4} } 
\end{align*}
\end{proof}

\begin{lemma}
\label{lemma:concentration-folded-normal}
    Let $X_1, X_2, ..., X_n$ be independent random variables from $\mathcal{N}(0, \sigma^2)$, then with at least $1 - \frac{1}{c_1}$ probability, following holds.
    \begin{align*}
        \max_{ i \in \{1,2,...,n\} } | X_i | \leq 2 c_1 \sigma \sqrt{2 \log n}
    \end{align*}
\end{lemma}
\begin{proof}
From \cite{maxgauss2012},
\begin{align*}
    \E \left[ \max_{i \in \{1,2,...,n\} } | X_i | \right] \leq \sigma \left( \sqrt{2 \log n} + 1 \right) \leq 2 \sigma \left( \sqrt{2 \log n} \right)
\end{align*}
Assuming $n \geq 2$, the last inequality follows.
Using Markov's inequality, 
\begin{align*}
     &\text{Pr} \left( \max_{i \in \{1, 2, ..., n\} } | X_i | \geq 2 c_1 \sigma \left( \sqrt{2 \log n} \right) \right) \leq \frac{1}{c_1} \\
    &\text{Pr} \left( \max_{i \in \{1, 2, ..., n\}  } | X_i | \leq 2 c_1 \sigma \left( \sqrt{2 \log n} \right) \right) \geq 1 - \frac{1}{c_1}
\end{align*}
\begin{align*}
    s
\end{align*}
\end{proof}

\begin{lemma}
\label{lemma:anti-concentration-zero-mean-normal}
For standard Gaussian random variable $X$ from $\N(0, \sigma^2)$, the following anti-concentration inequality holds:
\begin{align*}
    \Pr \rb{ \abs{X} \leq R } \leq \frac{2 R}{ \sigma \sqrt{2 \pi} }.
\end{align*}
\end{lemma}
\begin{proof}
(From \cite{du2018gradient}) For the standard Gaussian random variable $\frac{X}{\sigma}$, 
\begin{align*}
    \Pr \rb{ \abs{ \frac{X}{ \sigma } } \leq R } \leq \frac{2 R}{ \sqrt{2 \pi} }
\end{align*}
Using $R=\frac{R'}{ \sigma }$, we get the required result. 
\end{proof}

\begin{lemma}
\label{lemma:relation-std-lipschitz-coordinate-lipschitz}
Suppose function $f : \R^d \rightarrow \R$ is $L_g$-Lipschitz continuous and $L_i$-coordinate wise Lipschitz continuous i.e.
\begin{align*}
    \left| f( \mathbf{a} ) - f( \mathbf{b} ) \right| \leq & L_g \| \mathbf{a} - \mathbf{b} \| \\
    &\forall \mathbf{a}, \mathbf{b} \in \R^d \quad \text{(Standard Lipschitz continuity)} \\
    \left| f(a_1, a_2, ..., a_i, ..., a_d) - f( a_1, a_2, ..., b_i, ..., a_d ) \right| \leq & L_i | a_i - b_i | \\ 
    \forall a_1, a_2, ..., a_i, ..., a_d, b_i \in \R & \text{ and } \forall i \in [d] \quad \text{(Coordinate-wise Lipschitz continuity)} \\
\end{align*}
If a function $f$ satisfies $L_i$-coordinate wise Lipschitz continuity for all $i$, then function $f$ follows following inequality. 
\begin{align*}
    \abs{ f( a_1, a_2, ..., a_d ) - f( b_1, b_2, ..., b_d ) } \leq \sum_{i=1}^n L_i \abs{ a_i - b_i }
\end{align*}
Moreover, the function $f$ also satisfies standard Lipschitz continuity with $L_g$ Lipschitz constant where inequality between $L_g$ and $L_i$ is as follows.
\begin{align*}
    L_g \leq \sqrt{ \sum_{i=1}^d L_i^2 }
\end{align*}

\end{lemma}
\begin{proof}
Define $\mathbf{a} = \rb{ a_1, a_2, ..., a_d }$ and $\mathbf{b} = \rb{ b_1, b_2, ..., b_d }$. 
\begin{align*}
    \abs{ f( a_1, a_2, ..., a_d ) - f( b_1, b_2, ..., b_d ) } \leq & \abs{ f( a_1, a_2, ..., a_d ) - f( b_1, a_2, ..., a_d ) } \\ 
    &+ \abs{ f( b_1, a_2, a_3, ..., a_d ) - f( b_1, b_2, a_3, ..., a_d ) } \\
    &+ \abs{ f( b_1, b_2, a_3, ..., a_d ) - f( b_1, b_2, b_3, ..., a_d ) } \\
    &+ ... + \abs{ f( b_1, b_2, ..., b_{d-1}, a_d ) - f( b_1, b_2, b_3, ..., b_d ) } \\
    \leq & L_1 \abs{ a_1 - b_1 } + L_2 \abs{ a_2 - b_2 } + ... + L_d \abs{ a_d - b_d } \\
    \leq & \sqrt{ \sum_{i=1}^d L_i^2 } \| \mathbf{a} - \mathbf{b} \|_2
\end{align*}
where last inequality follows from Cauchy-Schwarz inequality. 
\end{proof}

\begin{fact}
\label{fact-hoeffding-on-binomial}
(Hoeffding's inequality on Binomial random variable) If we have a binomial random variable with parameters $n$ (total number of trials) and $p$ (probability of success). For $k \geq np$, following inequality holds.
\begin{align*}
    \text{Pr} \rb{X \geq k} \leq \exp \rb{-2 n \rb{ \frac{k}{n} - p }^2 }
\end{align*}

\end{fact}

\begin{fact} \label{fact:hoeffding-inequality} (Hoeffding's inequality) Let $X_1, X_2, \ldots, X_n$ be independent random variables where $X_i$ is bounded in the interval $\sqb{a_i, b_i}$. Then, for any $t \geq 0$, we have
\begin{align*}
\Pr \rb{ \abs{ \rb{ X_1 + X_2 + \ldots + X_n } - \E \sqb{ X_1 + X_2 + \ldots + X_n } } \geq t } \leq 2 \exp \rb{ - \frac{2 t^2}{ \sum_{i=1}^n \rb{ a_i - b_i }^2 } }.
\end{align*}
\end{fact}

\begin{fact}
\label{fact:mean-half-normal-distribution} (Half-normal distribution) If $X$ follows a normal distribution with with mean 0 and variance $\sigma^2$, $\N \rb{ 0, \sigma^2 }$, then $Y = \abs{X} = X \text{sign} \rb{ X }$ follows a half-normal distribution with mean $\E \left[ Y \right] = \frac{ \sigma \sqrt{2} }{ \sqrt{ \pi } }$.

\end{fact}
\todo{proof?}

\begin{fact}
\label{fact:anticoncentration-gaussian}
For a gaussian random variable $X \sim \mathcal{N}(0, \sigma^2)$, $\forall t \in (0, \sigma)$, we have
\begin{align*}
    \text{Pr}( |X| \geq t ) \geq 1 - \frac{4 t}{ 5 \sigma }
\end{align*}
\end{fact}

\begin{fact}
\label{fact:sum-inifinite-inverse-squares} The sum of reciprocals of the squares of the natural numbers is given by
\begin{align*}
    \sum_{n=1}^{\infty} \frac{1}{n^2} = \frac{\pi^2}{6} \leq 2
\end{align*}
\end{fact}

\begin{fact} 
\label{fact:bernouli-inequality} 
(Theorem $3.1(r_5^{'})$ of \cite{li2013some}) For any $\alpha > 1$ and $x \in \left[ 0, \frac{1}{ \alpha - 1 } \right)$, 
\begin{align*}
    \left( 1 + x \right)^{\alpha} \leq \frac{1}{ 1 - \frac{\alpha x}{ 1 + x } } = 1 + \frac{ \alpha x }{ 1 - \left( \alpha - 1 \right) x }
\end{align*}
\end{fact}

\begin{fact} (McDiarmid’s Inequality)
\label{fact:mcdiarmid-inequality} 
Let $V$ be some set and let $f: V^m \mapsto \R$ be a function such that for some $c_i > 0$, for all $i \in [m]$ and for all $x_1, \ldots, x_m, x_i' \in V$, we have
\begin{align*}
\abs{ f \rb{ x_1, \ldots, x_i, \ldots, x_m }  - f \rb{ x_1, \ldots, x_i', \ldots, x_m }  } \leq c_i.
\end{align*}
Let $X_1, X_2, \ldots X_m$ are independent random variables taking values in $V$. Then, 
\begin{align*}
\Pr \sqb{ f( X_1, X_2, \ldots, X_m ) - \E \rb{ X_1, X_2, \ldots, X_m } \geq \epsilon  } \leq \exp \rb{ \frac{-2 \epsilon^2}{ \sum_{i=1}^m c_i^2 } }.
\end{align*}

\end{fact}

\begin{fact}
\label{fact:sum-of-agp}
If Arithmetic-Geometric Progression(AGP) is as follows.
\begin{align*}
    a, (a + d)r, (a + 2d)r^2, (a + 3d)r^3, ...., \left[ a + (n-1)d \right] r^{n-1}
\end{align*}
where $a$ is the initial term, $d$ is the common difference and $r$ is the common ratio. The sum of the first $n$ terms of the AGP ($S_n$) is given by
\begin{align*}
    S_n = \frac{ a - \left[ a + (n-1)d \right] r^n }{1 - r} + \frac{dr \left( 1 - r^{n-1} \right) }{ (1 - r)^2 }
\end{align*}
\end{fact}

%

\begin{definition} Let $\mcf$ be a set of functions $\R^d \rightarrow \R$ and $\mcx = \rb{x_1, x_2, ..., x_n}$ be a finite set of samples. The empirical Rademacher complexity of $\mcf$ with respect to $\mcx$ is defined by 
\begin{align*}
    \empR \rb{ \mcx; \mcf } = \E_{ \xi \sim \{ \pm 1 \}^n } \left[ \sup_{f \in \mcf} \frac{1}{n} \sum_{i=1}^n \xi_i f(x_i) \right].
\end{align*}

\end{definition}

The following results are standard and can be found, e.g., in  \cite{allen2019learning}.
\begin{lemma} \label{lemma:rademacher-complexity-properties} Rademacher complexity has the following properties:
\begin{enumerate}[a.]
    \item \label{it:linear-rademacher-complexity} For any $d \in \R$ and $x \in \R^d$ with $\norm{x}_2 \leq 1$. The function class $\mcf = \{ x \mapsto  \langle w, x \rangle + b  \, \;\; | \;\; \, \norm{w}_2 \leq B, \abs{b} \leq B \}$ has Rademacher complexity $\empR \rb{ \mcx, \mcf } \leq \frac{2 B}{ \sqrt{n} } $.
    \item \label{it:addition-property-rademacher-complexity} Given classes $\mcf_1, \mcf_2$ functions, $\empR \rb{ \mcx; \mcf_1 + \mcf_2 } = \empR \rb{ \mcx; \mcf_1 } + \empR \rb{ \mcx; \mcf_2 }$.
    \item \label{it:neural-network-type-rademacher-complexity} Given classes $\mcf_1, \mcf_2, ..., \mcf_m$ of functions of type $\mcx \rightarrow \R$ and suppose $w \in \R^m$ is a fixed vector, then $\mcf' = \{ x \mapsto \sum_{r=1}^m w_r \sigma \rb{ f_r(x) } \,|\, f_r \in \mcf_r  \}$ satisfies $\empR \rb{ \mcx ; \mcf' } \leq 2 \| w \|_1 \max_{r \in [m]} \empR \rb{ \mcx; \mcf_r } $ where $\sigma$ is a 1-Lipschitz continuous function. 
\end{enumerate}
\end{lemma}
\begin{proof} These are standard results and can be found in \cite{allen2019learning} and \cite{shalev2014understanding}. 
\end{proof}
\todo{The follwoing is not clear. In particular, what's $\delta$? Do we need both conclusions or it sufficient to have the latter?}
\begin{fact} \label{fact:population-empirical-loss-diff-rademacher} (Rademacher Complexity) If $\mcf_1, \mcf_2, ..., \mcf_k$ are classes of functions of type $\R^d \rightarrow \R$ and $L_x: \R^d \rightarrow [-b, b]$ is a $L_g$-Lipschitz-continuous function for every $x$ in the support of $\mcd$, then 
\begin{align*}
    \sup_{ f_1 \in \mcf_1, ..., f_k \in \mcf_k } \abs{ \E_{x \in \mcd} \left[ L_x \rb{ f_1(x), ..., f_k(x) } \right] - \frac{1}{n} \sum_{i=1}^n L_x \rb{ f_1(x_i), ..., f_k(x_i) } } \leq 2 \empR \rb{ \mcx; \mathcal{L} } + b \sqrt{ \frac{\log \frac{1}{\delta}}{ 2n } }
\end{align*}
where $\mathcal{L}$ is set of functions obtained by composing $L_x$ with $\mcf_1, \mcf_2, ..., \mcf_k$, that is $\mathcal{L}:=\{L_x \circ (f_1, \ldots, f) \mid f_1 \in \mcf_1, \ldots, f_k \in \mcf_k \}$. Using vector contraction inequality from \cite{maurer2016vector}, we get
\begin{align*}
    \sup_{ f_1 \in \mcf_1, ..., f_k \in \mcf_k } & \abs{ \E_{x \in \mcd} \left[ L_x \rb{ f_1(x), ..., f_k(x) } \right] - \frac{1}{n} \sum_{i=1}^n L_x \rb{ f_1(x_i), ..., f_k(x_i) } } \\ 
    &\leq 2 \sqrt{2} L_g \rb{ \sum_{i=1}^k \empR \rb{ \mcx; \mathcal{F}_i } } + b \sqrt{ \frac{\log \frac{1}{\delta}}{ 2n } }.
\end{align*}
\end{fact}

\end{document}